\documentclass[pdflatex,sn-nature]{sn-jnl}

\usepackage{graphicx}%
\usepackage{multirow}%
\usepackage{amsmath,amssymb,amsfonts}%
\usepackage{amsthm}%
\usepackage{mathrsfs}%
\usepackage[title]{appendix}%
\usepackage{xcolor}%
\usepackage{textcomp}%
\usepackage{manyfoot}%
\usepackage{booktabs}%
\usepackage{algorithm}%
\usepackage{algorithmicx}%
\usepackage{algpseudocode}%
\usepackage{listings}%
\usepackage{url}            
\usepackage{amsfonts}       
\usepackage{nicefrac}       
\usepackage{microtype}      
\usepackage{times}
\usepackage{soul}
\usepackage{makecell}
\usepackage{bm}
\usepackage[stable]{footmisc}
\usepackage{mathrsfs}
\usepackage{enumitem}
\usepackage{colortbl}
\usepackage{lineno}
\usepackage[numbers,super,sort&compress]{natbib}
\usepackage{hyperref}
\usepackage{multirow} 
\usepackage{ragged2e} 
\usepackage{booktabs} 

\hypersetup{citecolor=black,linkcolor=black,urlcolor=black,colorlinks=true}
\makeatletter
\renewcommand\@biblabel[1]{#1.} 
\renewcommand\@cite[1]{#1}    
\renewenvironment{thebibliography}[1]
 {
  \list{\@biblabel{\arabic{enumiv}}}
       {\settowidth\labelwidth{\@biblabel{#1}}%
        \leftmargin\labelwidth
        \advance\leftmargin\labelsep
        \usecounter{enumiv}}%
  \sloppy\clubpenalty4000\@clubpenalty\clubpenalty\widowpenalty4000%
  \sfcode`\.\@m}
 {\endlist}
\makeatother

\theoremstyle{thmstyleone}%
\theoremstyle{thmstyletwo}%

\theoremstyle{thmstylethree}%

\newcommand{\method}{Pangaea}

\usepackage{multirow}
\usepackage{booktabs,makecell}

\begin{document}
\title{AI \method{}: Unifying Intelligence Islands for \\ Adapting Myriad Tasks}
\def\method{Pangaea}
\newcommand{\reviewone}[1]{\textcolor{red}{#1}}
\newcommand{\reviewtwo}[1]{\textcolor{green}{#1}}
\newcommand{\reviewthree}[1]{\textcolor{blue}{#1}}

\author*[1]{\fnm{Jianlong} \sur{Chang}}\email{}

\author[2]{\fnm{Haixin} \sur{Wang}}\email{Project Page: \href{https://jianlongchange.github.io/AI_Pangaea}{\textcolor{blue}{jianlongchange.github.io/AI\_Pangaea}}}

\author[1]{\fnm{Zhiyuan} \sur{Dang}}
\author[1]{\fnm{Li} \sur{Huang}}
\author[1]{\fnm{Zhiyu} \sur{Wang}}
\author[1]{\fnm{Ruoqi} \sur{Cao}}
\author[1]{\fnm{Shihao} \sur{Piao}}
\author[1]{\fnm{Dongzhe} \sur{Li}}
\author[1]{\fnm{Dianyu} \sur{Gao}}
\author[1]{\fnm{Dongsheng} \sur{Wang}}
\author[1]{\fnm{Yin} \sur{Li}}
\author[2]{\fnm{Jinan} \sur{Sun}}
\author*[3]{\fnm{Lu} \sur{Fang}}
\author*[4]{\fnm{Zhouchen} \sur{Lin}}

\affil[1]{\orgdiv{Huawei}, \state{Beijing}, \country{China}}

\affil[2]{\orgdiv{National Engineering and Research Center for Software Engineering, \\ Peking University}, \state{Beijing}, \country{China}}

\affil[3]{\orgdiv{Department of Electronic Engineering, Beijing National Research Center for Information Science and Technology, Tsinghua University}, \state{Beijing}, \country{China}}

\affil[4]
{\orgdiv{State Key Laboratory of General AI, School of Intelligence Science and Technology, Peking University}, \state{Beijing}, \country{China}}


\abstract{
The pursuit of artificial general intelligence continuously demands generalization in one model across myriad tasks, even those not seen before.
However, current AI models are isolated from each other for being limited to specific tasks, now first defined as Intelligence Islands.
To unify Intelligence Islands into one, we propose Pangaea, the first AI supercontinent akin to the geological Pangaea. 
Pangaea encodes any data into a unified format and accumulates universal knowledge through pre-training on 296 datasets across diverse modalities.
Eventually, it demonstrates remarkable generalization across 45 general tasks and 15 scientific tasks encompassing a wide range of scientific subjects.
By investigating Pangaea deeper, the scaling effect of modality is revealed, quantifying the universal knowledge accumulation across modalities as the cumulative distribution function of a geometric distribution. 
On the whole, Pangaea shows strong potential to handle myriad tasks, indicating a new direction toward artificial general intelligence.
}





\maketitle

\begin{figure}[!h]
\centering
\includegraphics[width=\textwidth,keepaspectratio=true]{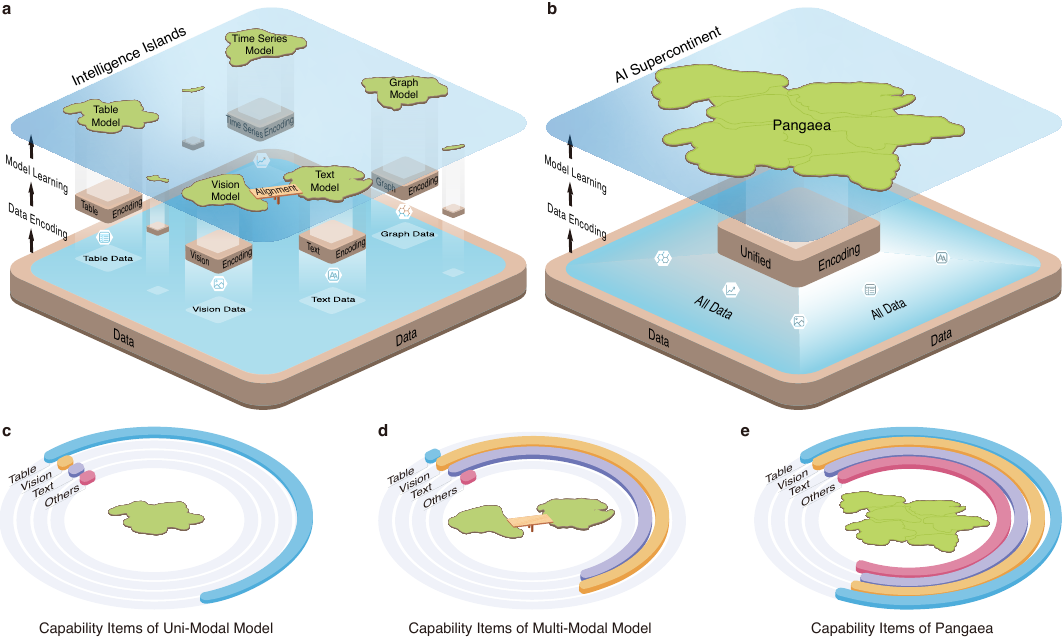}
\vspace{-0.5cm}
\caption{\textbf{Intelligence Islands and AI Supercontinent.} 
\textbf{a}, AI models build intelligence from data relying on modality-specific data encodings, leading to Intelligence Islands. This issue persists even in current multi-modal models, which merely align multiple uni-modal models.
\textbf{b}, \method{} unifies Intelligence Islands through unified data encoding, constructing an AI supercontinent.
\textbf{c}, Existing uni-modal models are only capable of handling tasks of predefined modalities.
\textbf{d}, Even with the alignment of multiple uni-modal models, existing multi-modal models are still limited to handling tasks of predefined modalities.
\textbf{e}, \method{} is capable of handling tasks of any modalities.
}
\vspace{-0.2cm}
\label{fig:framework}
\end{figure}

Since the Dartmouth workshop in 1956, AI models have evolved into powerful tools for emulating humans by learning knowledge from data.
Ranging from logistic regression~\cite{el2024logistic} to random forest~\cite{Delavaux2023forest}, convolutional neural networks~\cite{lecun2015deep}, graph neural networks~\cite{Gehrig2024gnns}, and transformers~\cite{xu2024whole}, researchers have focused mainly on building increasingly powerful yet distinct models for specific tasks, resulting in countless isolated intelligences that cannot scale across modalities.
To investigate this isolation, the building process from data to intelligence is revisited, including data encoding and model learning stages, as shown in Fig.~\ref{fig:framework}a.
For a start, data are categorized into modalities based on their diverse topological structures and statistical properties. 
These differences lead to distinct data encodings across modalities, \textit{e.g.}, text data is represented as sequences of word embeddings that capture semantic distributions, while image data is encoded as grids of pixel vectors that reflect local texture patterns.
Existing AI models are designed for these modality-specific data encodings and learn from specific modalities, resulting in isolated intelligences built on predefined modalities, as shown in Fig.~\ref{fig:framework}c.
This isolation even exists in current multi-modal models which merely align multiple uni-modal models and are still limited to predefined modalities as illustrated in Fig.~\ref{fig:framework}d. 
Until now, this isolation has not been acknowledged, let alone fundamentally addressed. For the first time, we define this isolation as \textit{Intelligence Islands} in this manuscript and attempt to handle it in a unified manner.

Intelligence Islands bring three significant limitations to future development of AI. 
First, Intelligence Islands prevent AI models from being standardized as an artificial general intelligence (AGI) model across countless real-world tasks. 
Despite the explosive growth of AI in recent years across tasks in vision~\cite{qi2017pointnet,DBLP:conf/cvpr/HeZRS16,vaswani2017attention,Tian2020Contrastive}, time series~\cite{Yuqietal-2023-PatchTST,zeng2023are,Du0FPQXW21}, graph~\cite{kipf2017semi,rong2020self,velickovic2023gnn}, table~\cite{hollmann2025accurate,XGBoost,Qin2010letor}, and text~\cite{radford2018improving,devlin-etal-2019-bert,achiam2023gpt} modalities, addressing these tasks requires architectures~\cite{fatemi2024talk,dinh2022lift} tailored to the unique characteristics of each modality.
In this way, AI models achieve performance improvements on specific tasks at the expense of handling other tasks.
Therefore, the numerous real-world tasks require endless unique models consuming massive amounts of data and computational resources, making it unattainable to support a rapid intelligence explosion beyond the human level.
Second, Intelligence Islands lead to a lack of holistic understanding and knowledge accumulation in AI models. Far from the AGI model that can accumulatively gain and integrate knowledge from multiple modalities, AI models designed for text recognition are not equipped to handle vision tasks. 
Transferring knowledge from one modality to another is challenging~\cite{shen2023crossmodal,mirchandani2023llm,lu2022frozen}, making it difficult to apply insights gained in one area to facilitate learning in another.
This prevents AI from achieving a holistic understanding and restricts mutual cognitive growth across different modalities.
Third, Intelligence Islands render existing AI models incapable of utilizing the full spectrum of accessible data. 
Empirically, the performance of AI models on specific tasks generally improves as data increases.
Nevertheless, due to the predefined modalities in AI models, only a small fraction of the entire available data can be utilized, leaving a large margin for improvement.
In some data-scarce scientific subjects including Humanities~\cite{Jiang2025Investigating}, Mathematical~\cite{trinh2024solving}, Physical~\cite{xiong2025bridging}, Biological~\cite{jumper2021highly}, and Earth and environment sciences~\cite{irrgang2021towards}, this margin further enlarges.
This data underutilization leads to inefficiency of AI models in many scenarios, particularly in those with high data acquisition costs.

The breakthrough appears to be occurring sooner than expected. In this manuscript, \method{} is proposed to unify Intelligence Islands and construct the first AI supercontinent shown in Fig.~\ref{fig:framework}b, similar to the geological Pangaea. 
Realizing that the root cause of Intelligence Islands lies in the differences between data encodings, \method{} proposes a unified encoding based on triplet sets and develops a triplet transformer suitable for the order-irrelevant and quantity-variant characteristics of triplet sets. 
Benefiting from such unified modeling, a parallel reconstruction strategy is designed to guide \method{} in learning universal knowledge from unlabeled, unaligned data across diverse modalities.
Pre-trained on 296 datasets across text, table, vision, graph, and time series modalities, \method{} was subsequently evaluated via extensive experiments. These evaluations covered 45 general tasks across seven modalities, with audio and point cloud newly added, and 15 scientific tasks spanning diverse scientific subjects.
The remarkable experimental results of all 60 tasks demonstrate that \method{} transcends modality restrictions, highlighting its capability to handle any modalities, as shown in Fig.~\ref{fig:framework}e.
Through a deeper study of \method{}, the scaling effect of modality is first revealed in AI, showing that integrating more modalities leads to the acquisition of richer universal knowledge. As a result, this charts a promising blueprint for AGI.

\begin{figure}[!h]
\centering
\includegraphics[width=\textwidth,keepaspectratio=true]{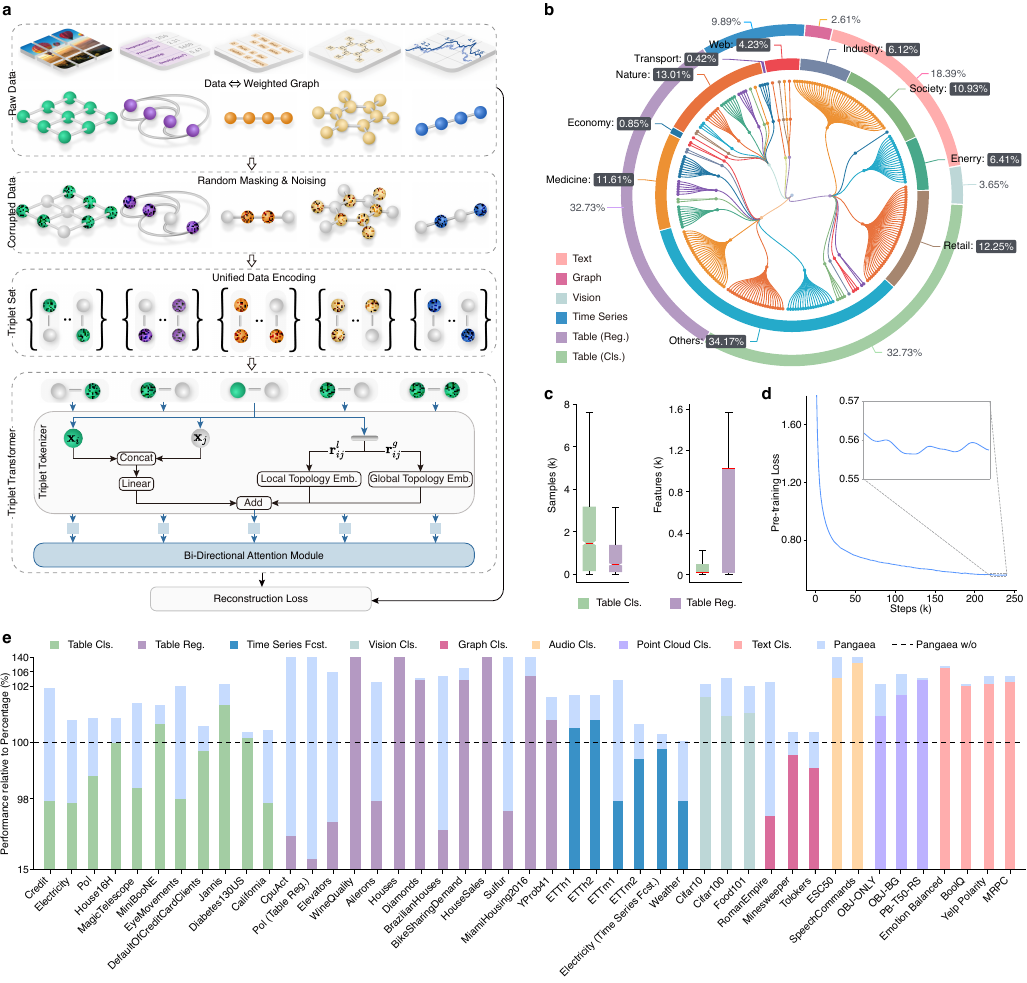}
\vspace{-0.5cm}
\caption{\textbf{Architecture, pre-training, datasets, and performances.} 
\textbf{a}, Unified encoding, triplet transformer and pre-training. Raw data is mathematically abstracted as a weighted graph, corrupted by masking and noising, and encoded into a triplet set for parallel reconstruction by the triplet transformer.
\textbf{b}, Pre-training dataset. The outer ring shows five modalities, the middle ring displays ten domains, and the inner tree diagram illustrates the relationship between datasets, domains, and modalities.
\textbf{c}, Sample and feature distributions of table datasets for pre-training.
In box plots, the central lines mark the median, hinges mark the 25th and 75th percentiles and whiskers show all values that, at maximum, fall within 1.5 times the interquartile range.
\textbf{d}, Pre-training convergence curve of \method{}.
\textbf{e}, Performances of the \method{} exceed those of both \method{}$_\text{w/o}$ and competitive models across 45 downstream tasks. Cls., Classification; Reg., Regression; Fcst., Forecasting; Emb., Embedding.
}
\vspace{-0.2cm}
\label{fig:main}
\end{figure}

\subsection*{Accumulating universal knowledge from all modalities}
Distinct data encodings arise from the varying topological structures and statistical properties across modalities, leading to Intelligence Islands.
To surmount this obstacle, \method{} unifies data encodings by converting any data to a triplet set, enabling scalability across all modalities.
Intrinsically, any data can be treated as a set of numerical values embedded within specific topological structures, which is mathematically abstracted as a weighted graph, as shown in Fig.~\ref{fig:main}a. For a weighted graph $\mathcal{G}=(\mathcal{V},\mathcal{E})$, $\mathcal{V}$ is a vertex set representing numerical values, $\mathcal{E}$ is an edge set that quantifies the topological relation between any two values. 
By deconstructing the graph structure, each weighted graph is equivalently transformed into a triplet set $\big\{(u,e_{uv},v)\mid u,v\in\mathcal{V}, e_{uv}\in\mathcal{E}\big\}$, where each triplet consists of two values connected by a weighted edge. 
Consequently, the triplet $(u,e_{uv},v)$ is regarded as the fundamental unit of data, like atoms in a molecule, serving as the basis for unifying data encodings.
To efficiently implement the triplet, its parts are vectorized respectively in practice, enabling any data $\textbf{x}$ to be structured as a triplet set $\big\{(\textbf{x}_i,\textbf{r}_{ij},\textbf{x}_j)\big\}_{ij}$.
In this context, $\textbf{x}_i$ and $\textbf{x}_j$ denote the $i$-th and $j$-th constructed numeric parts of $\textbf{x}$, represented as vectors of sampled numerical values. 
The topology part $\textbf{r}_{ij}$ of $\textbf{x}$ includes the local topology $\textbf{r}^l_{ij}$ to measure the relation between $\textbf{x}_i$ and $\textbf{x}_j$, and the global topology $\textbf{r}^g_{ij}$ to represent the absolute position of $\textbf{r}^l_{ij}$. By leveraging this unified encoding, data is encoded in a unified manner using triplets, paving the way for learning across all modalities.

With the aim of learning from such a triplet set and considering its order-irrelevant and quantity-variant characteristics, a triplet transformer is developed with a tokenizer and a bi-directional attention module, as illustrated in Fig.~\ref{fig:main}a.
The triplet tokenizer converts each triplet $\left(\textbf{x}_{i},\textbf{r}_{ij},\textbf{x}_{j}\right)$ into a triplet token. 
Specifically, the numeric parts $\textbf{x}_i$ and $\textbf{x}_j$ are concatenated and then mapped via a shared linear layer. The local topology part $\textbf{r}^l_{ij}$ is implemented as a learnable topology embedding, added to the mapped numeric part.
Eventually, these three parts of one triplet form a triplet token. 
Besides, the global topology part $\textbf{r}^g_{ij}$, which distinguishes this triplet from others, is represented as a position embedding in practice.
Taking these triplet tokens as input, the attention module is used to capture the relationships between them.
With this fine-grained processing, the triplet transformer captures the inter-triplet and intra-triplet correlations, thereby enabling high-level representation learning on triplet sets.

In light of the unified encoding and the triplet transformer, \method{} is pre-trained across various modalities in a parallel reconstruction strategy, as illustrated in Fig.~\ref{fig:main}a.
This strategy divides multi-modal reconstruction into uni-modal levels, where raw data from each modality is parallelly recovered using its corresponding triplet set.
Benefiting from this, \method{} is capable of simultaneously learning from various modalities without alignment or label.
As shown in Fig.~\ref{fig:main}b, \method{} is pre-trained across 296 datasets spanning text, table, vision, graph, and time series modalities. Notably, 243 of these datasets are table-based, significantly enhancing data diversity due to their variations in scale and structure depicted in Fig.~\ref{fig:main}c.
Technically, \method{} optimizes using a masked reconstruction paradigm and then backpropagates the average gradient to learn universal knowledge across modalities rather than modality-specific ones. 
In Fig.~\ref{fig:main}d, the loss curve converges, demonstrating the effectiveness of the pre-training strategy.

Extensive experiments have demonstrated that \method{} solves a wide range of tasks in a unified way with performance beyond expectations.
We evaluated \method{} on 45 downstream tasks, including 24 for table modality, six for time series, three for vision, two for audio, three for graph, three for point cloud, and four for text.
These tasks, especially the table ones, exhibited clear distinctions in both data structures and dimensions.
Traditionally, such diverse tasks would require at least seven types of models, one for each modality. 
Here, seven competitive models~\cite{DBLP:conf/cvpr/HeZRS16,gong2021ast,XGBoost,liu2023itransformer,bo2021lowfrequencyinformationgraphconvolutional,qi2017pointnetplusplus,devlin-etal-2019-bert} were selected for comparison.
Conversely, \method{} as a unified model successfully addressed all 45 tasks, demonstrating its potential to address any task.
Moreover, \method{} surpassed these competitive models on all 45 tasks, with an average improvement of 7.5\%.
Several of the competitive models were pre-trained, \textit{i.e.}, ResNet~\cite{DBLP:conf/cvpr/HeZRS16}, AST~\cite{gong2021ast}, BERT~\cite{devlin-etal-2019-bert} while others were not, \textit{i.e.}, XGB~\cite{XGBoost}, iTransformer~\cite{liu2023itransformer}, FAGCN~\cite{bo2021lowfrequencyinformationgraphconvolutional}, PointNet++~\cite{qi2017pointnetplusplus}.
The consistent improvement over these models demonstrated the acquisition of universal knowledge across diverse modalities, which seamlessly transferred to downstream tasks.
This conclusion was further supported by a 5.4\% average improvement of \method{} over its version without pre-training denoted \method{}$_\text{w/o}$, confirming the superiority of pre-training.

The universal knowledge learned from pre-training helps \method{} improve performance in more isolated scenarios, as well as adapt to new modalities.
Beyond the consistent improvement over competitive models, \method{} delivered a more impressive 10.7\% enhancement on table tasks.
Given that each table task typically requires an entirely distinct model, knowledge from specific data is isolated across tasks, even if data share the same topology.
Such a heavy reliance on the specific data exists widely in practical applications and is especially severe in table tasks, where there are only an average of 20,000 samples per task.
In contrast, universal knowledge mitigates the dependence on data, as evidenced by the further improvement of \method{} in data-scarce scenarios.
Even in the absence of audio and point cloud data during pre-training, \method{} still achieved a 6.5\% average improvement on audio and point cloud tasks compared with \method{}$_\text{w/o}$.
This advantage suggested that although universal knowledge was learned from predefined modalities, it kept the applicability in any modality, thus transcending modality boundaries.
Building on the advantages mentioned above, \method{} surmounts obstacles in AI development to broaden its application scope in the real-world.

\subsection*{Transferring universal knowledge to scientific tasks}
\begin{figure}[!t]
\centering
\includegraphics[width=\textwidth,keepaspectratio=true]{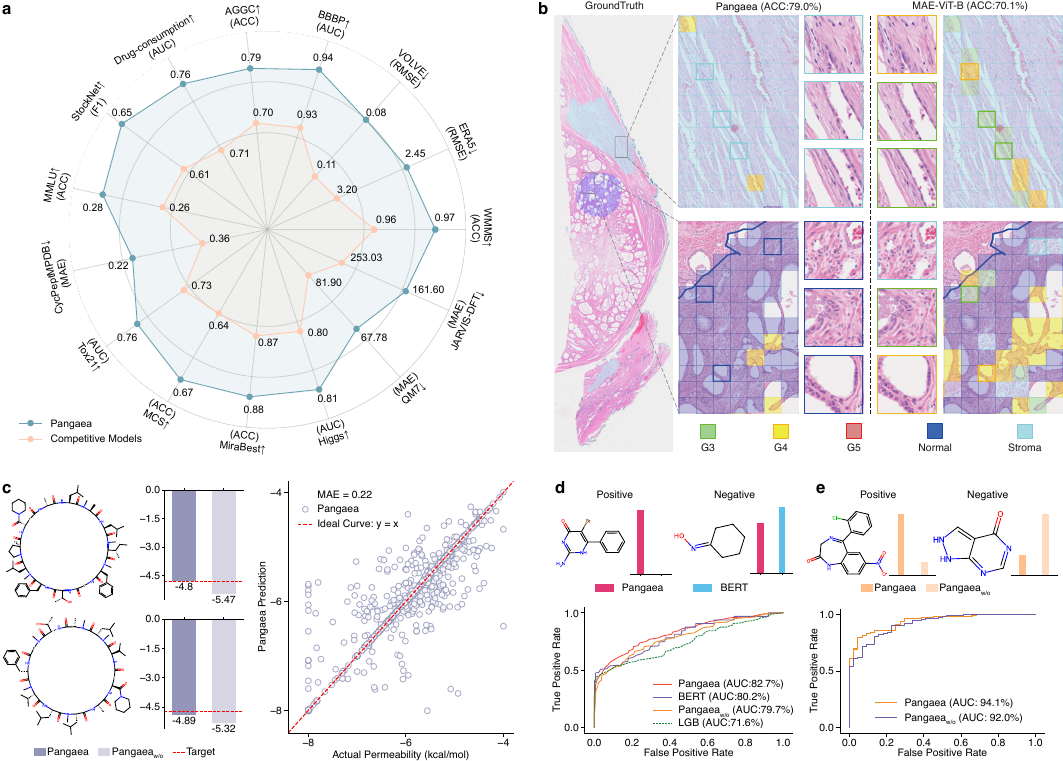}
\vspace{-0.5cm}
\caption{\textbf{\method{} is applied in Health and Biological sciences.}
\textbf{a}, Overview of the performance comparison between \method{} and competitive models on all 15 scientific tasks.
\textbf{b}, The left panel plots a whole slide image with annotations for prostate cancer grading, and case studies in Stroma and Normal regions visualize the classification results of ROI. 
\textbf{c}, Case studies and \method{} predictions of cyclic peptide membrane permeability.
\textbf{d}, ROC curves and case studies of \method{} and competitive models for drug molecule toxicity prediction on the NR-AR pathway.
\textbf{e}, Performance of \method{} and \method{}$_\text{w/o}$ on blood-brain barrier penetration prediction, including ROC curves and illustrative examples.
}
\vspace{-0.2cm}
\label{fig:ai4s}
\end{figure}

Compared to 45 general tasks, scientific tasks often struggle with severe Intelligence Islands due to data scarcity.
To mitigate this, one promising approach involves transferring universal knowledge to these scientific tasks.
Specifically, \method{} exhibits superior performance across 15 diverse scientific tasks, as shown in Fig.~\ref{fig:ai4s}a. Spanning nine diverse scientific subjects including Health, Biological, Earth and environmental, Physical, Business and commerce, Humanities, Astronomy, Mathematical, and Social sciences, these tasks validate its capability in extensive scientific tasks and its effectiveness in mitigating data scarcity through pre-training. A comprehensive performance overview for each task is detailed in Extended Data Table~\ref{tab:ai4s_res_info}.

\begin{figure}[!t]
\centering
\includegraphics[width=\textwidth,keepaspectratio=true]{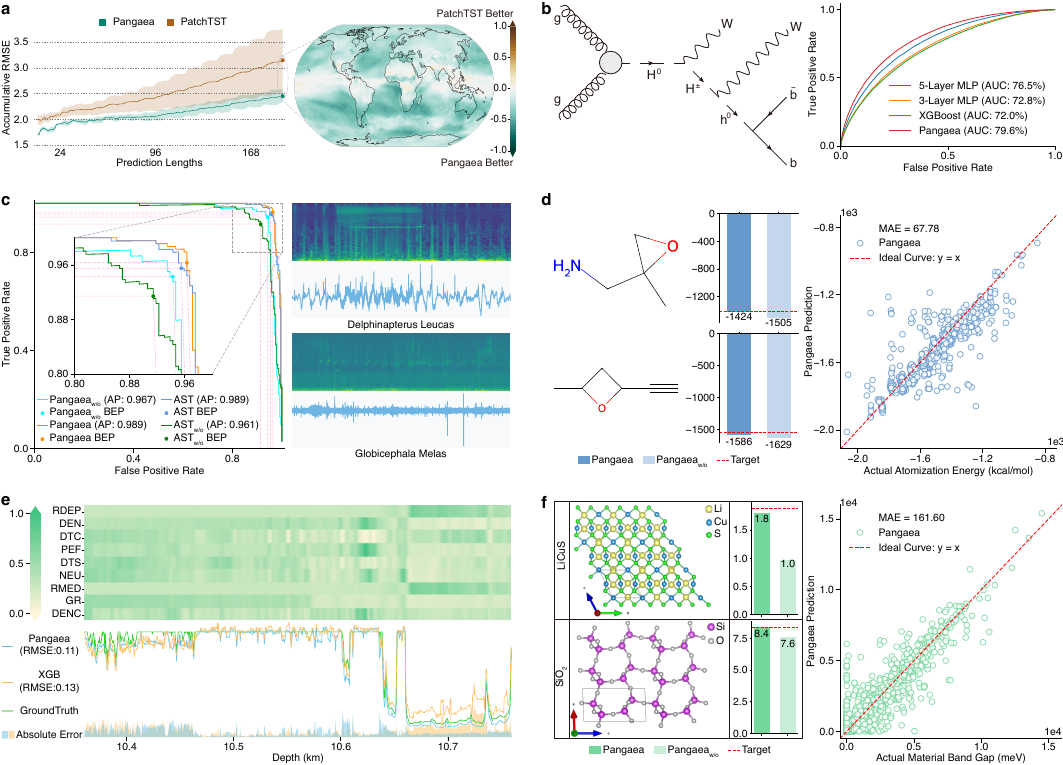}
\vspace{-0.5cm}
\caption{\textbf{\method{} is applied in Earth and environmental, and Physical sciences.}
\textbf{a}, Cumulative RMSE curve for 192 months of global temperature forecasts and heatmap of RMSE difference for worldwide temperature forecasts in the 192nd month.
\textbf{b}, The signal process involving new exotic Higgs bosons and ROC curve of \method{} and competitive models based on kinematic features.
\textbf{c}, Visualization of marine mammal voice samples and PR curves comparing \method{} with competitive models for vocalization classification.
\textbf{d}, Case studies and \method{} predictions of molecule electronic properties.
\textbf{e}, Raw data distribution with depth visualized as a heatmap for reservoir property estimation. Shale volume measurements and predictions by \method{} and XGB in well 103 are plotted along the shared depth axis.
\textbf{f}, Case studies and \method{} predictions of material band gap.
}
\vspace{-0.2cm}
\label{fig:ai4s2}
\end{figure}

\method{} was extensively evaluated across key tasks in both Health and Biological sciences. 
For Health sciences, we investigated prostate cancer grading using the AGGC dataset~\cite{huo2024comprehensive} and blood-brain barrier penetration prediction using the BBBP dataset~\cite{sakiyama2021prediction}, as shown in Fig.~\ref{fig:ai4s}b, e.
\method{} surpassed MAE-ViT-B~\cite{he2022masked} by 12.7\% on AGGC with 79.0\% accuracy, and demonstrated a 1.2\% improvement over MTL-BERT~\cite{zhang2022pushing} on BBBP.
In Biological sciences, it was assessed on the CycPeptMPDB~\cite{li2023cycpeptmpdb} dataset for cyclic peptide membrane permeability prediction and the Tox21 dataset~\cite{huang2016tox21challenge} for drug toxicity prediction, as detailed in Fig.~\ref{fig:ai4s}c, d.
\method{} achieved SOTA performance on CycPeptMPDB with a 0.2 MAE and showed AUC improvements of 4.6\% over pre-trained BERT and 6.9\% over LGB~\cite{NIPS2017_lgb} on Tox21.
Notably, \method{} with pre-training obtained improvements of 1.9\% on CycPeptMPDB and 7.3\% on Tox21 compared to \method{}$_\text{w/o}$, indicating the effective transfer of universal knowledge from pre-training.

\method{} exhibited robust performance in Earth and environmental sciences, effectively tackling three distinct yet critical tasks, as detailed in Fig.~\ref{fig:ai4s2}a, c, e. 
For global temperature forecasting on the ERA5 dataset~\cite{hersbach2020era5}, \method{} utilized temperature data from the past 96 months to accurately predict long-term trends for the subsequent 192 months, resulting in a 23.4\% reduction in RMSE compared to the PatchTST~\cite{Yuqietal-2023-PatchTST}. For marine mammal vocalization classification, it successfully distinguished voices on the WMMSD dataset~\cite{sayigh2016watkins} with an impressive 96.6\% accuracy. In reservoir property estimation on the VOLVE dataset~\cite{fu2024well}, \method{} demonstrated a 29.2\% performance improvement over the XGB model across four test oil wells. These results collectively underscore \method{}'s robust capability in processing various forms of data, thereby demonstrating strong generalization across diverse domains and tasks.

\method{} demonstrated its robustness across diverse tasks and data volumes in Physical science.
For high-energy particle identification, as shown in Fig.~\ref{fig:ai4s2}b, it achieved an 81.0\% AUC on the Higgs dataset~\cite{baldi2014searching}, which comprises 110,000,000 tabular samples, demonstrating its capability in handling large-scale datasets.
\method{} was also evaluated on two smaller, graph-formatted tasks, namely QM7~\cite{blum2009gdb13} with 7,160 samples for molecule electronic properties prediction, and JARVIS-DFT~\cite{choudhary2014jarvis} using 18,172 samples for material band gap prediction, as shown in Fig.~\ref{fig:ai4s2}d, f. 
On QM7, it achieved a 17.2\% MAE reduction compared to XGB.
For JARVIS-DFT, it notably achieved SOTA performance with a MAE of 161.6 and demonstrated a substantial 36.1\% MAE reduction over CartNet~\cite{sole2025cartesian}.
These results further demonstrate \method{}'s consistent efficiency across varying data volumes.

\begin{figure}[!t]
\centering
\includegraphics[width=\textwidth,keepaspectratio=true]{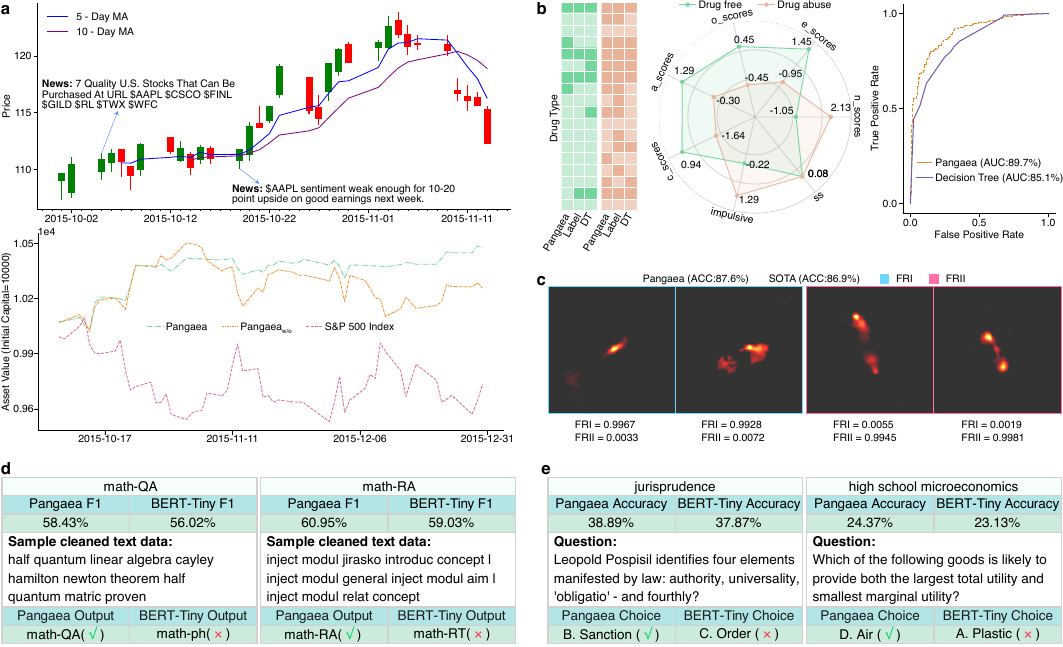}
\vspace{-0.5cm}
\caption{\textbf{\method{} is applied in Business and commerce, Humanities, Astronomy, Mathematical, and Social sciences.}
\textbf{a}, Daily stock movement of Apple Inc., and portfolio cumulative returns comparison between \method{} and \method{}$_\text{w/o}$ over 46 companies.
\textbf{b}, Prediction results and personality trait distribution for drug-free versus drug abuse individuals. Also shown is the ROC curve for \method{} and decision tree on cannabis.
\textbf{c}, Radio image samples and \method{} predictions for classifying active galactic nuclei.
\textbf{d}, Classification results of \method{} and BERT-Tiny for mathematical articles. 
\textbf{e}, Q\&A comparisons of \method{} and BERT-Tiny for massive multitask language understanding.
}
\vspace{-0.2cm}
\label{fig:ai4s3}
\end{figure}

\method{} also showcased its versatility in complex domains like Astronomy and Mathematics, as illustrated in Fig.~\ref{fig:ai4s3}c, d. 
For Astronomy tasks, \method{} achieved 87.6\% accuracy on the MiraBest dataset~\cite{porter2023mirabest} with 1,222 samples for classifying active galactic nuclei by their radio images, which is comparable to SOTA results~\cite{slijepcevic2022radio,scaife2021fanaroff}.
In Mathematics, \method{} achieved an accuracy of 67.4\% in subject classification on the MSC2020 dataset~\cite{dunne2020mathematics}, which consists of 159,175 samples derived by stemming the titles and abstracts of pre-print papers. It surpassed BERT-Tiny~\cite{devlin-etal-2019-bert}, a model with a comparable parameter size, by 3.7\%. 
This further illustrates the broad applicability of universal knowledge, extending to highly specialized and abstract fields.

\method{} further demonstrated excellent performance in other scientific subjects, as detailed in Fig.~\ref{fig:ai4s3}a, b, e.
In Business and commerce, \method{} was validated on a stock movement prediction task incorporating both Twitter data and historical stock prices, achieving an F1 score of 65.5\% and surpassing MAN-SF~\cite{sawhney2020deep} by 5.0\%.
In Humanities, using the drug consumption dataset~\cite{fehrman2017five}, \method{} classified drug consumer types based on ten personality traits. 
It outperformed DT~\cite{quinlan1986induction} across all 18 drug types, such as cannabis and LSD, with an average AUC improvement of 6.7\%.
In Social science, \method{} was validated on MMLU~\cite{hendrycks2021measuring}, a massive multitask test consisting of multiple-choice questions spanning 57 sub-tasks from various subjects. \method{} achieved 27.8\% overall accuracy, notably outperforming BERT-Tiny by 1.6\% and standing comparable to much larger models like RoBERTa~\cite{liu2019roberta} (27.9\%) and ALBERT~\cite{lan2020albert} (27.1\%).

In conclusion, \method{} consistently delivered strong performance across diverse scientific tasks spanning a wide range of subjects, demonstrating a robust capability to perform effectively even when data is scarce. This remarkable generalization stems from its leveraging of universal knowledge, which enabled \method{} to outperform competitive models without structural adjustments and operate as a unified model, marking a significant stride towards AGI. All detailed results are depicted in Extended Data Fig.~\ref{app_fig:prostate_grading}--\ref{app_fig:mmlu}.

\subsection*{Revealing the scaling effect of modality}
The remarkable generalization capability of \method{} across 60 tasks spanning seven modalities motivated further investigation. These gains arise from the universal knowledge accumulated across five pre-training modalities. 
This knowledge, which transcends modality restrictions, suggests the emergence of a scaling effect of modality where performance scales with the growing number of incorporated modalities.
Building on this insight, we quantify the performance gains of the scaling effect, reflecting the universal knowledge accumulation across modalities.

To leverage the duality, independence, and discreteness characteristics of the knowledge accumulation process, the Bernoulli distribution is employed to model the universal knowledge gained from a single modality, thereby revealing the scaling effect of modality as the cumulative distribution function of a geometric distribution. 
Universal knowledge learned from a single modality is modeled by a Bernoulli trial with success probability $p$, and the expectation of $p$ is the amount of learned universal knowledge.
When extended to $k\in\mathbb{N}^+$ modalities, the probability of universal knowledge learned from at least one is given by $1-(1-p)^k$.
Its expectation quantifies the accumulated amount of universal knowledge after learning from $k$ modalities. 
It aligns with the scaling effect of modality, which is expressed as $y=1 - (1 - p)^x$, where $x$ is the number of modalities and $y$ represents the accumulated amount of universal knowledge.
The upper bound of $1$ indicates that this amount eventually nears saturation, with diminishing returns of each additional modality.

\begin{figure}[!t]
\centering
\includegraphics[width=\textwidth,keepaspectratio=true]{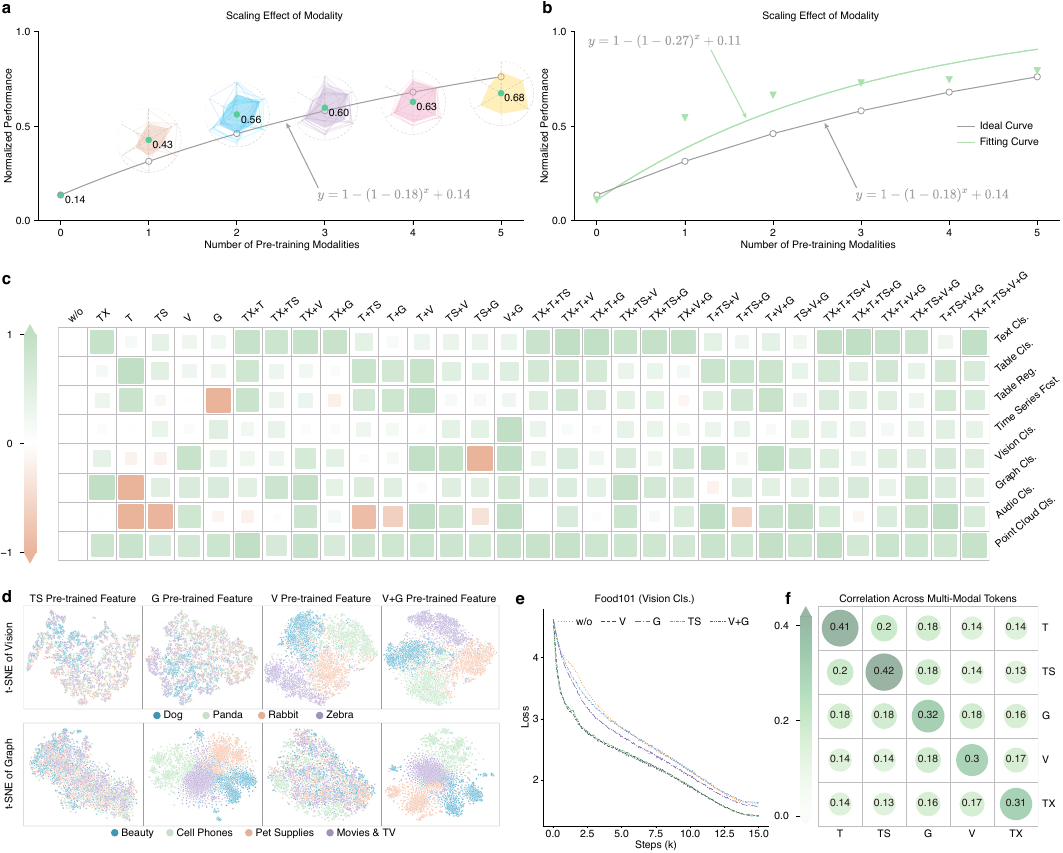}
\vspace{-0.5cm}
\caption{\textbf{Discoveries related to the scaling effect of modality.} \textbf{a}, Scaling effect of modality, with the extended equation $y=1-(1-0.18)^x+0.14$. Each vertex of the color-filled area in the radar plots, with the zoomed version presented in Extended Data Fig.~\ref{app_fig:radar_plot}, corresponds to the average normalized performance for a specific task type. \textbf{b}, Comparison between the ideal curve of scaling effect of modality and the fitting curve from performances (triangle mark) on unseen modalities. \textbf{c}, Affinity phenomenon of modality, comparing the fine-tuning performances of \method{} across 31 pre-training modality combinations to the ones of \method{}$_\text{w/o}$, with green indicating enhancement and red indicating reduction. 
\textbf{d}, The t-SNE visualization of vision and graph data of feature distributions derived by \method{} pre-trained on specific modality combinations. \textbf{e}, Convergence curves of fine-tuning loss on a vision task across different pre-training modality combinations. \textbf{f}, Averaged attention weights between tokens of pre-training modalities. TX, Text; T, Table; TS, Time Series; G, Graph; V, Vision; w/o, \method{}$_\text{w/o}$; $+$, and.
}
\vspace{-0.2cm}
\label{fig:ablation_exp}
\end{figure}

To determine an appropriate $p$ in the scaling effect of modality, we implemented all possible pre-training combinations across five modalities correlating the average normalized downstream performance $y$ with the number of pre-training modalities $x$, as shown in Fig.~\ref{fig:ablation_exp}a.
In these downstream experiments, two types of knowledge were quantified. Pre-training universal knowledge followed the scaling effect $y=1-(1-p)^{x}$, and fine-tuning knowledge was defined as a constant $c$ unaffected by $x$. Overall, the scaling effect was extended to $y=1-(1-p)^x+c$. 
After calculating $y$ for $x=\{0,1,2,3,4,5\}$ in the experiments, six scatter points were plotted in Fig.~\ref{fig:ablation_exp}a. 
By fitting these points, the scaling effect of modality was formulated as $y=1-(1-0.18)^x+0.14$.
Its applicability was further validated on downstream tasks involving unseen modalities, with the fitting curve of their results still closely aligning with the ideal curve, as shown in Fig.~\ref{fig:ablation_exp}b, even though audio and point cloud data were absent from pre-training.
Based on the quantitative results, we find that the performance improves with the increasing number of modalities. It implies that when learning from data with more diversity, AI tends to achieve a comprehensive understanding of the real-world, which aligns with the platonic representation hypothesis~\cite{huh2024platonic}. 

Beyond the scaling effect of modality, it is natural to explore how different modality combinations contribute to varying degrees of gains, referred to as the affinity phenomenon of modality. 
The study began by observing the diverse effectiveness of pre-training modality combinations across tasks, as depicted in Fig.~\ref{fig:ablation_exp}c and Extended Data Fig.~\ref{app_fig:model_scaling}, suggesting complex underlying interactions. 
This phenomenon was further emphasized in Fig.~\ref{fig:ablation_exp}d, where t-SNE visualizations~\cite{van2008visualizing} of feature distributions highlight how varied modality combinations influence the representation learning during pre-training, showcasing their unique affinities and contributions.
During fine-tuning, the impact of this affinity was evident in the convergence behavior, with curves in Fig.~\ref{fig:ablation_exp}e similarly reflecting the varied performance across different modality combinations.
To examine the cause of this affinity, we analyzed the attention weights between tokens from different modalities on the pre-trained \method{}, as shown in Fig.~\ref{fig:ablation_exp}f and Extended Data Fig.~\ref{app_fig:token_affinity}. These tokens exhibited varying degrees of relatedness and interaction, providing insight into the inherent connections~\cite{yu2020gradient,fifty2021efficiently} between modalities.
This comprehensive analysis underscores the importance of selecting pre-training modality combinations to achieve more discriminative feature distributions, thereby significantly enhancing downstream performance.

The scaling effect of modality quantifies universal knowledge accumulation as the cumulative distribution function of a geometric distribution. This effect suggests that \method{} accumulates progressively universal knowledge as it integrates more modalities. Moreover, the affinity phenomenon of modality highlights the importance of thoughtful design in modality combinations to maximize performance benefits. These two findings collectively form a theoretical and practical framework for multi-modal learning, laying the groundwork for AGI.

\subsection*{Discussion}
\method{} unifies Intelligence Islands by generalizing on data, architectures, and learning methods. 
With the unified encoding based on triplet sets and the triplet transformer to handle such triplet sets, \method{} learns from unlabeled, unaligned data across diverse modalities through the parallel reconstruction strategy.
Pre-trained on 296 datasets across five modalities, \method{} accumulates universal knowledge that transcends modality restrictions and constructs the first AI supercontinent. 
Its remarkable generalization is evident across 60 tasks with 15 scientific ones included.
Further analysis reveals a scaling effect of modality, quantifying how universal knowledge accumulates as the number of modalities increases. In summary, \method{} introduces a paradigm shift beyond the limitations of modality-specific AI development.

Despite the promising results \method{} demonstrates, some limitations remain. Similar to modern AI models, \method{} functions as a black box, largely driven by empirical performance rather than robust theoretical foundations. The scaling effect of modality is preliminarily expressed as the cumulative distribution function of a geometric distribution and offers a credible and intuitive scalable framework. However, this scaling effect still lacks rigorous theoretical proof to support its universality and scientific validity. In addition, while we attribute the affinity phenomenon of modality to the relatedness between modalities, its underlying cause requires further investigation.

In the future, \method{} serves as a potential path toward AGI. By leveraging the unified data encoding and scaling effect of modality, it offers broader scalability across data sizes, model parameters, and modality numbers. This enables extensive pre-training to continuously accumulate universal knowledge for the benefit of a wide range of real-world tasks. Furthermore, with the advancements in data encoding, model architecture and pre-training, the AI supercontinent constructed by \method{} will continue to expand at an unprecedented rate. 
As this growth accelerates, a form of intelligence resembling AGI may conceivably emerge in the end.

\section*{References}

\clearpage

\section*{Methods}

\subsection*{Evaluation metrics}
The supervised fine-tuning tasks in Fig.~\ref{fig:main}e can be roughly divided into two categories, \textit{i.e.}, classification and regression tasks. The classification tasks, namely table, vision, graph, point cloud and audio classification, use accuracy (ACC) to measure performance. Notably, the text classification tasks are assessed by the F1 score (for binary classification) and the weighted F1 score (for multi-class classification). The regression tasks, such as table regression and time series forecasting, utilize mean squared error (MSE) for performance measurement. The specific formulas are as follows:
\begin{equation}
    \text{ACC} = \frac{1}{n} \sum_{i=1}^{n} \mathbf{I}(y_i = \hat{y}_i),
\end{equation}
\begin{equation}
\text{F1} = \frac{2 \cdot \sum_{i=1}^{n} \mathbf{I}(y_i = 1 \wedge \hat{y}_i = 1)}{
\sum_{i=1}^{n} \mathbf{I}(y_i = 1) + \sum_{i=1}^{n} \mathbf{I}(\hat{y}_i = 1)},
\end{equation}
\begin{equation}
    \text{MSE} = \frac{1}{n} \sum_{i=1}^{n} (y_i - \hat{y}_i)^2,
\end{equation}
where $n$ is the number of samples in the dataset, $y_i$ is the ground truth, $\hat{y}_i$ is the prediction value, $\mathbf{I}(\cdot)$ is the indicator function where the condition is true, its value is 1, otherwise 0, and $C$ is the number of categories in tasks. Notably, two of graph tasks are class-imbalanced binary classification problems, evaluated using the area under ROC curve (AUC):
\begin{equation}
    \text{AUC} = \frac{1}{N^+*N^-} \sum_{n^+=1}^{N^+} \sum_{n^-=1}^{N^-}\mathbf{I}(\hat{y}_{n^+} > \hat{y}_{n^-}),
\end{equation}
where $N^+$ and $N^-$ are the number of positive and negative samples respectively, $\hat{y}_{n^+}$ is the prediction value of positive samples.

For 15 scientific tasks, we adopt task-specific evaluation metrics. AUC is used for the drug molecule toxicity prediction, drug consumer type classification, and blood-brain barrier permeability prediction. F1 score is used for stock movement prediction. ACC is used for the prostate cancer grading, marine mammal vocalization classification, MMLU, active galactic nuclei classification, high-energy particle identification, mathematical subject classification. Mean absolute error (MAE) is used for cyclic peptide membrane permeability prediction, molecule electronic properties prediction and material band gap prediction. And root mean square error (RMSE) is used for global temperature forecasting and reservoir property estimation tasks:
\begin{equation}
    \text{MAE} = \frac{1}{n} \sum_{i=1}^{n} |y_i - \hat{y}_i|,
\end{equation}
\begin{equation}
    \text{RMSE} = \sqrt{\frac{1}{n} \sum_{i=1}^{n} (y_i - \hat{y}_i)^2}.
\end{equation}

For Fig.~\ref{fig:main}e, we denote the results of \method{}$_\text{w/o}$ as $x_0$. For tasks evaluated using ACC/AUC/F1, the performance percentage of \method{} and competitive models compared with \method{}$_\text{w/o}$  is computed as follows:
\begin{equation}
    \text{Percentage}(x)_{ACC/AUC/F1} = \frac{x-x_0}{x_0} \times 100\% + 100\%,
\end{equation}where $x$ represents the performance of either \method{} or the competitive models. For tasks evaluated using MSE, the performance percentage of of \method{} and competitive models compared with \method{}$_\text{w/o}$ is computed as follows:
\begin{equation}
    \text{Percentage}(x)_{MSE} = \frac{x_0-x}{x_0} \times 100\% + 100\%.
\end{equation} And their performance improvement is computed from the following:
\begin{equation}
    \text{Improvement}(x) = \text{Percentage}(x) - 100\%.
\end{equation} Following the same way, we also report the performance percentage and improvement of \method{} compared with competitive models in the main text.

For Fig.~\ref{fig:ablation_exp}b, we first average the fine-tuning results for audio classification tasks across datasets, and then apply min-max normalization as follows:
\begin{equation}
\text{Norm}(x) = \frac{x - \min(x)}{\max(x) - \min(x)},
\end{equation}
to these averaged values, yielding the normalized results for different modality combinations. Next, we average the normalized scores according to the number of modalities in each combination, resulting in six scatter points corresponding to $[0,1,2,3,4,5]$ modalities. The same procedure is applied to the point cloud classification task. Finally, we average the scatter points from both tasks and visualize the aggregated results in a scatter plot.

Following the similar approach for Fig.~\ref{fig:ablation_exp}a, we average the normalized performance across six types of downstream tasks, namely text classification, table classification, table regression, time series forecasting, video classification, and graph classification, and present the results with a scatter plot. The accompanying radar plots of Fig.~\ref{fig:ablation_exp}a (with the zoomed version presented in Extended Data Fig.~\ref{app_fig:radar_plot}) are the performances of downstream tasks under different modality combinations. For example, in the two-modal case, we plot the performance of each combination with $C(5, 2) = 10$ curves, where each curve contains six values representing the performance across the six types of downstream tasks, and the average performance of modality combinations is shown with a gray color-filled area.

As for Fig.~\ref{fig:ablation_exp}c, we denote the results of \method{}$_\text{w/o}$ as $x_0$. Using the results, which have undergone min-max normalization for eight types of tasks, we then subsequently apply the specific following normalization:
\begin{equation} 
    \text{Norm}(x) = 
    \begin{cases} 
        -\frac{x - x_0}{\min(x) - x_0}  & \text{if } x < x_0, \\
        \frac{x - x_0}{\max(x) - x_0}  & \text{if } x \geq x_0.
    \end{cases}
\end{equation} In this way, the results will be mapped to $[-1,1]$ for better visualization.

In Extended Data Fig.~\ref{app_fig:model_scaling}, fine-tuning results for a specific downstream task on a single dataset are collected across various pre-training modality combinations. Min-max normalization is applied, and the final results are shown using scatter plots.

\subsection*{Details of data preparation}

To unify Intelligence Islands, data from various modalities were widely used during the pre-training phase. 
Specifically, we utilized modalities that are directly accessible by humans or sensors in production and daily life, such as table, time series, vision, text and graph. 
Audio and point cloud were treated as unseen modalities, meaning they were not included in pre-training but were considered in fine-tuning evaluations.

An overview of the datasets that were employed for pre-training is provided in Fig.~\ref{fig:main}b. The figure highlights the vast scale of data, nearly 28 million samples in total, with comprehensive coverage across various domains as shown in the middle ring of Fig.~\ref{fig:main}b, including medicine, nature, transport, web, industry, society, energy, economy, retail, and others. 
The outer ring of Fig.~\ref{fig:main}b represents the number of data samples divided by modality, including table (classification and regression), vision, time series, text and graph. The specific sampling and filtering strategy of pre-training datasets is shown in Extended Data Fig.~\ref{app_fig:data_description}b.
Specifically, the table data were sourced from OpenML~\cite{OpenML2013}, comprising 243 datasets spanning binary classification, multi-class classification, and regression tasks, with a total of 16 million samples. The specific characteristic distributions of table data are shown in Extended Data Fig.~\ref{app_fig:data_description}a.
The text data were drawn from the English Wikipedia corpus introduced in BERT~\cite{devlin-etal-2019-bert}, comprising approximately six million text samples, each consisting of 512 token IDs after tokenization.
The vision data consisted of over one million image samples from ImageNet~\cite{russakovsky2015imagenet}. 
The graph data were derived from the Node Property Prediction category within the Open Graph Benchmark~\cite{hu2020ogb}, consisting of three datasets with nearly one million samples. 
The time series data included open-source datasets from the Monash Time Series Forecasting Repository~\cite{godahewa2021monash} and a selection of time series datasets from Kaggle, covering domains such as air quality and stock markets. In total, there were 48 datasets with 3.47 million samples, where each sample is defined as a sequence of 256 time steps.
The details of all pre-training datasets, including names and sample counts, are summarized in Extended Data Table~\ref{tab:pretrain_data_size}, organized by domains and modalities.

The fine-tuning datasets of Fig.~\ref{fig:main}e are shown in Extended Data Table~\ref{tab:mian_res_info}. Specifically, 24 table datasets are also sourced from OpenML~\cite{OpenML2013}.
Six time series datasets, namely ETT-h1/h2/m1/m2, Electricity and Weather, are from SimMTM~\cite{dong2023simmtm}. The vision datasets include Cifar-10/100~\cite{Krizhevsky09learningmultiple}, and Food-101~\cite{bossard14}. The audio datasets comprise SpeechCommands V2~\cite{speechcommandsv2} and ESC-50~\cite{piczak2015dataset}. Three graph datasets, namely Roman-Empire, Minesweeper, and Tolokers are token from the heterophilous benchmark~\cite{platonovcritical}. 
Three point cloud datasets, namely OBJ-BG, OBJ-ONLY, PB-T50-RS, are selected from the ScanObjectNN benchmark~\cite{uy2019revisiting}. Finally, the text datasets include Emotion Balanced~\cite{saravia2018carer}, BoolQ~\cite{clark2019boolq}, Yelp Polarity~\cite{zhang2015character} and MRPC~\cite{dolan2005paraphrase}.

As for scientific tasks, the drug molecule toxicity prediction task is a primary focus. The Tox21 dataset~\cite{huang2016tox21challenge} from MoleculeNet~\cite{wu2018moleculenet}, which includes only 7,831 compounds, was selected for experiments. 
This dataset was designed to quickly and efficiently assess whether specific chemicals may disrupt processes in the human body that could lead to adverse health effects. 
Its preprocessing follows the Graph Property Prediction task from the Open Graph Benchmark~\cite{hu2020ogb}, such as OgbgMolhiv, where each graph represents a molecule with nodes as atoms and edges as chemical bonds. Consequently, this task is framed as a graph classification problem.

We investigated the BBBP task, utilizing its dataset~\cite{sakiyama2021prediction} of 2,050 compounds from MoleculeNet~\cite{wu2018moleculenet}. This task assesses if compounds cross the blood brain barrier, which is critical for Central Nervous System drug efficacy. In this task, we framed BBBP as a graph classification problem, the same as OgbgMolhiv.

For molecule electronic properties prediction, we studied the QM7 task, utilizing its dataset of 7,160 molecules obtained directly from MoleculeNet~\cite{wu2018moleculenet}. This task involves predicting molecular atomization energy, a fundamental property in quantum chemistry crucial for understanding molecular stability and reactivity. Its preprocessing also followed the Open Graph Benchmark Graph Property Prediction standard, such as OgbgMolhiv. This task is framed as a graph regression problem.

We also investigated the CycPeptMPDB task, utilizing its dataset of 7,451 cyclic peptides~\cite{li2023cycpeptmpdb}. This task focuses on predicting the membrane permeability of cyclic peptides, a crucial property for their development as effective therapeutics and ensuring their bioavailability. In this task, molecules are represented as graphs with atoms as nodes and chemical bonds as edges, following the Open Graph Benchmark Graph Property Prediction standard, such as OgbgMolhiv, and it is accordingly a graph regression problem.

Our work also encompassed the material band gap prediction task, specifically drawn from the JARVIS-DFT database. This involves predicting band gap (mBJ) properties from a dataset of 18,172 samples. Accurate band gap prediction is vital for accelerating the design and discovery of new semiconductor and insulator materials with desired electronic and optoelectronic properties. The preprocessing for this task similarly adhered to the Open Graph Benchmark Graph Property Prediction standard.

We tested \method{} on the global temperature forecasting task. The ERA5 dataset from the European Centre for Medium-Range Weather Forecasts~\cite{bougeault2010thorpex,hersbach2020era5} was selected for the task. This dataset covers global climate information from January 1940 to the present, estimating a wide range of atmospheric, land, and oceanic climate variables hourly. In the experiment, we only used the monthly average data from January 1940 to May 2024, with the target variable being the 2-meter surface temperature (t2m), and a spatial grid with a resolution of $0.25^{\circ} \times 0.25^{\circ}$. Consequently, this task is approached as a standard time series forecasting problem.

We validated \method{}'s performance on the reservoir property estimation task.
Interpreting and processing well logs, which are detailed records of geological formations captured during drilling, to estimate on-site reservoir characteristics (such as petrophysics, geomechanics, and geochemistry) is crucial for reservoir modeling, reserve estimation, and production forecasting. 
The VOLVE dataset used in this task was selected from the 2021 SPWLA PDDA Machine Learning Competition~\cite{fu2024well}. The primary objective is to estimate reservoir characteristics, including shale volume and porosity, based on a set of common well logs, such as depth, gamma ray (GR), bulk density (DEN), neutron porosity (POR), resistivity, and sonic. In the training experiment, the input consists of 10-dimensional features, including depth, and the output corresponds to one of the two reservoir characteristics, making this a table regression task. 

We additionally explored the drug consumption task~\cite{fehrman2017five}. This task, guided by the principles of the Five Factor Model of personality, involves predicting the consumption patterns of various legal and illegal drugs from 12 features encompassing personality and demographic factors. Utilizing a dataset of 1,885 samples, it is critical for public health insights and behavioral science research. Its preprocessing involved standard tabular data techniques for numerical and categorical features. Consequently, this task is framed as a multitask binary classification problem, predicting whether an individual has used each of the 18 different drug types.

Pangaea was also tested on the Higgs boson classification task~\cite{baldi2014searching}. This challenge involves identifying Higgs boson signals amidst vast background noise in the Higgs dataset, which comprises 11 million simulated particle collision events. Each event is characterized by 28 high dimensional tabular features, with a binary output label indicating the presence of a signal or background. This task is paramount in high energy physics for advancing particle discovery and fundamental understanding. Preprocessing involved standard tabular data techniques, such as feature scaling and normalization. Consequently, this task is framed as a binary classification problem on high dimensional tabular data.

We evaluated \method{} on the prostate cancer grading task. The dataset was from the Automated Gleason Grading Challenge 2022 (AGGC) for prostate adenocarcinoma tissue classification~\cite{huo2024comprehensive}, which consists of H\&E-stained whole slide image (WSI) of prostatectomy and biopsy specimens, annotated by experienced pathologists. For this downstream task, we only selected subset 1 (whole mount images scanned by an Akoya Biosciences scanner) of the dataset. Consequently, this task is viewed as a typical vision classification problem.

We also addressed the involves classifying the active galactic nuclei task. This task involves a dataset Mirabest~\cite{porter2023mirabest} of 1,222 radio images. Morphological classification of radio galaxies is crucial for understanding galaxy evolution and active galactic nuclei in the universe. Its preprocessing involved standard image processing techniques, such as resizing and normalization. Consequently, this task is also framed as a vision classification problem.

For marine mammal vocalization classification, we utilized the ``best of'' cuts section of the Watkins Marine Mammal Sound Database (WMMSD)~\cite{sayigh2016watkins}. This dataset comprises 1,694 high-quality, low-noise sound cuts from 32 different marine mammal species. After removing multi-label entries, our final dataset consisted of 1,290 cuts across 32 distinct categories. These sound cuts were then effectively transformed into Mel spectrograms~\cite{gong2021ast}, allowing us to approach this as a typical vision classification problem.

The evaluation of Pangaea also extended to the Massive Multitask Language Understanding (MMLU) task~\cite{hendrycks2021measuring}. This task evaluates a model's broad knowledge and complex reasoning ability across 57 highly diverse subjects, utilizing a comprehensive set of multiple choice questions. This task serves as a crucial benchmark for evaluating model capabilities in general intelligence and understanding complex reasoning across diverse domains. Its preprocessing involved standard natural language processing techniques, such as tokenization and formatting for multiple choice questions. Consequently, this task is framed as a multiclass text classification problem requiring selection from multiple choice answers.

We also applied Pangaea to the Mathematics Subject Classification (MSC)~\cite{dunne2020mathematics} task. This task involves classifying 164,230 mathematical articles into specific subject areas according to the hierarchical MSC scheme. It is crucial for organizing the vast body of mathematical literature and enabling efficient knowledge retrieval. Its preprocessing involved standard natural language processing techniques for text data, such as tokenization and feature extraction. Consequently, this task is framed as a multiclass text classification problem.

We adopted the StockNet dataset~\cite{xu2018stock} for stock movement prediction. This dataset features high-trade-volume S\&P 500 stocks from the NYSE and NASDAQ markets. Stock-specific tweets were extracted using regex queries based on NASDAQ ticker symbols (\textit{e.g.}, \$AMZN), while price data was sourced from Yahoo Finance. Samples were generated by applying a 5-day lag window along trading days. These samples were labeled based on the closing price's movement percentage: positive for movements $\geq0.55\%$ and negative for movements $\le-0.5\%$. This yielded 26,614 samples, with an approximate 49.78\% positive and 50.22\% negative class distribution. Consequently, this task is viewed as a binary classification problem with the text and time series inputs.

\subsection*{Details of \method{} Architecture}
We designed a custom-built \method{} with approximately 20 million parameters. 
The triplet transformer of \method{} comprises a triplet tokenizer and a bi-directional attention module. The triplet tokenizer includes a linear layer with dimensions $768 \times 512$ and a topology embedding layer with dimensions $1000 \times 512$. 
The attention module consists of eight transformer blocks, each block having a hidden size of $256$, an intermediate size of $512$ and eight attention heads, achieving a balance between computational efficiency and model complexity. 
The attention mask is bi-directional, enhancing the model's capability to contextualize and interpret inputs across both directions.
Rotary positional embedding~\cite{su2024RoFormer} was used as the global topology embedding layer for the global topology part in each triplet, enabling the model to discern the triplet token order.
The implementation of these components draws inspiration from the LLaMA~\cite{touvron2023llama} model.
For pre-training, an additional multi-layer perceptron (MLP) head was appended to the triplet transformer to reconstruct the raw data. During fine-tuning, the MLP was repurposed to predict task-specific values.

\subsection*{Details of triplet construction}
The unified data encoding of Pangaea is that any data modality can be mathematically abstracted as a weighted graph, which is then decomposed into a set of triplets. To provide a modality-agnostic overview of the triplet construction process, we detail how raw data is systematically transformed into a triplet set. For a given data sample $X_i$, the $j$-th triplet $t_{i,j}$ consists of two numeric parts, $t_{i,j}^{\text{num}_1}$ and $t_{i,j}^{\text{num}_2}$, along with a local topology part $I_{i,j}^{\text{indices}}$, and a global topology part represented by the triplet index $j$. The numeric parts (``node'' in the weighted graph) are distinct yet related numerical subsets extracted directly from the raw data. For instance, in table, they are two randomly sampled subvectors; in images, they are two adjacent patches; in text, they are text token IDs from two consecutive words; for point clouds, they are two local point groups derived from farthest point sampling and k-nearest neighbors. These raw parts are then padded to a consistent dimension before being concatenated and mapped into a shared numerical embedding space. The topology part is vital for providing context, essentially defining the relationship between the two numeric parts and their position within the data sample. We distill this into two parts. Local topology (``edge'' in the graph) captures the immediate connection, typically represented by the original indices or unique identifiers of the raw parts within the data sample. For modalities like images, time series, and text, this means the 1D sequential indices of the paired patches or tokens. These indices are projected into a shared topology embedding space and then averaged to obtain a consistent dimension (``weight'' of edge). Beyond local relationships, we also incorporate global topology, which specifies the absolute position of each triplet within the sequence of all triplets generated from a single data sample. This part is mapped by a shared position embedding layer, allowing Pangaea to understand the overall structure of the set of triplets. The detailed setting of triplets for each modality is listed in Extended Data Table~\ref{tab:triplet_info}.

By consistently applying this construction strategy, Pangaea defines triplets with explicit numeric and topology parts, embeds each part independently into a shared representation space, and integrates them into triplet tokens. This approach enables Pangaea to process and align data from a wide range of modalities within an unified framework. We provide below the specific formulations used to generate triplets across all seven incorporated modalities.

\subsubsection*{Triplet construction process of table modality}
For a table sample $X_i \in \mathbb{R}^d$, we first construct a set of $d$ raw triplets, $\mathcal{T}_i$. The $j$-th triplet $t_{i,j}$ includes two numerical sub-vectors from $X_i$ (its numerical parts,, $t_{i,j}^{\text{num}_1}, t_{i,j}^{\text{num}_2}$), their original feature indices $I_{i,j}^{\text{indices}}$ to capture local topology, and an absolute position $j$ for global topological context among the $d$ triplets. These raw triplets are then transformed into their corresponding triplet tokens, $t_{i,j}^{\text{token}}$, through a series of padding, linear mapping, and embedding steps, combining numerical, local topological, and global topological information. The precise construction process, including all dimensions and transformations, is outlined as below.

\vspace{-1em}
\begin{nolinenumbers}
{\fontsize{9pt}{11pt}
\begin{flalign*}
	& \mathcal{D}_{\text{table}} \in \mathbb{R}^{N \times d} & \text{(Table dataset, } N \text{ samples, } d \text{ features)} \\
	& X_i \in \mathbb{R}^{d}, \quad X_i \sim \mathcal{D}_{\text{table}} & \text{(Sample from table dataset)} \\
	& \text{For each } j \in \{1, \dots, d\} \text{ to form triplet } t_{i,j}: & \\
	& \quad t_{i,j}^{\text{num}_1}, t_{i,j}^{\text{num}_2} \subset X_i,\quad t_{i,j}^{\text{num}_1}, t_{i,j}^{\text{num}_2} \in \mathbb{R}^{d/2} & \text{(Random sub-vectors from } X_i\text{)} \\
	& \quad I_{i,j}^{\text{indices}} \in \{1, \dots, d\}^{d} \in \mathbb{Z}^{d} & \text{(Original feature indices of } t_{i,j}^{\text{num}_1}, t_{i,j}^{\text{num}_2}\text{)} \\
	& \quad t_{i,j} = (t_{i,j}^{\text{num}_1}, t_{i,j}^{\text{num}_2}, I_{i,j}^{\text{indices}}, j) & \text{(The } j \text{-th raw triplet)} \\
	& \mathcal{T}_i = \{ t_{i,1}, t_{i,2}, ..., t_{i,d} \} & \text{(Set of } d \text{ raw triplets for } X_i\text{)}\\
	& \text{For each } t_{i,j} \text{ to form triplet token } t_{i,j}^{\text{token}}: & \\
	& \quad t_{i,j}^{\text{num}_1'} = \text{ZeroPad}(t_{i,j}^{\text{num}_1}) \in \mathbb{R}^{384} & \\
	& \quad t_{i,j}^{\text{num}_2'} = \text{ZeroPad}(t_{i,j}^{\text{num}_2}) \in \mathbb{R}^{384} & \\
	& \quad t_{i,j}^{\text{num}_{\text{concat}}} = \text{Concat}(t_{i,j}^{\text{num}_1'}, t_{i,j}^{\text{num}_2'}) \in \mathbb{R}^{768} & \\
	& \quad t_{i,j}^{\text{num}} = \text{Linear}(t_{i,j}^{\text{num}_{\text{concat}}}) \in \mathbb{R}^{512} & \text{(Numerical embedding)} \\
	& \quad t_{i,j}^{\text{topo}^{\text{local}}_{\text{emb}}} = \text{TopologyEmbedding}(I_{i,j}^{\text{indices}}) \in \mathbb{R}^{d \times 512} & \\
	& \quad t_{i,j}^{\text{topo}^{\text{local}}} = \text{MeanReduce}(t_{i,j}^{\text{topo}^{\text{local}}_{\text{emb}}}) \in \mathbb{R}^{512} & \text{(Local topology embedding)} \\
	& \quad t_{i,j}^{\text{topo}^{\text{global}}} = \text{PositionEmbedding}(j) \in \mathbb{R}^{512} & \text{(Global topology embedding)} \\
	& \quad t_{i,j}^{\text{token}} = t_{i,j}^{\text{num}} + t_{i,j}^{\text{topo}^{\text{local}}} + t_{i,j}^{\text{topo}^{\text{global}}} \in \mathbb{R}^{512} & \text{(Final triplet token)} \\
	& \mathcal{T}_i^{\text{token}} = \{ t_{i,1}^{\text{token}}, t_{i,2}^{\text{token}}, ..., t_{i,d}^{\text{token}} \} & \text{(Set of } d \text{ triplet tokens for } X_i\text{)}
\end{flalign*}}
\end{nolinenumbers}
\vspace{0em}

\subsubsection*{Triplet construction process of time series modality}
For a time series sample $X_i \in \mathbb{R}^{256}$, we first construct a set of $8$ raw triplets, $\mathcal{T}_i$. The $j$-th triplet $t_{i,j}$ includes two sequential time segments from $X_i$ (its numerical parts, $t_{i,j}^{\text{num}_1}, t_{i,j}^{\text{num}_2}$), their original temporal indices $I_{i,j}^{\text{indices}}$ to capture local topology, and an absolute position $j$ for global topological context among the $8$ triplets. These raw triplets are then transformed into their corresponding triplet tokens, $t_{i,j}^{\text{token}}$, through a series of padding, linear mapping, and embedding steps, combining numerical, local topological, and global topological information. The precise construction process, including all dimensions and transformations, is further  outlined in the equations below.

\vspace{-1em}
\begin{nolinenumbers}
{\fontsize{9pt}{11pt}
\begin{flalign*}
	& \mathcal{D}_{\text{ts}} \in \mathbb{R}^{N \times T}, \quad T = 256 \text{ time steps} & \text{(Time series dataset, } N \text{ segments)} \\
	& X_i \in \mathbb{R}^{T}, \quad X_i \sim \mathcal{D}_{\text{ts}} & \text{(Sampled time series segment)} \\
	& \text{For each } j \in \{1, \dots, 8\} \text{ to form triplet } t_{i,j}: & \\
	& \quad \text{offset}_j = 32(j-1) & \\
	& \quad t_{i,j}^{\text{num}_1} = X_i[\text{offset}_j : \text{offset}_j + 15] \in \mathbb{R}^{16} & \text{(1st numeric segment)} \\
	& \quad t_{i,j}^{\text{num}_2} = X_i[\text{offset}_j + 16 : \text{offset}_j + 31] \in \mathbb{R}^{16} & \text{(2nd numeric segment)} \\
	& \quad I_{i,j}^{\text{indices}} = \{\text{offset}_j, \dots, \text{offset}_j + 31\} \in \mathbb{Z}^{32} & \text{(Original temporal indices of segments)} \\
	& \quad t_{i,j} = (t_{i,j}^{\text{num}_1}, t_{i,j}^{\text{num}_2}, I_{i,j}^{\text{indices}}, j) & \text{(The } j \text{-th raw triplet)} \\
	& \mathcal{T}_i = \{ t_{i,1}, t_{i,2}, ..., t_{i,8} \} & \text{(Set of } 8 \text{ raw triplets for } X_i\text{)}\\
	& \text{For each } t_{i,j} \text{ to form triplet token } t_{i,j}^{\text{token}}: & \\
	& \quad t_{i,j}^{\text{num}_1'} = \text{ZeroPad}(t_{i,j}^{\text{num}_1}) \in \mathbb{R}^{384} & \\
	& \quad t_{i,j}^{\text{num}_2'} = \text{ZeroPad}(t_{i,j}^{\text{num}_2}) \in \mathbb{R}^{384} & \\
	& \quad t_{i,j}^{\text{num}_{\text{concat}}} = \text{Concat}(t_{i,j}^{\text{num}_1'}, t_{i,j}^{\text{num}_2'}) \in \mathbb{R}^{768} & \\
	& \quad t_{i,j}^{\text{num}} = \text{Linear}(t_{i,j}^{\text{num}_{\text{concat}}}) \in \mathbb{R}^{512} & \text{(Numerical embedding)} \\
	& \quad t_{i,j}^{\text{topo}^{\text{local}}_{\text{emb}}} = \text{TopologyEmbedding}(I_{i,j}^{\text{indices}}) \in \mathbb{R}^{32 \times 512} & \\
	& \quad t_{i,j}^{\text{topo}^{\text{local}}} = \text{MeanReduce}(t_{i,j}^{\text{topo}^{\text{local}}_{\text{emb}}}) \in \mathbb{R}^{512} & \text{(Local topology embedding)} \\
	& \quad t_{i,j}^{\text{topo}^{\text{global}}} = \text{PositionEmbedding}(j) \in \mathbb{R}^{512} & \text{(Global topology embedding)} \\
	& \quad t_{i,j}^{\text{token}} = t_{i,j}^{\text{num}} + t_{i,j}^{\text{topo}^{\text{local}}} + t_{i,j}^{\text{topo}^{\text{global}}} \in \mathbb{R}^{512} & \text{(Final triplet token)} \\
	& \mathcal{T}_i^{\text{token}} = \{ t_{i,1}^{\text{token}}, t_{i,2}^{\text{token}}, ..., t_{i,8}^{\text{token}} \} & \text{(Set of } 8 \text{ triplet tokens for } X_i\text{)}
\end{flalign*}}
\end{nolinenumbers}
\vspace{0em}

\subsubsection*{Triplet construction process of image modality}
For an image sample $X_i \in \mathbb{R}^{224\times 224\times 3}$, we first construct a set of $196$ raw triplets, $\mathcal{T}_i$. The $j$-th triplet $t_{i,j}$ includes two adjacent image patches from $X_i$ (its numerical parts, $t_{i,j}^{\text{num}_1}, t_{i,j}^{\text{num}_2}$), their original patch indices $I_{i,j}^{\text{indices}}$ to capture local topology, and an absolute position $j$ for global topological context among the $196$ triplets. These raw triplets are then comprehensively transformed into their corresponding triplet tokens, $t_{i,j}^{\text{token}}$, through a series of flattening, linear mapping, and embedding steps, combining numerical, local topological, and global topological information. The precise construction process, including all dimensions and transformations, is outlined in the equations below.

\vspace{-1em}
\begin{nolinenumbers}
{\fontsize{9pt}{11pt}
\begin{flalign*}
	& \mathcal{D}_{\text{img}} \in \mathbb{R}^{N \times H \times W \times C}, \quad H=W=224, C=3 & \text{(Image dataset, } N \text{ images)} \\
	& X_i \in \mathbb{R}^{H \times W \times C}, \quad X_i \sim \mathcal{D}_{\text{img}} & \text{(Sampled image)} \\
	& \text{Image split into } 14 \times 28 \text{ patches (size } 16 \times 8 \times 3 \text{ each).} & \\
	& \text{This yields } 392 \text{ patches (indexed } 1 \text{ to } 392 \text{ in row-major order).} & \\
	& \text{For each } j \in \{1, \dots, 196\} \text{ to form triplet } t_{i,j}: & \\
	& \quad P_{i, \text{left\_idx}}, P_{i, \text{right\_idx}} \in \mathbb{R}^{16 \times 8 \times 3} & \text{(Adjacent image patches)} \\
	& \quad t_{i,j}^{\text{num}_1} = P_{i, \text{left\_idx}} & \\
	& \quad t_{i,j}^{\text{num}_2} = P_{i, \text{right\_idx}} & \\
	& \quad I_{i,j}^{\text{indices}} = \{\text{left\_idx}, \text{right\_idx}\} \in \mathbb{Z}^{2} & \text{(Original 1D indices of patches)} \\
	& \quad t_{i,j} = (t_{i,j}^{\text{num}_1}, t_{i,j}^{\text{num}_2}, I_{i,j}^{\text{indices}}, j) & \text{(The } j \text{-th raw triplet)} \\
	& \mathcal{T}_i = \{ t_{i,1}, t_{i,2}, ..., t_{i,196} \} & \text{(Set of } 196 \text{ raw triplets for } X_i\text{)}\\
	& \text{For each } t_{i,j} \text{ to form triplet token } t_{i,j}^{\text{token}}: & \\
	& \quad t_{i,j}^{\text{num}_{\text{concat}}} = \text{Concat}(t_{i,j}^{\text{num}_1}, t_{i,j}^{\text{num}_2}) \in \mathbb{R}^{16 \times 16 \times 3} & \\
	& \quad t_{i,j}^{\text{num}_{\text{flat}}} = \text{Flatten}(t_{i,j}^{\text{num}_{\text{concat}}}) \in \mathbb{R}^{768} & \\
	& \quad t_{i,j}^{\text{num}} = \text{Linear}(t_{i,j}^{\text{num}_{\text{flat}}}) \in \mathbb{R}^{512} & \text{(Numerical embedding)} \\
	& \quad t_{i,j}^{\text{topo}^{\text{local}}_{\text{emb}}} = \text{TopologyEmbedding}(I_{i,j}^{\text{indices}}) \in \mathbb{R}^{2 \times 512} & \\
	& \quad t_{i,j}^{\text{topo}^{\text{local}}} = \text{MeanReduce}(t_{i,j}^{\text{topo}^{\text{local}}_{\text{emb}}}) \in \mathbb{R}^{512} & \text{(Local topology embedding)} \\
	& \quad t_{i,j}^{\text{topo}^{\text{global}}} = \text{PositionEmbedding}(j) \in \mathbb{R}^{512} & \text{(Global topology embedding)} \\
	& \quad t_{i,j}^{\text{token}} = t_{i,j}^{\text{num}} + t_{i,j}^{\text{topo}^{\text{local}}} + t_{i,j}^{\text{topo}^{\text{global}}} \in \mathbb{R}^{512} & \text{(Final triplet token)} \\
	& \mathcal{T}_i^{\text{token}} = \{ t_{i,1}^{\text{token}}, t_{i,2}^{\text{token}}, ..., t_{i,196}^{\text{token}} \} & \text{(Set of } 196 \text{ triplet tokens for } X_i\text{)}
\end{flalign*}}
\end{nolinenumbers}
\vspace{0em}

\subsubsection*{Triplet construction process of audio modality}
For an audio sample $X_i$, we use Mel spectrograms~\cite{gong2021ast} to convert it into an image format with dimensions of $\mathbb{R}^{512\times 128\times 3}$. For this audio sample, we first construct a set of $256$ raw triplets, $\mathcal{T}_i$. The $j$-th triplet $t_{i,j}$ includes two adjacent spectrogram patches from $X_i$ (its numerical parts, $t_{i,j}^{\text{num}_1}, t_{i,j}^{\text{num}_2}$), their original patch indices $I_{i,j}^{\text{indices}}$ to capture local topology, and an absolute position $j$ for global topological context among the $256$ triplets. These raw triplets are then transformed into their corresponding triplet tokens, $t_{i,j}^{\text{token}}$, through a series of flattening, linear mapping, and embedding steps, combining numerical, local topological, and global topological information. The precise construction process, including all dimensions and transformations, is outlined in the equations below.

\vspace{-1em}
\begin{nolinenumbers}
{\fontsize{9pt}{11pt}
\begin{flalign*}
	& \mathcal{D}_{\text{audio}} \in \mathbb{R}^{N \times H \times W \times C}, \quad H=512, W=128, C=3 & \text{(Audio dataset, } N \text{ spectrograms)}  \\
	& \quad \text{where raw audio is converted to Mel spectrogram}, & \\
	& \quad \text{and its single channel is repeated to 3}. & \\
	& X_i \in \mathbb{R}^{H \times W \times C}, \quad X_i \sim \mathcal{D}_{\text{audio}} & \text{(Sampled Mel spectrogram)} \\
	& \text{Spectrogram split into } 32 \times 16 \text{ patches (size } 16 \times 8 \times 3 \text{ each).} & \\
	& \text{This yields } 512 \text{ patches (indexed } 1 \text{ to } 512 \text{ in row-major order).} & \\
	& \text{For each } j \in \{1, \dots, 256\} \text{ to form triplet } t_{i,j}: & \\
	& \quad P_{i, \text{left\_idx}}, P_{i, \text{right\_idx}} \in \mathbb{R}^{16 \times 8 \times 3} & \text{(Adjacent patches)} \\
	& \quad t_{i,j}^{\text{num}_1} = P_{i, \text{left\_idx}} & \\
	& \quad t_{i,j}^{\text{num}_2} = P_{i, \text{right\_idx}} & \\
	& \quad I_{i,j}^{\text{indices}} = \{\text{left\_idx}, \text{right\_idx}\} \in \mathbb{Z}^{2} & \text{(Original 1D indices of patches)} \\
	& \quad t_{i,j} = (t_{i,j}^{\text{num}_1}, t_{i,j}^{\text{num}_2}, I_{i,j}^{\text{indices}}, j) & \text{(The } j \text{-th raw triplet)} \\
	& \mathcal{T}_i = \{ t_{i,1}, t_{i,2}, ..., t_{i,256} \} & \text{(Set of } 256 \text{ raw triplets for } X_i\text{)}\\
	& \text{For each } t_{i,j} \text{ to form triplet token } t_{i,j}^{\text{token}}: & \\
	& \quad t_{i,j}^{\text{num}_{\text{concat}}} = \text{Concat}(t_{i,j}^{\text{num}_1}, t_{i,j}^{\text{num}_2}) \in \mathbb{R}^{16 \times 16 \times 3} & \\
	& \quad t_{i,j}^{\text{num}_{\text{flat}}} = \text{Flatten}(t_{i,j}^{\text{num}_{\text{concat}}}) \in \mathbb{R}^{768} & \\
	& \quad t_{i,j}^{\text{num}} = \text{Linear}(t_{i,j}^{\text{num}_{\text{flat}}}) \in \mathbb{R}^{512} & \text{(Numerical embedding)} \\
	& \quad t_{i,j}^{\text{topo}^{\text{local}}_{\text{emb}}} = \text{TopologyEmbedding}(I_{i,j}^{\text{indices}}) \in \mathbb{R}^{2 \times 512} & \\
	& \quad t_{i,j}^{\text{topo}^{\text{local}}} = \text{MeanReduce}(t_{i,j}^{\text{topo}^{\text{local}}_{\text{emb}}}) \in \mathbb{R}^{512} & \text{(Local topology embedding)} \\
	& \quad t_{i,j}^{\text{topo}^{\text{global}}} = \text{PositionEmbedding}(j) \in \mathbb{R}^{512} & \text{(Global topology embedding)} \\
	& \quad t_{i,j}^{\text{token}} = t_{i,j}^{\text{num}} + t_{i,j}^{\text{topo}^{\text{local}}} + t_{i,j}^{\text{topo}^{\text{global}}} \in \mathbb{R}^{512} & \text{(Final triplet token)} \\
	& \mathcal{T}_i^{\text{token}} = \{ t_{i,1}^{\text{token}}, t_{i,2}^{\text{token}}, ..., t_{i,256}^{\text{token}} \} & \text{(Set of } 256 \text{ triplet tokens for } X_i\text{)}
\end{flalign*}}
\end{nolinenumbers}
\vspace{0em}

\subsubsection*{Triplet construction process of graph modality}
For an anchor node $X_i \in\mathbb{R}^{d}$ in a graph, we first construct a set of $32$ raw triplets, $\mathcal{T}_i$. The $j$-th triplet $t_{i,j}$ comprises the anchor node $X_i$ and one of its neighbors (its numerical parts, $t_{i,j}^{\text{num}_1}, t_{i,j}^{\text{num}_2}$), their original feature indices $I_{i,j}^{\text{indices}}$ to capture local topology, and an absolute position $j$ for global topological context among the $32$ triplets. These raw triplets are then transformed into their corresponding triplet tokens, $t_{i,j}^{\text{token}}$, through a series of padding, linear mapping, and embedding steps, combining numerical, local topological, and global topological information. The precise construction process, including all dimensions and transformations, is outlined in the equations below.

\vspace{-1em}
\begin{nolinenumbers}
{\fontsize{9pt}{11pt}
\begin{flalign*}
	& \mathcal{G} = (\mathcal{V}, \mathcal{E}), \quad \mathcal{V} \in \mathbb{R}^{N \times d}, \quad d < 384 & \text{(Graph dataset, } N \text{ nodes with } d \text{ features)} \\
	& v_i \in \mathcal{V}, \quad X_i \in \mathbb{R}^{d} & \text{(Sampled anchor node } v_i \text{ and its features } X_i\text{)} \\
	& \mathcal{N}(v_i) = \{ v_{i,k} \mid k = 1, \dots, 32 \} \subseteq \mathcal{V} & \text{(Sampled 32 neighbors of } v_i\text{)} \\
	& \text{For each } j \in \{1, \dots, 32\} \text{ to form triplet } t_{i,j}: & \\
	& \quad v_{i,j} \in \mathcal{N}(v_i), \quad X_{i,j} \in \mathbb{R}^{d} & \text{(The } j \text{-th  neighbor } v_{i,j} \text{ and its features } X_{i,j}\text{)} \\
	& \quad t_{i,j}^{\text{num}_1} = X_i \in \mathbb{R}^{d} & \text{(Features of the anchor node)} \\
	& \quad t_{i,j}^{\text{num}_2} = X_{i,j} \in \mathbb{R}^{d} & \text{(Features of the } j \text{-th neighbor)} \\
	& \quad I_{i,j}^{\text{indices}} \in \{1, \dots, d\}^{d} \in \mathbb{Z}^{d} & \text{(Original feature indices of } X_i\text{)} \\
	& \quad t_{i,j} = (t_{i,j}^{\text{num}_1}, t_{i,j}^{\text{num}_2}, I_{i,j}^{\text{local}}, j) & \text{(The } j \text{-th raw triplet)} \\
	& \mathcal{T}_i = \{ t_{i,1}, t_{i,2}, ..., t_{i,32} \} & \text{(Set of } 32 \text{ raw triplets for } X_i\text{)}\\
	& \text{For each } t_{i,j} \text{ to form triplet token } t_{i,j}^{\text{token}}: & \\
	& \quad t_{i,j}^{\text{num}_1'} = \text{ZeroPad}(t_{i,j}^{\text{num}_1}) \in \mathbb{R}^{384} & \\
	& \quad t_{i,j}^{\text{num}_2'} = \text{ZeroPad}(t_{i,j}^{\text{num}_2}) \in \mathbb{R}^{384} & \\
	& \quad t_{i,j}^{\text{num}_{\text{concat}}} = \text{Concat}(t_{i,j}^{\text{num}_1'}, t_{i,j}^{\text{num}_2'}) \in \mathbb{R}^{768} & \\
	& \quad t_{i,j}^{\text{num}} = \text{Linear}(t_{i,j}^{\text{num}_{\text{concat}}}) \in \mathbb{R}^{512} & \text{(Numerical embedding)} \\
	& \quad t_{i,j}^{\text{topo}^{\text{local}}_{\text{emb}}} = \text{TopologyEmbedding}(I_{i,j}^{\text{indices}}) \in \mathbb{R}^{d \times 512} &  \\
	& \quad t_{i,j}^{\text{topo}^{\text{local}}} = \text{MeanReduce}(t_{i,j}^{\text{topo}^{\text{local}}_{\text{emb}}}) \in \mathbb{R}^{512} & \text{(Local topology embedding)} \\
	& \quad t_{i,j}^{\text{topo}^{\text{global}}} = \text{PositionEmbedding}(j) \in \mathbb{R}^{512} & \text{(Global topology embedding)} \\
	& \quad t_{i,j}^{\text{token}} = t_{i,j}^{\text{num}} + t_{i,j}^{\text{topo}^{\text{local}}} + t_{i,j}^{\text{topo}^{\text{global}}} \in \mathbb{R}^{512} & \text{(Final triplet token)} \\
	& \mathcal{T}_i^{\text{token}} = \{ t_{i,1}^{\text{token}}, t_{i,2}^{\text{token}}, ..., t_{i,32}^{\text{token}} \} & \text{(Set of } 32 \text{ triplet tokens for } X_i\text{)}
\end{flalign*}}
\end{nolinenumbers}
\vspace{0em}

\subsubsection*{Triplet construction process of text modality}
After adopting a random sentence chunking strategy and a pre-trained BERT text tokenizer~\cite{devlin-etal-2019-bert}, a sampled text segment $X_i$ consists of 512 text token IDs, \textit{i.e.}, $X_i \in \mathbb{Z}^{S}$.
Based on this text sample, we first construct a set of $256$ raw triplets, $\mathcal{T}_i$. The $j$-th triplet $t_{i,j}$ comprises two sequential text token IDs (its numerical parts, $t_{i,j}^{\text{num}_1}, t_{i,j}^{\text{num}_2}$), their original sequential indices $I_{i,j}^{\text{indices}}$ to capture local topology, and an absolute position $j$ for global topological context among the $256$ triplets. These raw triplets are then transformed into their corresponding triplet tokens, $t_{i,j}^{\text{token}}$, through a series of word embedding~\cite{devlin-etal-2019-bert}, padding, linear mapping, and embedding steps, combining numerical, local topological, and global topological information. The precise construction process, including all dimensions and transformations, is outlined as below.

\vspace{-1em}
\begin{nolinenumbers}
{\fontsize{9pt}{11pt}
\begin{flalign*}
	& \mathcal{D}_{\text{text}} \in \mathbb{R}^{N\times S}, \quad S = 512 & \text{(Text dataset, } N \text{ segments, } S \text{ text token IDs)} \\
	& X_i \in \mathbb{Z}^{S}, \quad X_i \sim \mathcal{D}_{\text{text}} & \text{(Sampled text segment, sequence of token IDs)} \\
	& \text{For each } j \in \{1, \dots, 256\} \text{ to form triplet } t_{i,j}: & \\
	& \quad t_{i,j}^{\text{num}_1} = X_i[2j-2] \in \mathbb{Z} & \text{(Raw token ID 1)} \\
	& \quad t_{i,j}^{\text{num}_2} = X_i[2j-1] \in \mathbb{Z} & \text{(Raw token ID 2)} \\
	& \quad I_{i,j}^{\text{indices}} = \{2j-2, 2j-1\} \in \mathbb{Z}^{2} & \text{(Original sequential indices of token IDs)} \\
	& \quad t_{i,j} = (t_{i,j}^{\text{num}_1}, t_{i,j}^{\text{num}_2}, I_{i,j}^{\text{indices}}, j) & \text{(The } j \text{-th raw triplet)} \\
	& \mathcal{T}_i = \{ t_{i,1}, t_{i,2}, ..., t_{i,256} \} & \text{(Set of } 256 \text{ raw triplets for } X_i\text{)}\\
	& \text{For each } t_{i,j} \text{ to form triplet token } t_{i,j}^{\text{token}}: & \\
	& \quad t_{i,j}^{\text{num}_1'} = \text{WordEmbedding}(t_{i,j}^{\text{num}_1}) \in \mathbb{R}^{256} & \text{(Token ID embedding, refer to BERT)} \\
	& \quad t_{i,j}^{\text{num}_2'} = \text{WordEmbedding}(t_{i,j}^{\text{num}_2}) \in \mathbb{R}^{256} & \text{(Token ID embedding, refer to BERT)} \\
	& \quad t_{i,j}^{\text{num}_1''} = \text{ZeroPad}(t_{i,j}^{\text{num}_1'}) \in \mathbb{R}^{384} & \\
	& \quad t_{i,j}^{\text{num}_2''} = \text{ZeroPad}(t_{i,j}^{\text{num}_2'}) \in \mathbb{R}^{384} & \\
	& \quad t_{i,j}^{\text{num}_{\text{concat}}} = \text{Concat}(t_{i,j}^{\text{num}_1''}, t_{i,j}^{\text{num}_2''}) \in \mathbb{R}^{768} & \\
	& \quad t_{i,j}^{\text{num}} = \text{Linear}(t_{i,j}^{\text{num}_{\text{concat}}}) \in \mathbb{R}^{512} & \text{(Numerical embedding)} \\
	& \quad t_{i,j}^{\text{topo}^{\text{local}}_{\text{emb}}} = \text{TopologyEmbedding}(I_{i,j}^{\text{indices}}) \in \mathbb{R}^{2 \times 512} & \\
	& \quad t_{i,j}^{\text{topo}^{\text{local}}} = \text{MeanReduce}(t_{i,j}^{\text{topo}^{\text{local}}_{\text{emb}}}) \in \mathbb{R}^{512} & \text{(Local topology embedding)} \\
	& \quad t_{i,j}^{\text{topo}^{\text{global}}} = \text{PositionEmbedding}(j) \in \mathbb{R}^{512} & \text{(Global topology embedding)} \\
	& \quad t_{i,j}^{\text{token}} = t_{i,j}^{\text{num}} + t_{i,j}^{\text{topo}^{\text{local}}} + t_{i,j}^{\text{topo}^{\text{global}}} \in \mathbb{R}^{512} & \text{(Final triplet token)} \\
	& \mathcal{T}_i^{\text{token}} = \{ t_{i,1}^{\text{token}}, t_{i,2}^{\text{token}}, ..., t_{i,256}^{\text{token}} \} & \text{(Set of } 256 \text{ triplet tokens for } X_i\text{)}
\end{flalign*}}
\end{nolinenumbers}
\vspace{0em}

\subsubsection*{Triplet construction process of point cloud modality}
After adopting the pre-processing method (\textit{i.e.}, farthest point sampling) used in PointGPT~\cite{chen2023pointgpt}, a point cloud with $S$ raw 3D points is converted into $g$ center points, each having $k$ nearest neighbors represented by their 3D coordinates (x, y, z).
Based on this point cloud sample $X_i^{\text{groups}} \in \mathbb{R}^{g \times k \times 3}$, we first construct a set of $g/2$ raw triplets, $\mathcal{T}_i$. The $j$-th triplet $t_{i,j}$ comprises two point groups (its numerical parts, $t_{i,j}^{\text{num}_1}, t_{i,j}^{\text{num}_2}$), their original indices $I_{i,j}^{\text{indices}}$ to capture local topology, and an absolute position $j$ for global topological context among the $g/2$ triplets. These raw triplets are then transformed into their corresponding triplet tokens, $t_{i,j}^{\text{token}}$, through a series of point embedding~\cite{chen2023pointgpt}, padding, linear mapping, and embedding steps, combining numerical, local topological, and global topological information. The precise construction process, including all dimensions and transformations, is outlined in the equations below.

\vspace{-1em}
\begin{nolinenumbers}
{\fontsize{9pt}{11pt}
\begin{flalign*}
	& \mathcal{D}_{\text{point cloud}} \in \mathbb{R}^{N\times S\times 3} & \text{(Point cloud dataset, } N \text{ 3D models, } S \text{ points)} \\
	& X_i \in \mathbb{R}^{S\times 3}, \quad X_i \sim \mathcal{D}_{\text{point cloud}} & \text{(Sampled 3D model)} \\
	& \text{Pre-processing for } X_i: && \\
	& \quad \text{Farthest Point Sampling to select } g \text{ center points.} & \\
	& \quad \text{For each center, gather } k \text{ nearest neighbors.} & \\
	& X_i^{\text{groups}} \in \mathbb{R}^{g \times k \times 3} & \text{(Collection of } g \text{ local point groups)} \\
	& \text{For each } j \in \{1, \dots, g/2\} \text{ to form triplet } t_{i,j}: & \\
	& \quad t_{i,j}^{\text{num}_1} = X_i^{\text{groups}}[2(j-1)] \in \mathbb{R}^{k \times 3} & \text{(Raw point group 1)} \\
	& \quad t_{i,j}^{\text{num}_2} = X_i^{\text{groups}}[2j-1] \in \mathbb{R}^{k \times 3} & \text{(Raw point group 2)} \\
	& \quad I_{i,j}^{\text{indices}} = \{2(j-1), 2j-1\} \in \mathbb{Z}^{2} & \text{(Original indices of the two point groups)} \\
	& \quad t_{i,j} = (t_{i,j}^{\text{num}_1}, t_{i,j}^{\text{num}_2}, I_{i,j}^{\text{indices}}, j) & \text{(The } j \text{-th raw triplet)} \\
	& \mathcal{T}_i = \{ t_{i,1}, t_{i,2}, ..., t_{i,g/2} \} & \text{(Set of } g/2 \text{ raw triplets for } X_i\text{)}\\
	& \text{For each } t_{i,j} \text{ to form triplet token } t_{i,j}^{\text{token}}: & \\
	& \quad t_{i,j}^{\text{num}_1'} = \text{PointEmbedding}(t_{i,j}^{\text{num}_1}) \in \mathbb{R}^{d} & \text{(Point embedding, refer to PointGPT)} \\
	& \quad t_{i,j}^{\text{num}_2'} = \text{PointEmbedding}(t_{i,j}^{\text{num}_2}) \in \mathbb{R}^{d} & \text{(Point embedding, refer to PointGPT)} \\
	& \quad t_{i,j}^{\text{num}_1''} = \text{ZeroPad}(t_{i,j}^{\text{num}_1'}) \in \mathbb{R}^{384} & \\
	& \quad t_{i,j}^{\text{num}_2''} = \text{ZeroPad}(t_{i,j}^{\text{num}_2'}) \in \mathbb{R}^{384} & \\
	& \quad t_{i,j}^{\text{num}_{\text{concat}}} = \text{Concat}(t_{i,j}^{\text{num}_1''}, t_{i,j}^{\text{num}_2''}) \in \mathbb{R}^{768} & \\
	& \quad t_{i,j}^{\text{num}} = \text{Linear}(t_{i,j}^{\text{num}_{\text{concat}}}) \in \mathbb{R}^{512} & \text{(Numerical embedding)} \\
	& \quad t_{i,j}^{\text{topo}^{\text{local}}_{\text{emb}}} = \text{TopologyEmbedding}(I_{i,j}^{\text{indices}}) \in \mathbb{R}^{2 \times 512} & \\
	& \quad t_{i,j}^{\text{topo}^{\text{local}}} = \text{MeanReduce}(t_{i,j}^{\text{topo}^{\text{local}}_{\text{emb}}}) \in \mathbb{R}^{512} & \text{(Local topology embedding)} \\
	& \quad t_{i,j}^{\text{topo}^{\text{global}}} = \text{PositionEmbedding}(j) \in \mathbb{R}^{512} & \text{(Global topology embedding)} \\
	& \quad t_{i,j}^{\text{token}} = t_{i,j}^{\text{num}} + t_{i,j}^{\text{topo}^{\text{local}}} + t_{i,j}^{\text{topo}^{\text{global}}} \in \mathbb{R}^{512} & \text{(Final triplet token)} \\
	& \mathcal{T}_i^{\text{token}} = \{ t_{i,1}^{\text{token}}, t_{i,2}^{\text{token}}, ..., t_{i,g/2}^{\text{token}} \} & \text{(Set of } g/2 \text{ triplet tokens for } X_i\text{)}
\end{flalign*}}
\end{nolinenumbers}
\vspace{0em}

In summary, Pangaea leverages triplet encoding to unify disparate data representation methods across various modalities, effectively paving the way for truly comprehensive multi-modal learning. Triplet encoding represents a significant step towards general-purpose data encoding, and our future work will focus on further optimizing its efficiency and investigating alternative, equally robust encoding paradigms.

\subsection*{Details of unified pre-training process}
The pre-training of \method{} with 20 million parameters (20M) was conducted on a cluster equipped with 64 $\times$ Ascend NPU D910B cards, totaling approximately 55 hours. Ascend-specific operators were adapted using \textsc{torch\_npu}. The entire pre-training lasted for 240,000 steps, employing a cosine with restarts training strategy with a warm-up ratio of 0.03 and a base learning rate of $2\times10^{-4}$. The optimizer was AdamW with a weight decay of 0.05, and the total batch size was 4,096.

We have also adopted a series of techniques to stabilize the pre-training process.
First, missing values in table and image data were handled as follows. We imputed missing discrete values with the most frequently occurring value. For continuous variables, we used the mean as a robust estimator that minimizes the impact of outliers and maintains the central tendency of the data. 
For time series data, missing values (denoted as NaN) were filled with zeros to maintain the continuity and temporal coherence of the series. 
For image data, we referred to the pre-processing method proposed in MAE~\cite{he2022masked}. 
For text data, we followed the random sentence chunking strategy and adopted pre-trained text tokenizer used in BERT~\cite{devlin-etal-2019-bert}.
Second, normalization was applied across different modalities. Table data were normalized column-wise per dataset, while time series data were locally normalized over input sequences of $256$ time steps. Normalization of image data followed MAE~\cite{he2022masked}. Due to the significant numerical disparities in graph data, it was not normalized.
Third, for data augmentation, we integrated masking~\cite{he2022masked} and noising~\cite{vincent2008extracting} techniques. For table, time series, and graph data, we initially applied a 10\% random masking (setting values to 0) and added Gaussian noise with a variance of 0.1 to the remaining 90\% values of the data. For text data, following the masked language modeling strategy introduced in BERT, we replace 15\% of the token IDs with the \texttt{[MASK]} token ID. For image data, following the masked reconstruction strategy in MAE~\cite{he2022masked}, 75\% of the triplets were replaced with a shared learnable token.

During the reconstruction process, after completing missing value imputation, normalization, and data augmentation, the raw data were encoded into a triplet set and then mapped to a set of triplet tokens. After unified encoding, the resulting triplet tokens were concatenated with a randomly initialized pre-training token, also referred to as a reconstruction token, placed at the beginning to prevent the number of triplets from influencing the reconstruction token. 
This concatenated input was fed into the triplet transformer, which utilized an additional MLP head to output the reconstructed raw data. 
For table, time series, and graph data, the pre-training token was leveraged to decode and reconstruct the entirety of the raw data, with the reconstruction objective defined as mean square loss. For image data, reconstruction was performed by independently decoding each learnable token to recover the raw pixel grids of the corresponding masked triplet, with the mean squared loss used as the reconstruction objective. For text data, we decoded triplet tokens containing \texttt{[MASK]} token IDs to reconstruct the two raw text token IDs of the corresponding triplet, with the cross-entropy (CE) loss as the objective.

During the five-modal pre-training process, the uni-modal pre-training loss curves under different modality combinations converged to stable values, as shown in Extended Data Fig.~\ref{app_fig:effectiveness_modality_combinations}, demonstrating the effectiveness of the pre-training strategy employed by \method{}. However, due to varying difficulty levels, the convergence values differed among these combinations. As illustrated in Extended Data Fig.~\ref{app_fig:effectiveness_loss_backpropagation}, \method{} achieved a faster convergence rate compared to \method{}$_\textit{\textbf{CT}}$. Here, \method{}$_\textit{\textbf{CT}}$ represents \method{} processes each modality in parallel, and then backpropagates the loss of each modality independently. Regarding different model parameters shown in Extended Data Fig.~\ref{app_fig:effectiveness_model_parameters}, larger \method{} models exhibited faster convergence rates and lower convergence values, benefiting from their strong representation capabilities. The small difference in convergence values between the 20M and 100M models suggested that the current pre-training data size was insufficient for the 100M model. Therefore, we should continue to increase data size to help large-scale parameter models achieve faster and lower convergence values, aligning with the conclusions of traditional scaling laws~\cite{kaplan2020scaling}.

\subsection*{Details of supervised fine-tuning}
During the downstream fine-tuning phase, the MLP head used during pre-training will be re-initialized, while all other modules will inherit the pre-trained weights. 
The performance percentages of \method{}, \method{}$_\text{w/o}$, and competitive models are shown in Fig.~\ref{fig:main}e, with detailed results provided in Extended Data Table~\ref{tab:mian_res_info}. For the competitive models, we selected XGB~\cite{XGBoost} for 24 table tasks, iTransformer~\cite{liu2023itransformer} for six time-series tasks, ResNet~\cite{DBLP:conf/cvpr/HeZRS16} for three vision tasks, AST~\cite{gong2021ast} for two audio tasks, FAGCN~\cite{bo2021lowfrequencyinformationgraphconvolutional} for three graph tasks, PointNet++~\cite{qi2017pointnetplusplus} for three point cloud tasks, and BERT-Tiny~\cite{devlin-etal-2019-bert} for four text tasks. Specifically, XGB was run with default hyperparameters. iTransformer and FAGCN results were taken directly from their respective papers~\cite{liu2023itransformer, platonovcritical}. ResNet50 was fine-tuned using pretrained MoCo weights~\cite{he2020moco}. AST used a ViT-S backbone pretrained on ImageNet with DINO~\cite{caron2021emerging}. PointNet++ results followed the original benchmark~\cite{uy2019revisiting}, and BERT-Tiny was fine-tuned on the corresponding text datasets.

We selected various table classification and regression datasets from the tabular benchmark~\cite{grinsztajn2022tree} for 24 supervised fine-tuning tasks as shown in Extended Data Table~\ref{tab:mian_res_info}. The training and testing datasets were partitioned in an 8:2 ratio. For table classification tasks, the MLP head decoded the pre-training token and outputs the number of categories for the dataset. These classification tasks employ binary cross-entropy (BCE) loss and are evaluated by ACC. For table regression tasks, the MLP head decoded the pre-training token to output regression values for the task, guided by mean square loss and evaluated by MSE.

For time series forecasting tasks, we followed the competitive models~\cite{Yuqietal-2023-PatchTST,liu2023itransformer} to predict future 96 time steps based on historical 96 time steps using six time series datasets as shown in Extended Data Table~\ref{tab:mian_res_info}. The MLP head decoded the pre-training token and directly output values for the future 96 time steps, guided by the mean square loss and evaluated by MSE. The data split method in iTransformer~\cite{liu2023itransformer} was utilized in all time series forecasting tasks.

For vision classification tasks, we standardized images to a uniform size of $224 \times 224 \times 3$ for classification tasks. Using three datasets as shown in Extended Data Table~\ref{tab:mian_res_info}, we fed the resized images into \method{} without additional masking, and evaluated the performance through the pre-partitioned training and testing splits of the corresponding dataset. In cases where the original datasets were not pre-partitioned, a uniform 8:2 training and test set split was applied. The MLP head decoded the pre-training token and directly output the number of categories in the image dataset. Model was evaluated using ACC and optimized with CE loss.

For audio classification tasks, we used two audio datasets as shown in Extended Data Table~\ref{tab:mian_res_info}. Following the established common setting~\cite{gong2021ast}, we converted audio files into Mel spectrograms ($512 \times 128 \times 3$) using the \textsc{TorchAudio} library, treating them as visualized audio images for classification. The MLP head decoded the pre-training token and directly outputs the number of categories in the dataset, guided by CE loss and evaluated by ACC. The partition method was the same as the vision classification task.

For graph classification tasks, we used three graph datasets as shown in Extended Data Table~\ref{tab:mian_res_info}, and evaluated the model performance through the pre-partitioned training and testing splits of the corresponding dataset. The MLP head decoded the pre-training token and directly output the number of categories in the dataset, guided by CE loss and evaluated by ACC.

We conducted text classification tasks on four datasets, shown in Extended Data Table~\ref{tab:mian_res_info}. Model performance was assessed using their pre-partitioned training and testing splits. An MLP head decoded the pre-training token, directly outputting the number of categories. This process was guided by CE loss and evaluated via the F1 score.

For point cloud classification tasks, we apply the same pre-processing method described in PointGPT~\cite{chen2023pointgpt} to the sampled points of three datasets. We then evaluate model performance using the pre-partitioned training and testing splits for each corresponding dataset. The MLP head decodes the concatenated pre-training token (as introduced in PointGPT) and directly outputs the number of categories, guided by CE loss and evaluated by ACC.

For scientific tasks, the performance percentages of \method{} and competitive models are shown in Fig.~\ref{fig:ai4s}a with detailed results provided in Extended Data Table~\ref{tab:ai4s_res_info}. We carefully selected competitive models for each task: MTL-BERT~\cite{zhang2022pushing} for blood-brain barrier penetration prediction; MAE-ViT-B~\cite{he2022masked}, ResNet, ViT-B~\cite{dosovitskiy2020image}, and Swin-Tiny~\cite{liu2021swin} for prostate cancer grading; XGB for high-energy particle identification, reservoir property estimation, and molecule electronic properties prediction; BERT, LGB, Galactica (GAL)~\cite{taylor2022galacticalargelanguagemodel}, MoMu~\cite{su2022molecularmultimodalfoundationmodel}, KV-PLM~\cite{zeng2022predicting}, and GROVER~\cite{rong2020self} for drug toxicity prediction; CycPeptMP~\cite{li2024cycpeptmp} for cyclic peptide membrane permeability prediction; Radio Galaxy Zoo~\cite{slijepcevic2022radio} and E2CNNRadGal~\cite{scaife2021fanaroff} for classifying active galactic nuclei; decision tree (DT)~\cite{quinlan1986induction} for drug consumption prediction; PatchTST~\cite{Yuqietal-2023-PatchTST} for global temperature forecasting; AST~\cite{gong2021ast} for marine mammal vocalization classification; MAN-SF~\cite{sawhney2020deep} for stock movement prediction; CartNet~\cite{sole2025cartesian} for material band gap prediction; and BERT-Tiny~\cite{devlin-etal-2019-bert} for MMLU and MSC2020 tasks. Specifically, results for MTL-BERT, GAL, MoMu, KV-PLM, GROVER, CycPeptMP, MAN-SF, and CartNet were taken directly from their respective papers, while results for other models were generated using their default hyperparameters.

As for the drug molecule toxicity prediction task, the Tox21 dataset was divided using scaffold splitting~\cite{wu2018moleculenet}, with the number of samples for training, validation, and testing being 6,264, 783, and 784 respectively. Twelve binary classification subtasks were derived from 12 pathways to determine toxicity, evaluated using AUC and optimized with BCE loss.

For the global temperature forecasting task, the 360 grid points at the equatorial region were set as training datasets, inference the worldwide grid points ($360 \times 181$). Each point, derived from integer latitude and longitude, was considered an independent time series. The training/validation/testing split of series was 629/192/192, and RMSE was used as the evaluation metric, with mean square loss directly guiding the learning process. \method{} input the past 96 time steps (monthly averaged data in here) and forecast the future 192 time steps, using a rolling window training approach~\cite{Yuqietal-2023-PatchTST}.

For the reservoir property estimation task, the training and test sets consist of well logs from 9 and 4 wells respectively, with RMSE as the evaluation metric, using mean square loss to guide the learning process. Other settings were the same as the table regression task.

For the prostate cancer grading task, the randomly selected 50 and 13 cases from Subset 1 of the original training set were viewed as the final training and test sets, respectively. A grand total of 497,059 images and 87,001 images of size $224 \times 224 \times 3$ at $1\times$ magnification were extracted from the WSIs in the training and test sets, using a threshold of $>75\%$ annotated mask and $<85\%$ black/white pixel count. This 5-class classification task used ACC for evaluation and CE loss as the objective.

For Higgs boson classification~\cite{baldi2014searching}, we conducted experiments on three distinct feature sets, which included all 28 features, the first 21 low level kinematic features, and the last seven high level features derived by physicists. The last 500,000 examples of the dataset were allocated as the test set, with the remaining data utilized for training. Pangaea directly outputted a binary label, guided by BCE loss and evaluated by accuracy.

For blood brain barrier penetration classification~\cite{wu2018moleculenet}, the BBBP dataset was split into a training set of approximately 1,642 samples, a validation set of about 205 samples, and a test set of approximately 205 samples. All samples were transformed into graph format from SMILES notation. Pangaea directly outputted a binary label for each compound, guided by BCE loss and evaluated by AUC.

For molecule electronic properties prediction, the QM7 dataset was split into a training set of approximately 5,728 samples, a validation set of about 716 samples, and a test set of approximately 716 samples. All samples were transformed into graph format from the origin SMILES notation format. Pangaea directly predicted a single real valued atomization energy, guided by mean square loss and evaluated by MAE.

For material band gap prediction, we utilized a dataset of 18,172 samples. These samples were randomly split into a training set of approximately 14,538 samples and a test set of approximately 3,634 samples. Pangaea directly predicted a single real valued band gap (mBJ), guided by mean square loss and evaluated by MAE.

For membrane permeability prediction, the dataset was split into a training set of 5,961 samples and a test set of 1,490 samples. Cyclic peptides were transformed into molecular graphs from SMILES notation, with atoms as nodes and chemical bonds as edges to capture their structural and chemical properties. Pangaea directly outputted a binary label indicating membrane permeability, guided by BCE loss and evaluated by ACC.

For drug consumption prediction, the dataset of 1,885 samples was randomly split into a training set of approximately 1,508 samples and a test set of approximately 377 samples. Input consisted of 17 tabular features encompassing personality and demographic factors, guided by the Five Factor Model of personality. Pangaea directly outputted binary labels for each of the 18 different drug types, guided by BCE loss and evaluated by ACC.

For marine mammal vocalization classification, we converted audio files into Mel spectrograms, treating them as images for classification. We split each category's samples into an 80\% training set and a 20\% test set (with a minimum of one sample per test set category). The MLP head decoded the pre-training token to directly output the number of categories, guided by CE loss and evaluated by ACC.

For the massive multitask language understanding task, multiple choice questions served as input, processed using standard natural language processing techniques for tokenization and formatting. The model was trained on more than 99,800 samples. Subsequently, it was tested on all 57 diverse MMLU subjects. Each of these subjects has predefined training, validation, and test splits. For our evaluation, results were aggregated over all test sets, totaling approximately 7,125 questions. The model outputted a classification across four choice answers, guided by CE loss and evaluated by ACC.

For the radio galaxy classification task, images of radio galaxies served as input, treated as a classic image classification problem. The full  dataset comprised 1,222 samples, and further divided into two parts: a confident part and an uncertain part. The full dataset was randomly split into approximately 1,069 training samples and 153 testing samples. The confident part similarly included 729 training samples and 104 testing samples. Consistent with competitive models, the small uncertain part was excluded from our testing. The model outputted a morphological classification label for each galaxy, guided by CE loss and evaluated by ACC.

For the mathematics subject classification task, input consisted of mathematical text, specifically paper abstracts. These textual inputs were stemming and processed using standard natural language processing techniques for tokenization and conversion into numerical embeddings. The dataset comprised 159,175 articles, randomly split into a training set of 101,426 articles, a validation set of 17,994 articles, and a test set of 39,755 articles. The model outputted a specific MSC classification code for one of the 31 top-level classes, guided by CE loss and evaluated by ACC.

For stock movement prediction, we used the data split method from StockNet. The training set consists of 20,339 movements from January 1, 2014, to August 1, 2015. We used 2,555 movements from August 1, 2015, to October 1, 2015, for development, and 3,720 movements from October 1, 2015, to January 1, 2016, for testing. The MLP head decoded the pre-training token to output the binary prediction, guided by BCE loss and evaluated by the F1 score.

\subsection*{Ablation study}
The ablation study was conducted on modality combinations, model parameters, and data sizes, with results presented as follows.

In the fine-tuning results for modality combinations (Extended Data Fig.~\ref{app_fig:finetuning_modal_combination1} and \ref{app_fig:finetuning_modal_combination2}), the performance of the four-modal pre-trained model generally exceeded that of other combinations, particularly in random initialization (\method{}$_\text{w/o}$, w/o). This demonstrated that integrating more pre-training modalities enhances downstream task performance, with specific combinations leading to varying gains. Notably, as shown in Extended Data Fig.~\ref{app_fig:finetuning_modal_combination1}a, the four-modal pre-trained model better mitigated overfitting, benefiting from richer and more comprehensive accumulated universal knowledge.

Regarding model parameters (Extended Data Fig.~\ref{app_fig:finetuning_model_scale1} and \ref{app_fig:finetuning_model_scale2}), larger models consistently achieved faster convergence rates and lower convergence values due to their strong representation capabilities. However, on smaller datasets, larger models were more prone to overfitting, as depicted in Extended Data Fig.~\ref{app_fig:finetuning_model_scale1}a. In such cases, we should avoid full fine-tuning whenever possible and consider the LoRA technique~\cite{hu2021loralowrankadaptationlarge}.

For pre-training data sizes of each modality (Extended Data Fig.~\ref{app_fig:finetuning_precentage_data1} and \ref{app_fig:finetuning_precentage_data2}), increasing pre-training data for each modality correlated with higher performance gains, especially in random initialization (\method{}$_\text{w/o}$, w/o). This indicated that as pre-training data grow, the accumulated universal knowledge leads to improved performance and reduced overfitting, as illustrated in Extended Data Fig.~\ref{app_fig:finetuning_precentage_data1}g, and Extended Data Fig.~\ref{app_fig:finetuning_precentage_data2}d, g.

Overall, these findings highlight the effectiveness of \method{} in scaling across modalities and accumulating universal knowledge. The performance gains from modality combinations, model parameters, and pre-training data sizes emphasize \method{}'s adaptability and generalization across various downstream tasks. \method{} enhances performance while mitigating overfitting, demonstrating the advantages of utilizing multi-modal data.

\subsection*{Understanding modality, universal knowledge, and scaling effects}
This section clarifies our core definitions of modality and universal knowledge, subsequently demonstrating the necessary emergence of the scaling effect of modality.

We propose defining a modality as a distinct category of data characterized by its inherent topological structures and statistical properties. The topological structures define the fundamental organization and relationships among data elements, while the statistical properties describe their overall information distribution and characteristics. For example, text modality comprises sequential words with grammatical relationships, exhibiting distinct semantic and frequency distributions. Image modality consists of pixel grids with spatial hierarchies, defined by its unique pixel value and texture patterns. Notably, while text (with its semantic and frequency distributions) and time series (characterized by temporal trends and statistical stationarity) share similar sequential topological structures, their fundamentally distinct statistical properties lead us to categorize them as different modalities.

The formal modality formulation clarifies the fundamental nature of universal knowledge. By leveraging our proposed unified data encoding, the topological structure of each modality is effectively deconstructed into a consistent form, thereby enabling the  accumulation of its statistical properties. This accumulation constitutes the modality-invariant, generalizable knowledge that can be seamlessly transferred across diverse downstream tasks. We define this comprehensively acquired knowledge, which significantly enhances performance in downstream applications, as universal knowledge in our work.

These definitions also clarify the scaling effect of modality. As the number of modalities increases, the volume of accumulated statistical properties from these modalities grows correspondingly. This comprehensive coverage directly contributes to the accumulation of richer universal knowledge, which in turn leads to improved performance on downstream tasks. This phenomenon clearly demonstrates the necessary emergence of the scaling effect of modality.

\subsection*{The derivation of the scaling effect of modality}
To leverage the duality (learned or not learned), independence (the accumulation of universal knowledge from each modality is independent), and discreteness (discrete distribution) characteristics presented in the knowledge accumulation process, the Bernoulli distribution is employed to model the universal knowledge gained from a single modality, thereby revealing the scaling effect of modality as the cumulative distribution function of a geometric distribution.

According to our universal knowledge definition, universal knowledge can be accumulated even through uni-modal pre-training. We assume this process is a binary and discrete event (either the knowledge is learned or it is not), which could be modeled using a Bernoulli distribution with probability $p$. Specifically, for a single modality, a random variable $X\sim Bernoulli(p)$ is introduced to represent whether universal knowledge is learned, where $\text{Pr}(X=1)=p$ is the probability of success. 
The expectation $\mathbb{E}(\text{Pr}(X=1))=p$ is the amount of universal knowledge learned from a single modality. 

Due to the parallel reconstruction strategy employed during multi-modal pre-training, the accumulation of universal knowledge from each modality can be modeled as an independent Bernoulli trial. As these trials are repeated, the accumulation process from multi-modal pre-training naturally follows the cumulative distribution function of a geometric distribution. Specifically, as the number of modalities increases to $k\in\mathbb{N}^+$, a random variable $Y$ is defined to represent the index of the modality where universal knowledge is first learned, \textit{i.e.}, $Y\sim Geometric(p)$. Mathematically, $\text{Pr}(Y=k)=(1-p)^{k-1}p$ means the probability that universal knowledge is first learned from $k$-th modality after the failure of the previous $k-1$ modalities.
The expectation $\mathbb{E}(\text{Pr}(Y=k))=(1-p)^{k-1}p$ can be regarded as the amount of universal knowledge first learned from the $k$-th modality. This amount is initially substantial and diminishes as $k$ increases, aligning with intuition.
After accumulating from $k$ modalities, the probability that universal knowledge has been learned from at least one modality is given by the cumulative distribution function of a geometric distribution $\text{Pr}(Y\leq k)=1-(1-p)^k$. Its expectation is used to quantify the accumulated amount of universal knowledge after learning from $k$ modalities. This aligns with the scaling effect of modality, which can be expressed as $y=1 - (1 - p)^x$, where $x$ is the number of modalities and $y$ represents the accumulated amount of universal knowledge. The upper bound of $1$ in the scaling effect, indicates that this amount eventually nears saturation, with diminishing contributions from each additional modality.

\subsection*{Related work}
\subsubsection*{Rationale for choosing graph-based modeling}
In line with our modality definition, data's statistical properties are organized by its inherent topological structures. This means data, a set of numerical values embedded within these structures, is naturally modeled as weighted graph.

This insight that reformulating all data modalities into weighted graphs aligns with a key insight from the survey~\cite{velickovic2023gnn}, namely the broad applicability of graph-structured data. However, the weighted graphs derived from different modalities are significantly heterogeneous in both structure and semantics, making it difficult to directly share a graph representation across them. To address this challenge, we introduce triplets as the atomic units of weighted graphs, and further regard them as the fundamental units of data across modalities, enabling a unified representation. This representation serves as the foundation for unified pre-training and universal knowledge transfer across modalities.

\subsubsection*{Advantages of triplet encoding compared to language encoding}
Our unified data encoding offers three key advantages over approaches primarily converting modalities to language~\cite{fatemi2024talk,shen2023crossmodal,dinh2022lift,mirchandani2023llm,lu2022frozen}, by converting data from any modality into a triplet format.

First, the triplet format enables unified pre-training across diverse modalities, accumulating universal knowledge beyond single modality scopes. Unlike methods~\cite{fatemi2024talk,shen2023crossmodal,dinh2022lift,mirchandani2023llm} converting data to a language format for downstream tasks, which is highly dependent on the capabilities of pre-trained language models designed for text tasks, Pangaea allows modalities to mutually benefit through unified pre-training.
Experiments on five modalities show Pangaea improves 2.3\% over text-only pre-training, indicating multi-modal pre-training enhances downstream performance.

Second, our approach utilizes a shared triplet tokenizer for triplet format of any modality, eliminating the need for modality-specific adaptation. In contrast, Large language models (LLMs) typically require additional encoders~\cite{shen2023crossmodal,dinh2022lift,mirchandani2023llm,lu2022frozen} or prompt engineering~\cite{fatemi2024talk} to align data into a language format. This alignment step often necessitates training separate models or components for each new task, thereby limiting generalization. Our experimental results on 45 downstream tasks and 15 scientific tasks demonstrate Pangaea’s strong potential to handle arbitrary tasks with one unified model. 

Thirdly, the triplet format enables Pangaea to process raw values directly, thereby preserving inherent precision, especially for continuous data. Conversely, LLMs convert data to discrete tokens, leading to unavoidable information loss. This loss is especially detrimental in scientific tasks requiring the capture of subtle differences between samples. Experiments on BBBP and Tox21 from MoleculeNet classification~\cite{wu2018moleculenet} demonstrate Pangaea's 26.2\% performance improvement over Galactica~\cite{taylor2022galacticalargelanguagemodel}, a model pre-trained on a large-scale scientific corpus. This supports Pangaea's advantage in tasks involving continuous data.

In summary, our triplet encoding demonstrates advantages in facilitating cross-modal pre-training, eliminating modality-specific encoders, and preserving data fidelity for diverse modalities, making it a compelling alternative for modeling diverse modalities.

\subsubsection*{Challenge in optimizing modality combinations}
As shown in Fig.~\ref{fig:ablation_exp}c, experimental results from various modality combinations reveal varying performances, suggesting both a modality affinity phenomenon and complex modality interactions. This variability indicates inherent conflicting information between different combinations, a challenge analogous to those encountered in multi-task learning~\cite{yu2020gradient,fifty2021efficiently}, where diverse tasks may present conflicting signals that impede generalization. 

Addressing this challenge would help us understand the modality affinity phenomenon, but it's beyond the present scope of our work. We'll leave this for future work, aiming to design a comprehensive framework to achieve state-of-the-art performance on downstream tasks by optimizing modality combinations.

\subsection*{The name of \method{}}
Pangaea is a geological concept that describes a supercontinent that existed during the late Paleozoic and early Mesozoic eras (see \url{https://en.wikipedia.org/wiki/Pangaea}). During this time, nearly all of Earth's continental plates were joined together, forming a massive landmass. Similarly, in the field of AI, we aim to construct such an AI supercontinent through unifying Intelligence Islands. Our model, \method{}, plays a crucial role in this endeavor. It is designed to learn from all modalities, accumulating universal knowledge from diverse modalities. By doing so, it is expected to gradually approach the true essence of general-purpose intelligence, thereby unifying Intelligence Islands and constructing an AI supercontinent.

\section*{Data Availability}\label{sec13}
All data needed to evaluate the conclusions are present below.
The table pre-training and fine-tuning datasets were adopted from OpenML (\url{https://www.openml.org/search?type=data&sort=runs&status=active}). All text datasets are sourced from the \url{https://huggingface.co/datasets}. The time series pre-training datasets are from the Monash Time Series Forecasting Repository obtained via the \textsc{GluonTS} library (\url{https://ts.gluon.ai/stable/api/gluonts/gluonts.dataset.repository.datasets.html}). The time series fine-tuning datasets are sourced from \url{https://github.com/thuml/SimMTM}. Graph datasets for pre-training are obtained from the node prediction task of the Open Graph Benchmark (\url{https://ogb.stanford.edu/docs/nodeprop/}) and the graph datasets for fine-tuning (Roman-Empire, Tolokers, Minesweeper) are obtained from the HeterophilousGraphDataset class of the \textsc{PyTorch Geometric} library. For vision datasets, ImageNet is sourced from the website \url{https://image-net.org/}, CIFAR-10/100 from \url{https://www.cs.toronto.edu/~kriz/cifar.html}, and Food101 from \url{https://data.vision.ee.ethz.ch/cvl/datasets_extra/food-101/}. For audio datasets, ESC-50 is available at \url{https://github.com/karolpiczak/ESC-50}, SpeechCommands from \url{http://download.tensorflow.org/data/speech_commands_v0.02.tar.gz}. Point cloud datasets are from the ScanObjectNN benchmark \url{https://hkust-vgd.ust.hk/scanobjectnn/}.

For 15 scientific tasks, the Tox21 dataset is from \url{https://ogb.stanford.edu/docs/graphprop/}, the VOLVE dataset is from \url{https://github.com/pddasig/Machine-Learning-Competition-2021/tree/main}, the ERA5 dataset is from \url{https://cds.climate.copernicus.eu/datasets/reanalysis-era5-single-levels-monthly-means?tab=overview}, the AGGC dataset is from \url{https://aggc22.grand-challenge.org/}, the StockNet dataset is from \url{https://github.com/yumoxu/stocknet-dataset}, the Watkins Marine Mammal Sound Database is from \url{https://cis.whoi.edu/science/B/whalesounds/index.cfm}, the MiraBest dataset is from \url{https://zenodo.org/records/4288837}, the BBBP and QM7 are from \url{https://moleculenet.org/}, the Higgs and Drug-consumption are from \url{https://archive.ics.uci.edu/dataset/}, the JARVIS-DFT is from \url{https://figshare.com/articles/dataset/jdft_3d-7-7-2018_json/6815699}, the MMLU is from \url{https://huggingface.co/datasets/cais/mmlu}, the CycPeptMPDB is from \url{http://cycpeptmpdb.com/}, the MSC2020 is from \url{https://mathscinet.ams.org/mathscinet/msc/msc2020.html}.

\section*{Code Availability}\label{sec14}
The pre-training setup utilized the \textsc{PyTorch} deep learning framework, accelerated by \textsc{DeepSpeed}, with the base framework (including training, evaluation, hyper-parameter configuration, argument parser, \textit{etc}.) derived from the \textsc{Transformers} library. For other details, please refer to the pre-training details of Methods section. We used the \textsc{torch\_npu} (\url{https://gitee.com/ascend/pytorch}) library to adapt \textsc{PyTorch} for the Ascend NPUs, along with other Python libraries such as \textsc{NumPy}, \textsc{Matplotlib}, \textsc{Scikit-Learn}, and \textsc{Pandas}.

\vspace{6pt}
\setcounter{enumiv}{41} 

\backmatter

\bmhead*{Author contributions}
J.C. designed and managed the research project.
H.W., Z.D. and R.C. wrote the paper.
R.C., Z.D., L.H., D.L., D.W. and Z.W. collected the pre-training and fine-tuning data.
L.H., Z.W. and Z.D. designed the model architecture.
L.H. ran the pre-training experiment.
Z.W., D.L. and S.P. ran the table data fine-tuning experiment.
Z.W. and Z.D. ran the time series and graph data fine-tuning experiment.
L.H. ran the vision data experiment.
R.C., D.G., D.W. and Z.W. collected the data of scientific tasks and ran its downstream experiments separately.
Z.L., L.F., J.S. and Y.L. improved the details on both techniques and paper writing.

\newpage
\setcounter{figure}{0}
\renewcommand{\figurename}{Extended Data Fig.}
\counterwithin*{figure}{part}
\setcounter{table}{0}
\renewcommand{\tablename}{Extended Data Table}
\counterwithin*{table}{part}

\clearpage

\begin{figure}[h]
    \centering
    \includegraphics[width=0.80\textwidth,keepaspectratio=true]{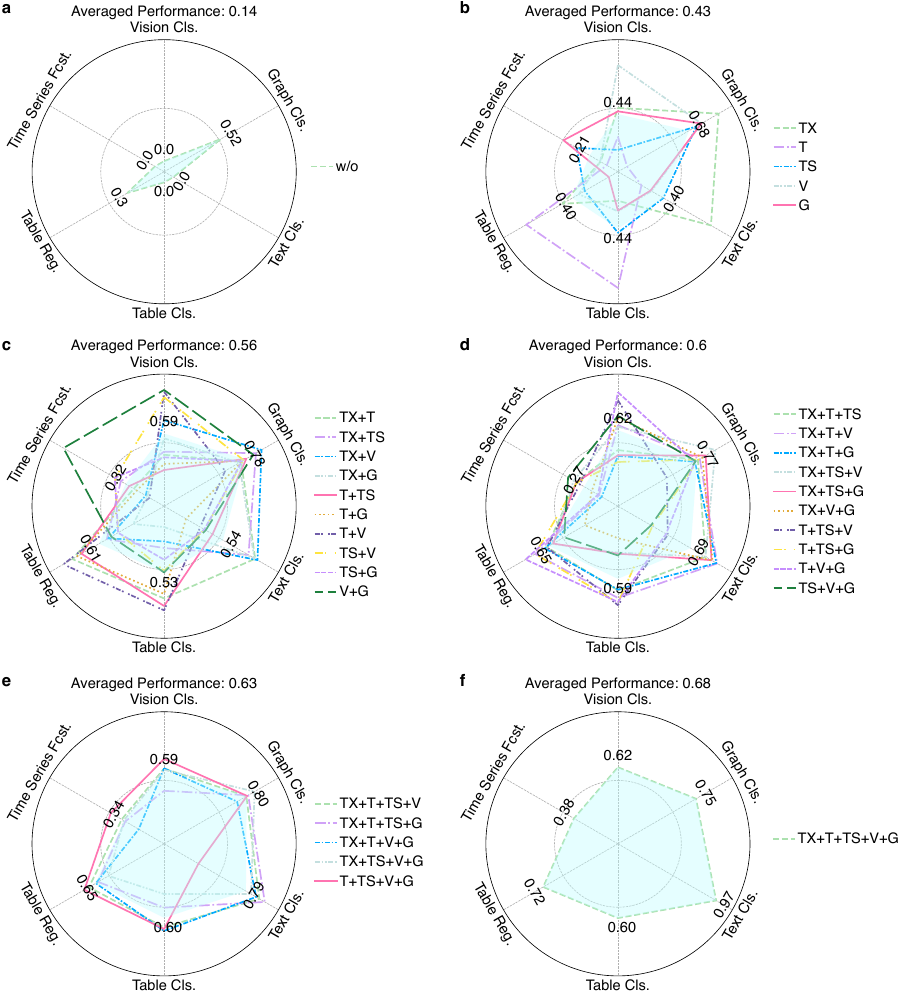}
    \caption{\textbf{Detailed normalized performances of various modality combinations across six types of downstream tasks.} This figure provides the magnified view of the radar plots originally presented in Fig.~\ref{fig:ablation_exp}a. Each line depicts the specific performance of a given combination, while the gray-filled area represents its average performance across these tasks. \textbf{a}, Downstream performances of \method{}$_\text{w/o}$. \textbf{b}, Downstream performances of five one-modal combinations (TX, T, TS, G, V). \textbf{c}, Downstream performances of ten two-modal combinations (TX+T, TX+TS, TX+V, TX+G, T+TS, T+G, T+V, TS+V, TS+G, V+G). \textbf{d}, Downstream performances of ten three-modal combinations (TX+T+TS, TX+T+V, TX+T+G, TX+TS+V, TX+TS+G, TX+V+G, T+TS+V, T+TS+G, T+V+G, TS+V+G). \textbf{e}, Downstream performances of five four-modal combinations (TX+T+TS+V, TX+T+TS+G, TX+T+V+G, TX+TS+V+G, T+TS+V+G). \textbf{f}, Downstream performances of one five-modal combination (TX+T+TS+V+G).
    }\label{app_fig:radar_plot}
\end{figure}

\clearpage

\begin{figure}[h]
    \centering
    \includegraphics[width=\textwidth,keepaspectratio=true]{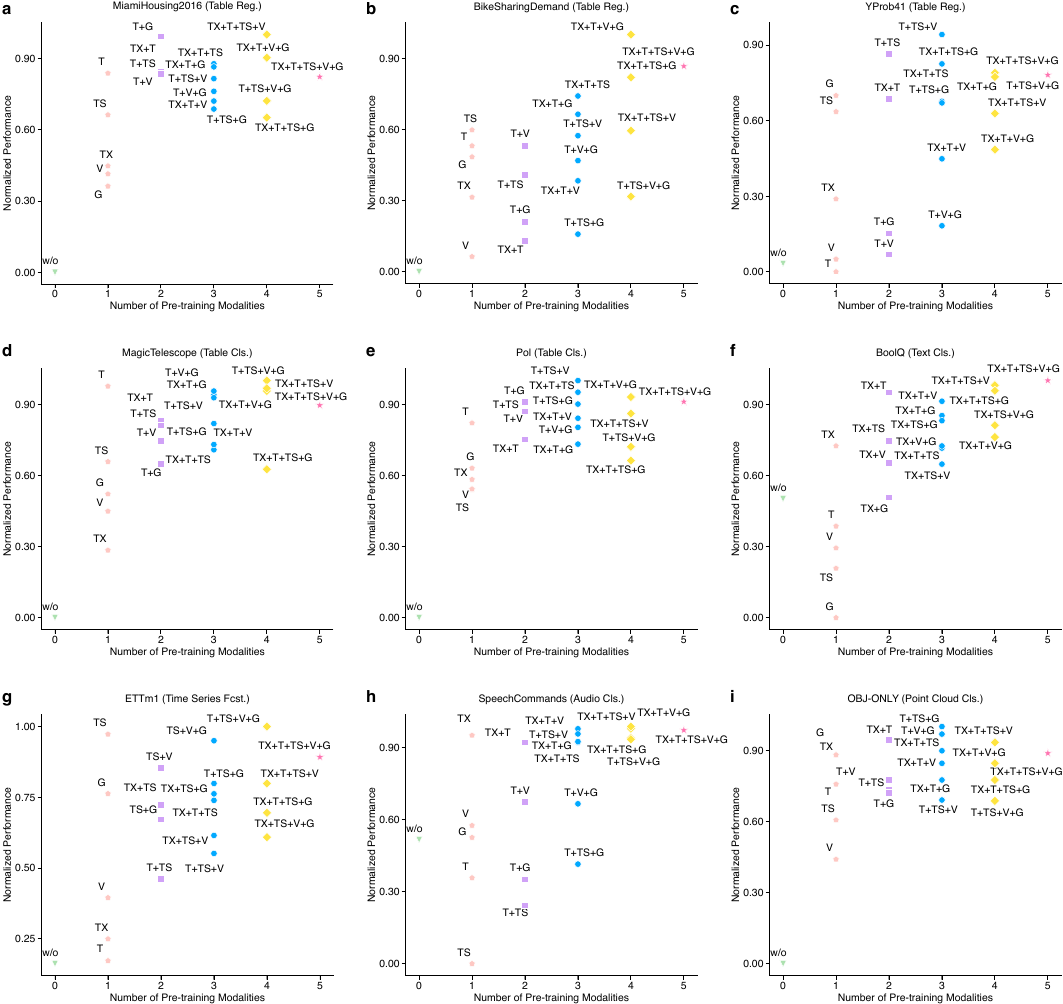}
    \caption{\textbf{Scaling effect of modality and modality affinity phenomenon across downstream tasks.} This figure presents detailed results from Fig.~\ref{fig:ablation_exp}a and c. \textbf{a}-\textbf{i}, Normalized performances across various downstream tasks, plotted against the number of pre-training modalities involved. These subfigures illustrate a general trend where incorporating more modalities tends to yield improved performance. They also highlight how different modality combinations contribute to varying degrees of gains, thereby demonstrating the modality affinity phenomenon.
    }\label{app_fig:model_scaling}
\end{figure}

\clearpage

\begin{figure}[h]
    \centering
    \includegraphics[width=0.80\textwidth,keepaspectratio=true]{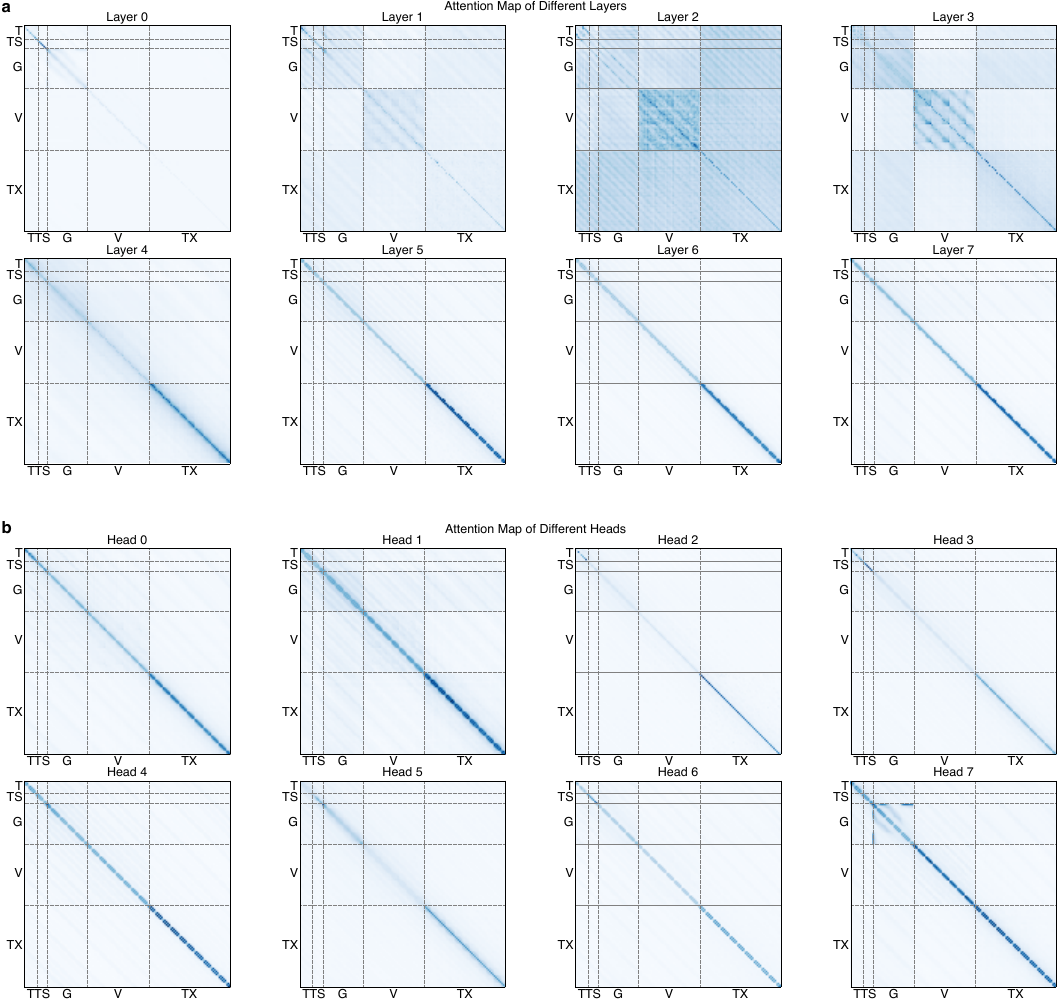} 
    \caption{\textbf{Layer-wise and head-wise attention maps across five modalities.} These attention maps, generated from pre-trained \method{}, illustrate token similarity across diverse modalities, as referenced in Fig.~\ref{fig:ablation_exp}f. Increased blueness indicates higher similarity between tokens. As expected, tokens on the diagonal (similarity to themselves) and sub-diagonal (similarity to adjacent tokens) consistently show higher similarity. \textbf{a}, Layer-wise Attention. This subfigure presents attention maps between tokens from the five pre-training modalities at the last (8th) head across different layers. \textbf{b}, Head-wise Attention. This subfigure displays attention maps between tokens from the five pre-training modalities at the last (8th) layer across different attention heads.
    }\label{app_fig:token_affinity}
\end{figure}

\clearpage

\begin{figure}[h]
    \centering
    \includegraphics[width=0.78\textwidth,keepaspectratio=true]{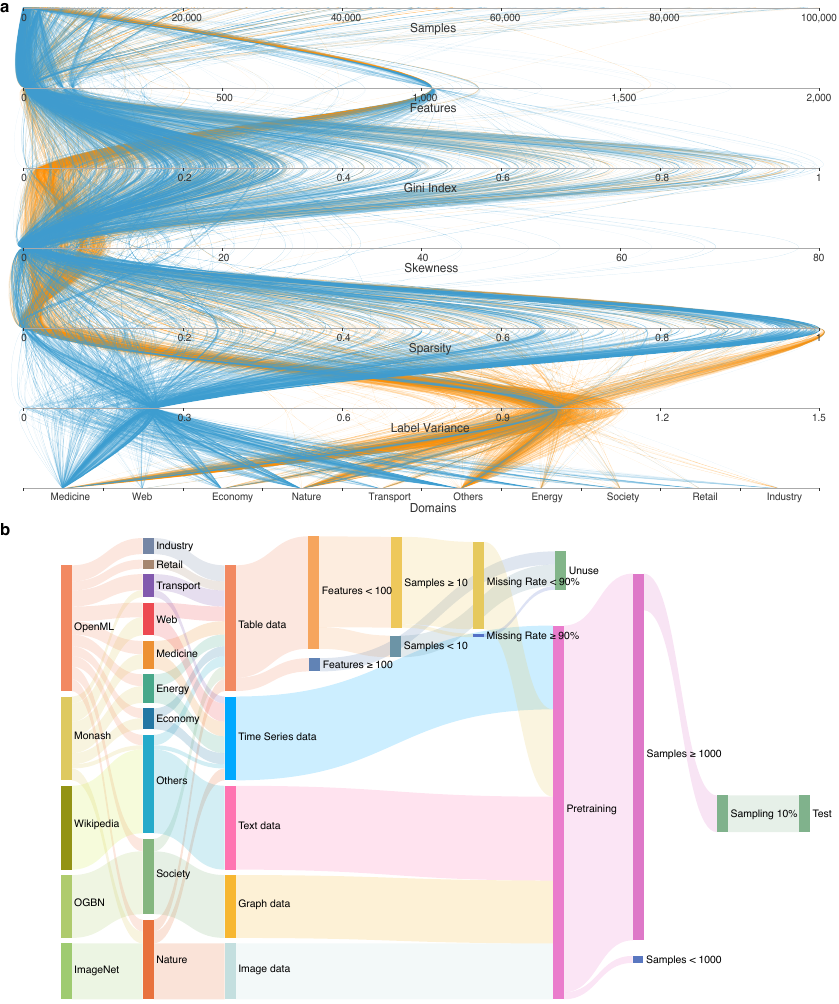}
    \caption{\textbf{Characteristics and processing of the pre-training data.} \textbf{a}, Characteristic distributions of table classification (orange) and table regression (blue) datasets. Each line means an individual table dataset. \textbf{b}, Sampling and filtering strategies applied to pre-training datasets.
    }\label{app_fig:data_description}
\end{figure}

\clearpage

\begin{figure}[h]
    \centering
    \includegraphics[width=\textwidth,keepaspectratio=true]{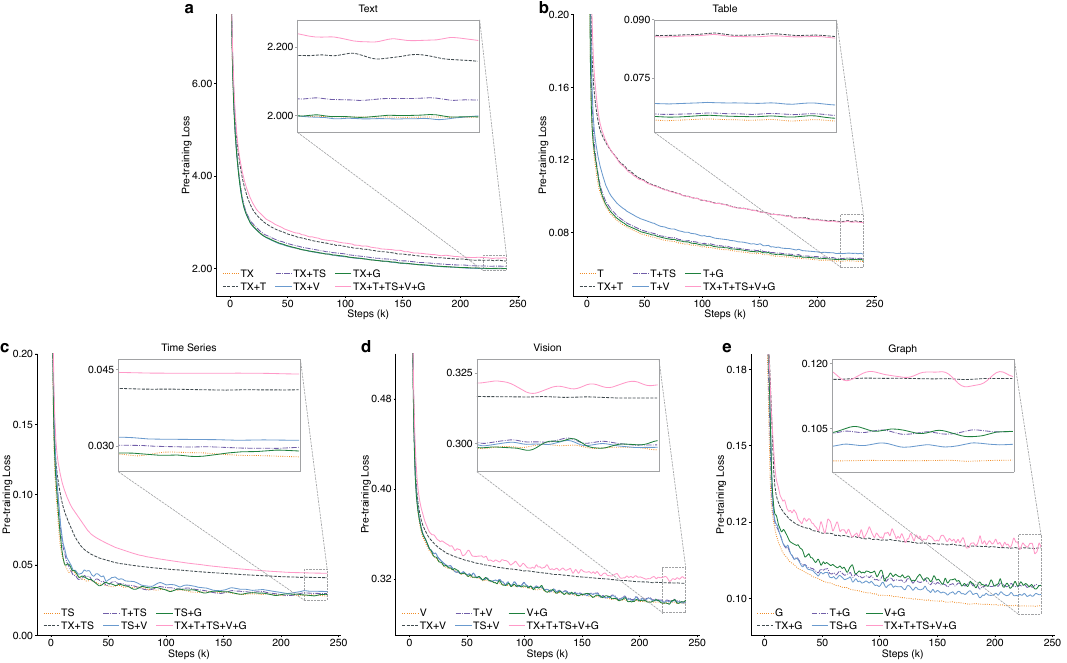} 
    \caption{\textbf{Effectiveness of the pre-training modality combinations.} \textbf{a}-\textbf{e}, Convergence curves of uni-modal pre-training loss under different pre-training modality combinations.
    }\label{app_fig:effectiveness_modality_combinations}
\end{figure}

\clearpage

\begin{figure}[h]
    \centering
    \includegraphics[width=0.6\textwidth,keepaspectratio=true]{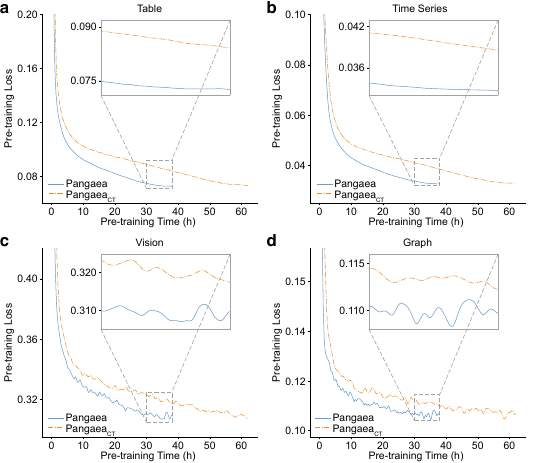}
    \caption{\textbf{Effectiveness of the pre-training loss backpropagation strategies.} \textbf{a}-\textbf{d}, Convergence curves of uni-modal pre-training loss under different strategies of pre-training loss backpropagation. 
    }\label{app_fig:effectiveness_loss_backpropagation}
\end{figure}

\clearpage

\begin{figure}[h]
    \centering
    \includegraphics[width=\textwidth,keepaspectratio=true]{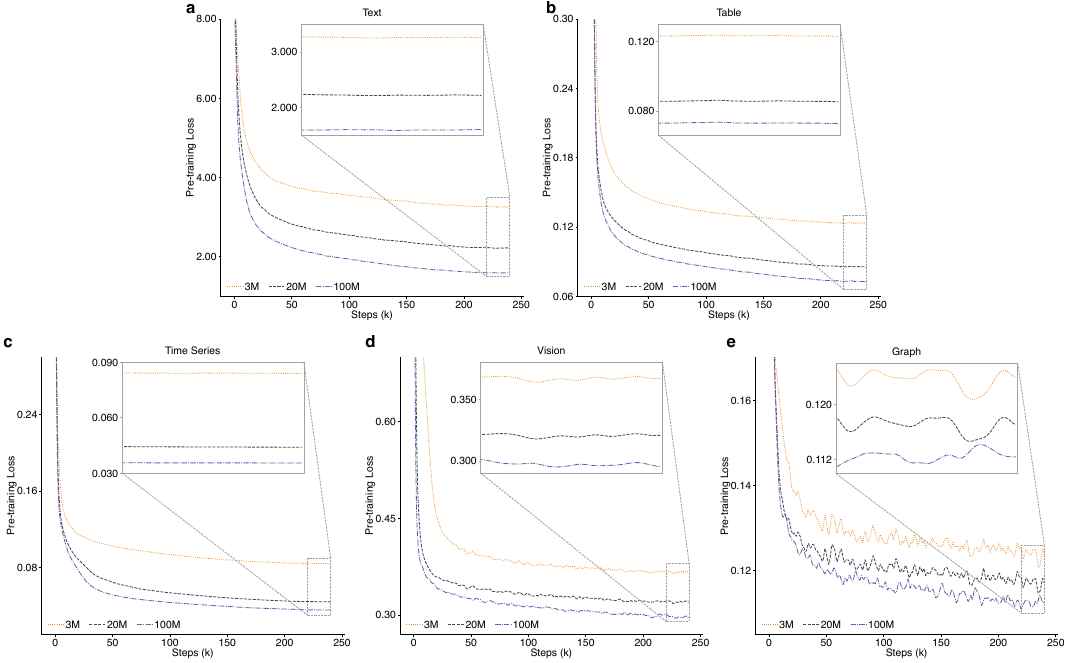} 
    \caption{\textbf{Effectiveness of the pre-training model parameters.} \textbf{a}-\textbf{e}, Convergence curves of uni-modal pre-training loss under different pre-training model parameters.
    }\label{app_fig:effectiveness_model_parameters}
\end{figure}

\clearpage

\begin{figure}[h]
    \centering
    \includegraphics[width=\textwidth,keepaspectratio=true]{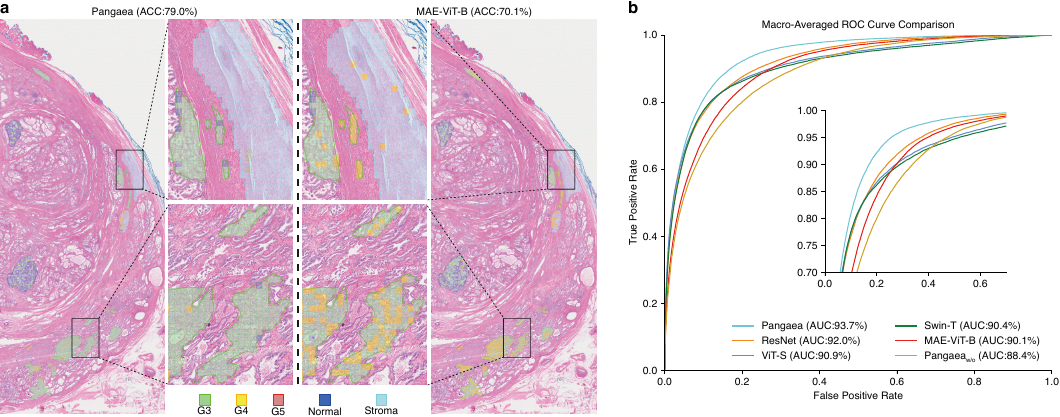}
    \caption{\textbf{Detailed results of \method{} on the prostate cancer grading task.} \textbf{a}, Detailed visualizations of ROI classification comparisons for prostate adenocarcinoma tissue. \textbf{b}, Comparison of macro-averaged ROC curves over various competitive models, with AUC indicated in the legend.
    }\label{app_fig:prostate_grading}
\end{figure}
\clearpage

\begin{figure}[h]
    \centering
    \includegraphics[width=0.86\textwidth,keepaspectratio=true]{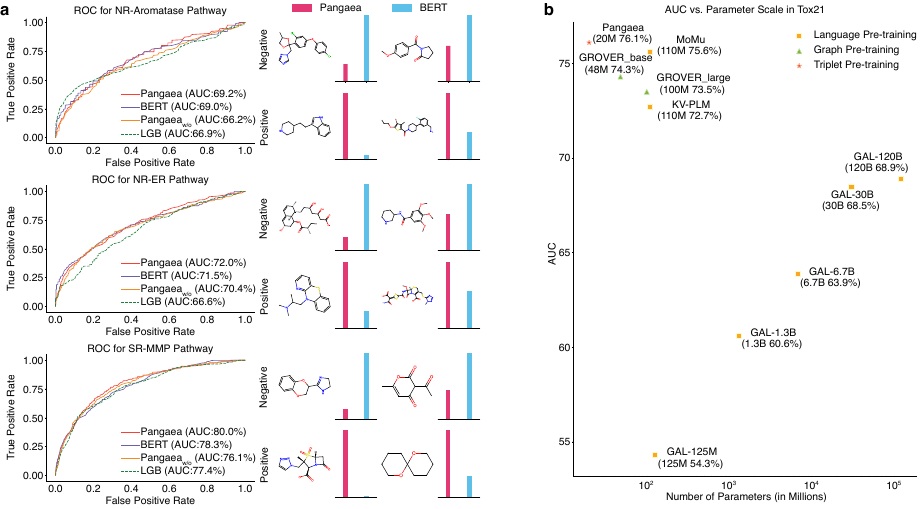}
    \caption{\textbf{Performance and efficiency of \method{} in the drug molecule toxicity prediction task.} \textbf{a}, ROC curves and case studies on the remaining 3 tasks, compared with competitive models. \textbf{b}, AUC versus model parameter in the Tox21 dataset. \method{} achieves a better AUC with fewer model parameters.
    }\label{app_fig:molecule_classification}
\end{figure}

\clearpage

\begin{figure}[h]
    \centering
    \includegraphics[width=\textwidth,keepaspectratio=true]{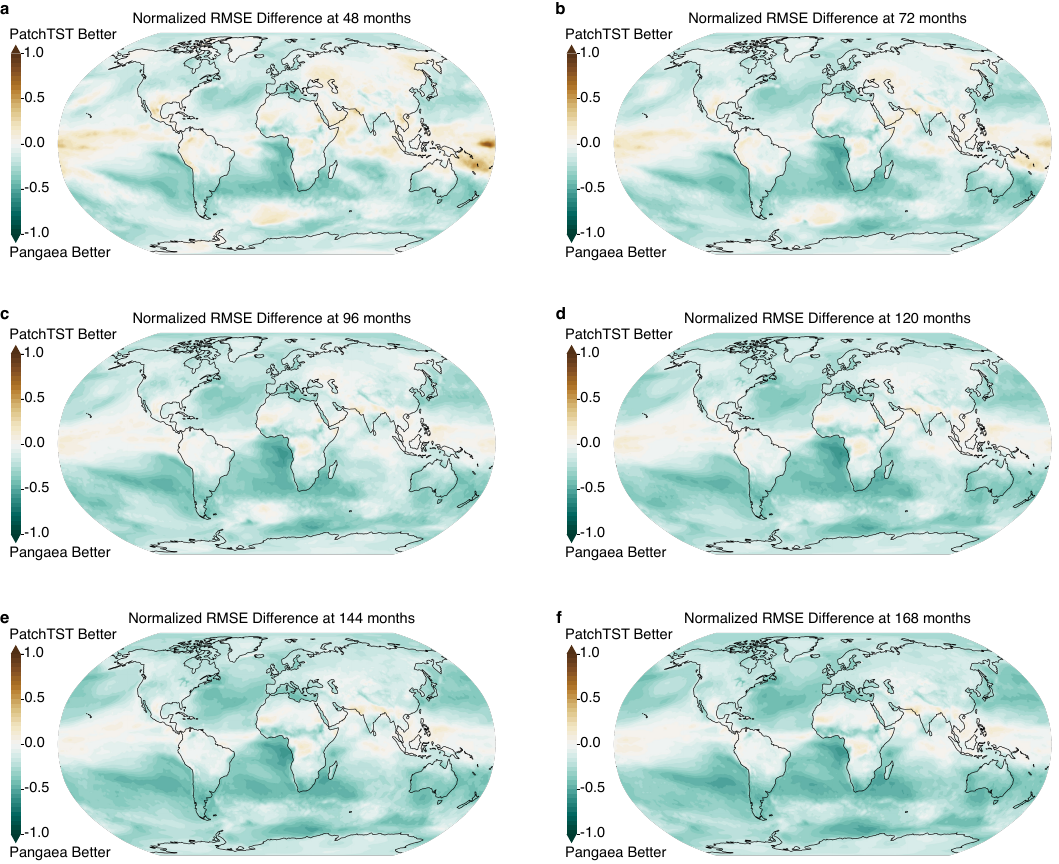}
    \caption{\textbf{Detailed results of \method{} for the global temperature forecasting task.} \textbf{a}-\textbf{f}, More visualization comparisons of \method{} and PatchTST among various prediction lengths (48-168 months), demonstrating the significant long-term predictive capability of \method{}.
    }\label{app_fig:global_temperature}
\end{figure}

\clearpage

\begin{figure}[h]
    \centering
    \includegraphics[width=0.8\textwidth,keepaspectratio=true]{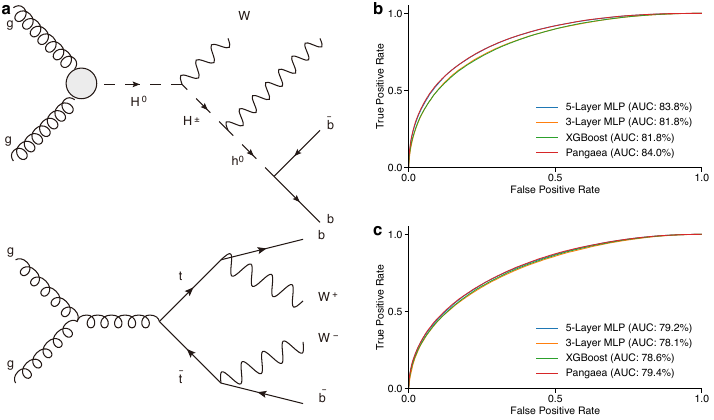}
    \caption{\textbf{Details of Higgs benchmark and results of \method{} on the high-energy particle identification task.} \textbf{a}, Diagrams for Higgs benchmark. This subfigure provides diagrams for the Higgs benchmark. The positive label (top) illustrates the signal process involving new exotic Higgs bosons, while the negative label (bottom) shows the background process involving top quarks. \textbf{b}, ROC curve of Pangaea and competitive models on all 28 features. \textbf{c}, ROC curve of Pangaea and competitive models on seven high-level features. \method{} consistently outperforms BERT-Tiny across these scenarios.
    }\label{app_fig:higgs_prediction}
\end{figure}

\clearpage

\begin{figure}[h]
    \centering
    \includegraphics[width=\textwidth,keepaspectratio=true]{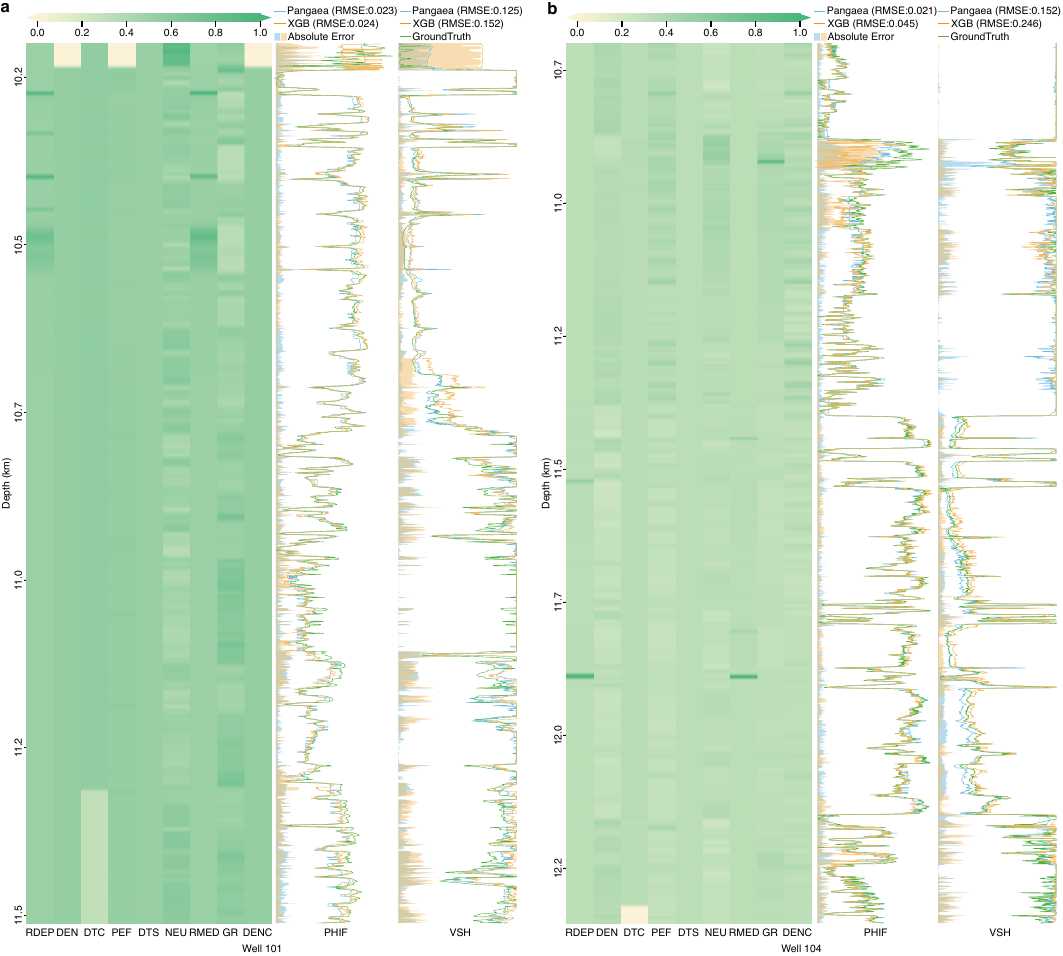}
    \caption{\textbf{Prediction results on remaining wells by \method{} for the reservoir property estimation task.} \textbf{a}, Performance comparison of \method{} and XGB in well 101. \textbf{b}, Performance comparison of \method{} and XGB in well 104. VSH, shale volume; PHIF, porosity.
    }\label{app_fig:well_logging}
\end{figure}





\clearpage
\begin{figure}[h]
    \centering
    \includegraphics[width=0.8\textwidth,keepaspectratio=true]{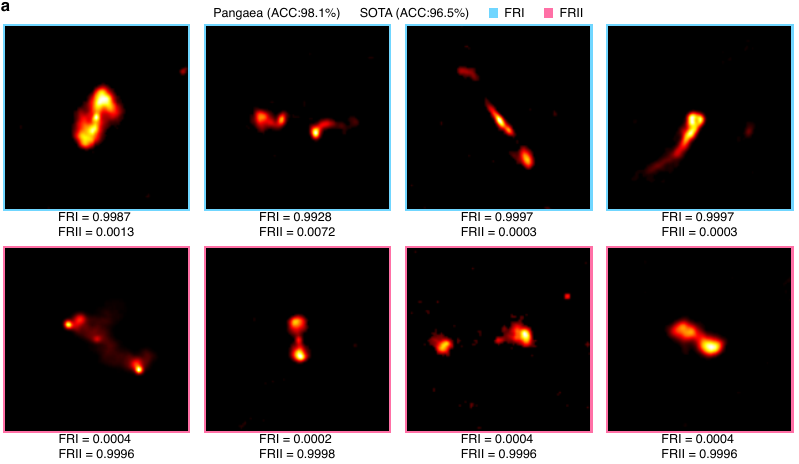}
    \caption{\textbf{\method{} predictions of the active galactic nuclei classification task.} \method{} still outperform SOTA results on Confident subset of the Mirabest dataset. The detailed prediction of \method{} on some radio samples are also presented.
    }\label{app_fig:nuclei_classification}
\end{figure}

\clearpage


\begin{figure}[h]
    \centering
    \includegraphics[width=\textwidth,keepaspectratio=true]{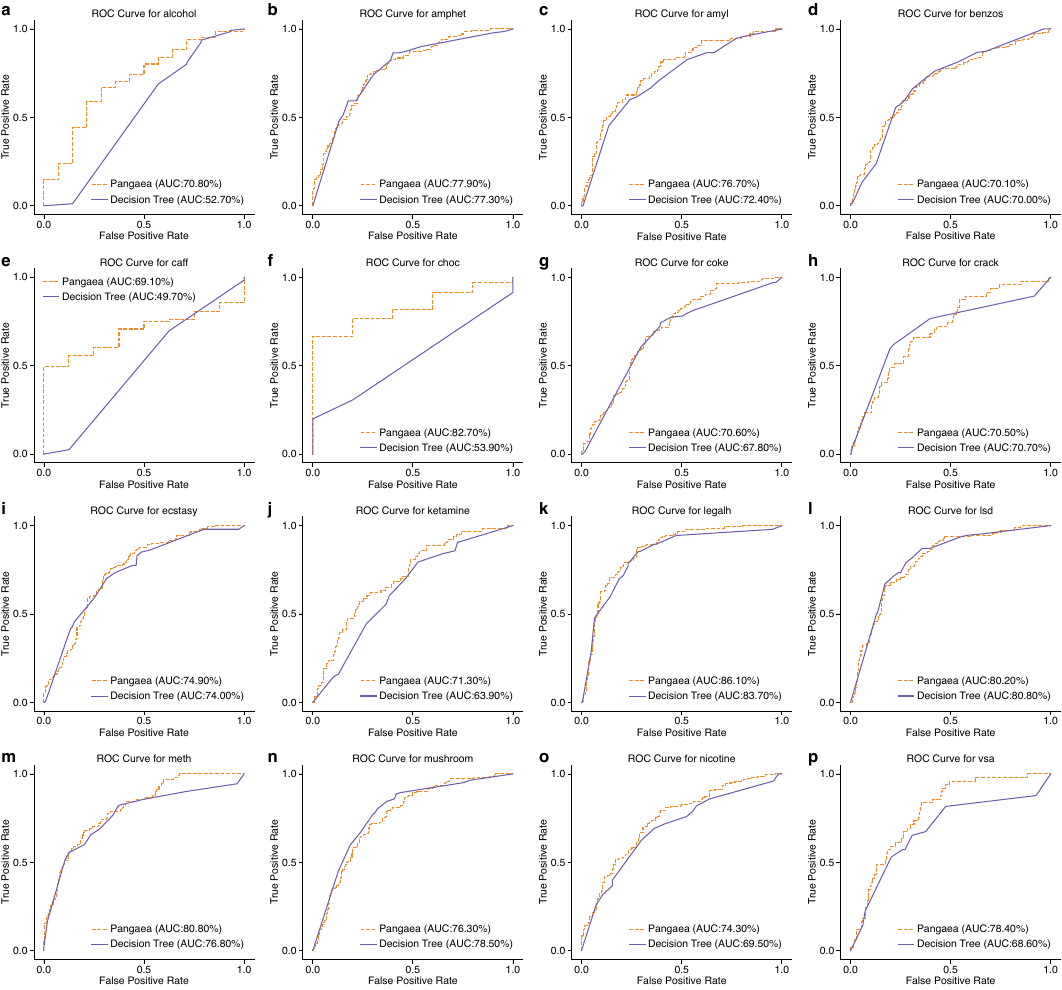}
    \caption{\textbf{ROC curves between \method{} and decision tree on the drug consumption prediction task.} \method{} consistently outperforms decision tree across 16 drug types.
    }\label{app_fig:drug_consumption}
\end{figure}

\clearpage

\begin{figure}[h]
    \centering
    \includegraphics[width=\textwidth,keepaspectratio=true]{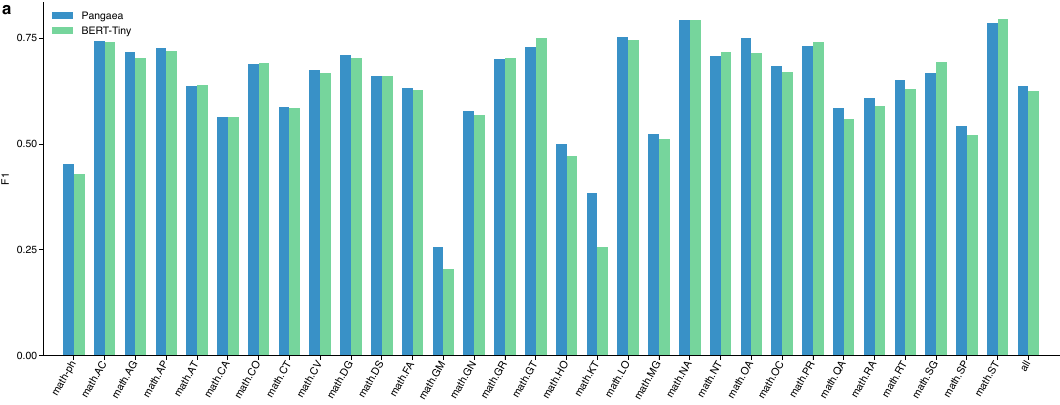}
    \caption{\textbf{Performance comparisons of \method{} on the mathematics subject classification task.} \method{} consistently outperforms BERT-Tiny across almost all 31 categories.
    }\label{app_fig:msc}
\end{figure}

\clearpage

\begin{figure}[h]
    \centering
    \includegraphics[width=\textwidth,keepaspectratio=true]{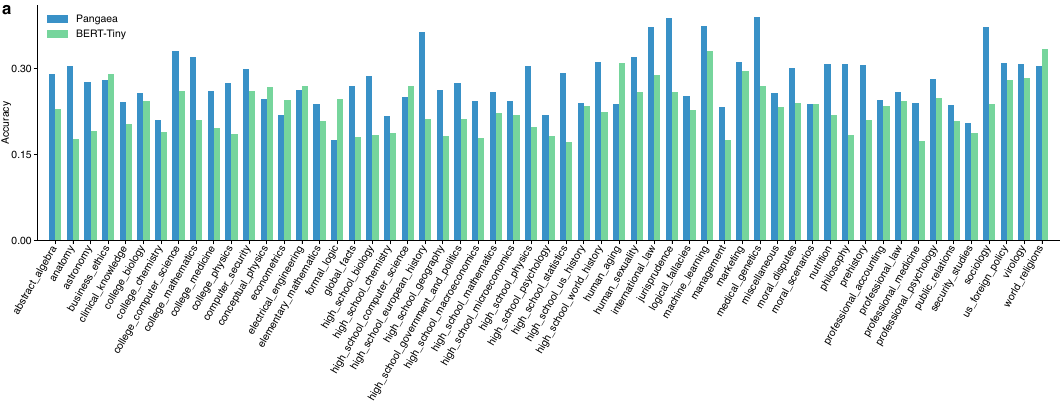}
    \caption{\textbf{Performance comparisons of \method{} on the massive multitask language understanding task.} \method{} consistently outperforms BERT-Tiny across almost all 57 sub-tasks.
    }\label{app_fig:mmlu}
\end{figure}

\clearpage

\begin{figure}[h]
    \centering
    \includegraphics[width=\textwidth,keepaspectratio=true]{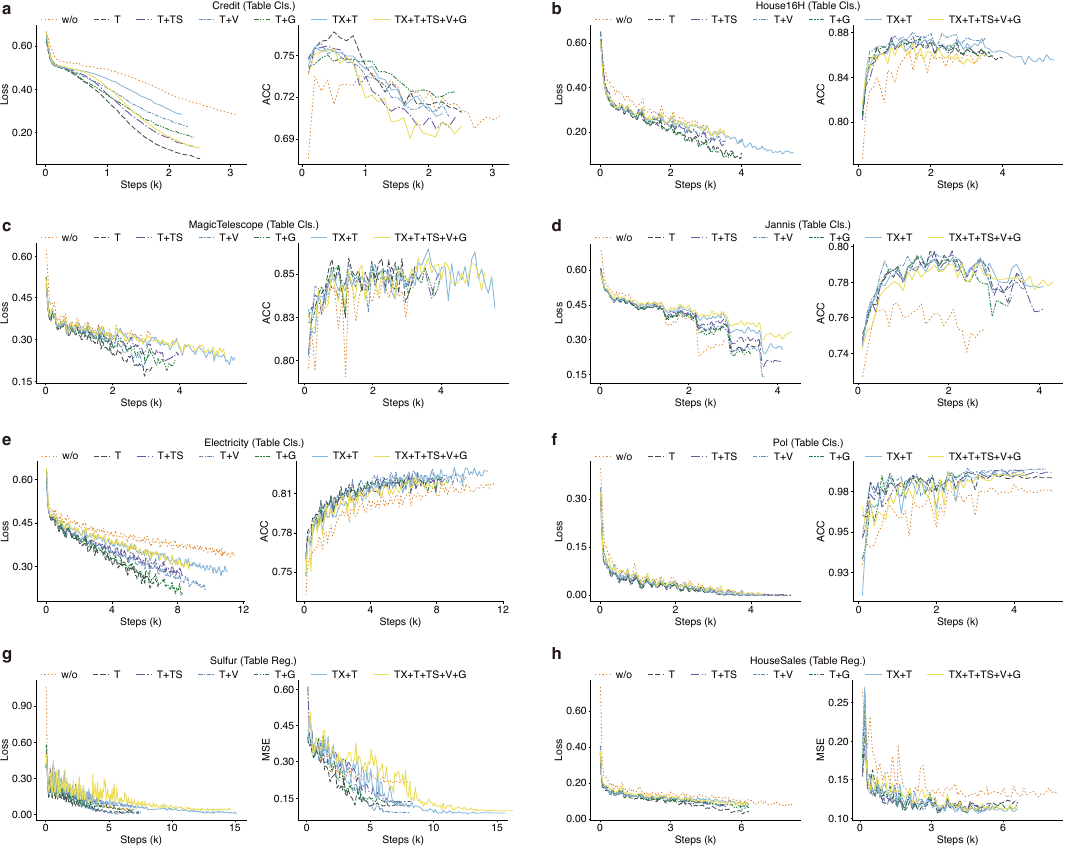}
    \caption{\textbf{Convergence curves of \method{} across different pre-training modality combinations in downstream fine-tuning.} Five-modal pre-trained models consistently outperform in downstream tasks. \textbf{a}-\textbf{h}, Precision curves (ACC or MSE) and fine-tuning loss across various modality-specific downstream tasks, including table classification and regression. 
    }\label{app_fig:finetuning_modal_combination1}
\end{figure}

\clearpage

\begin{figure}[h]
    \centering
    \includegraphics[width=\textwidth,keepaspectratio=true]{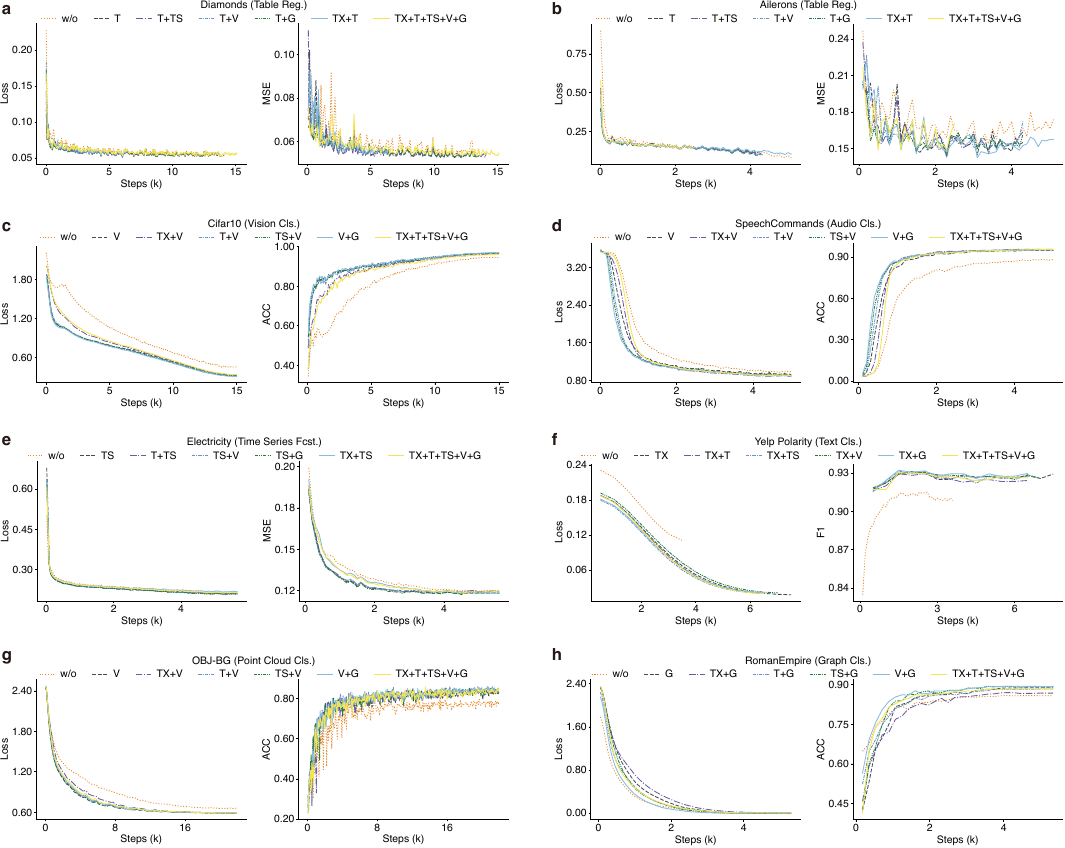}
    \caption{\textbf{Convergence curves of \method{} across different pre-training modality combinations in downstream fine-tuning.} Five-modal pre-trained models consistently outperform in downstream tasks. \textbf{a}-\textbf{h}, Precision curves (ACC or MSE) and fine-tuning loss across various modality-specific downstream tasks, including table regression, vision classification, audio classification and time series forecasting. 
    }\label{app_fig:finetuning_modal_combination2}
\end{figure}

\clearpage

\begin{figure}[h]
    \centering
    \includegraphics[width=\textwidth,keepaspectratio=true]{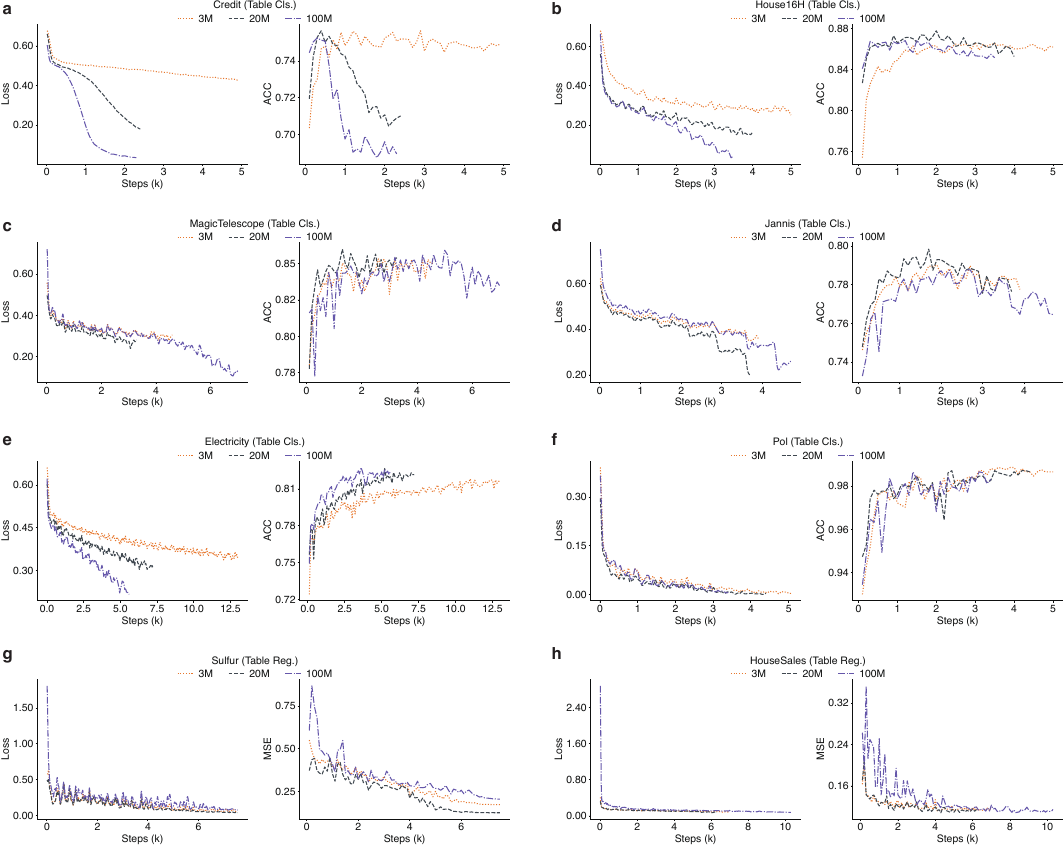}
    \caption{\textbf{Convergence curves of four-modal pre-trained \method{} across different pre-training model parameters (3M, 20M, 100M) in downstream fine-tuning.} Larger models improve performance but overfit with scarce data. \textbf{a}-\textbf{h}, Precision curves (ACC or MSE) and fine-tuning loss across various modality-specific downstream tasks, including table classification and regression. 
    }\label{app_fig:finetuning_model_scale1}
\end{figure}

\clearpage

\begin{figure}[h]
    \centering
    \includegraphics[width=\textwidth,keepaspectratio=true]{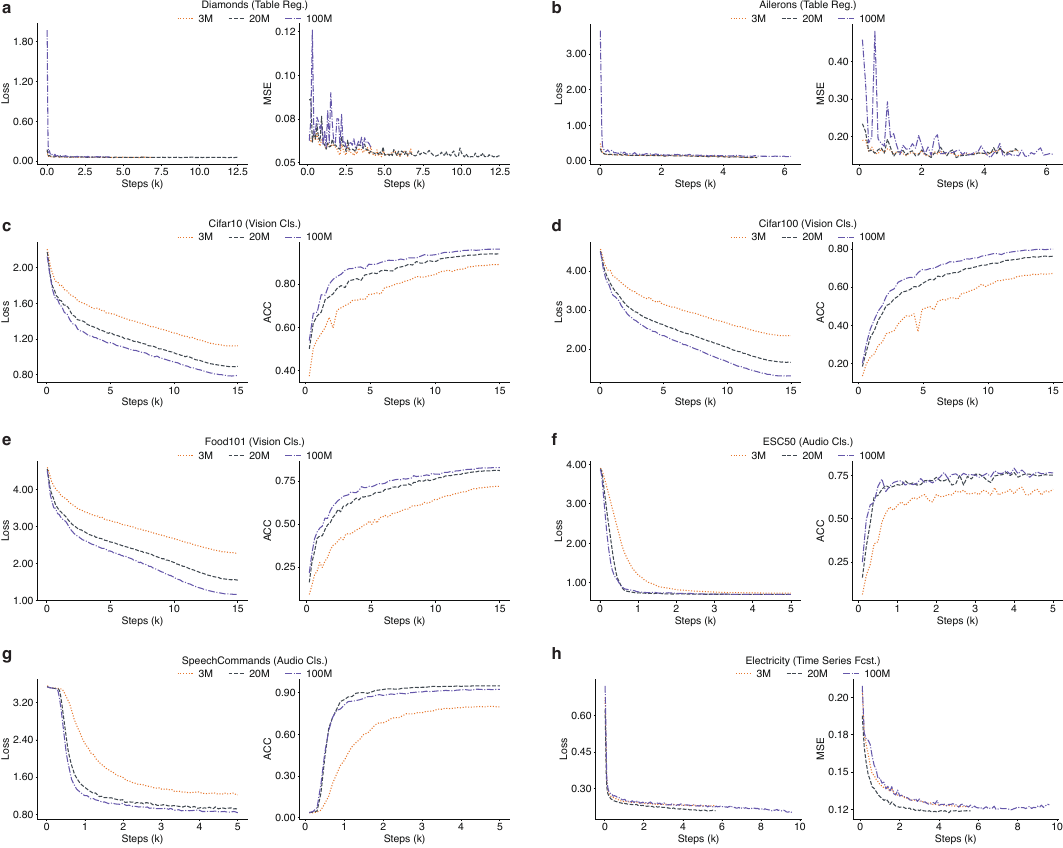}
    \caption{\textbf{Convergence curves of four-modal pre-trained \method{} across different pre-training model parameters (3M, 20M, 100M) in downstream fine-tuning.} Larger models improve performance but overfit with scarce data. \textbf{a}-\textbf{h}, Precision curves (ACC or MSE) and fine-tuning loss across various modality-specific downstream tasks, including table regression, vision classification, audio classification and time series forecasting. 
    }\label{app_fig:finetuning_model_scale2}
\end{figure}

\clearpage

\begin{figure}[h]
    \centering
    \includegraphics[width=\textwidth,keepaspectratio=true]{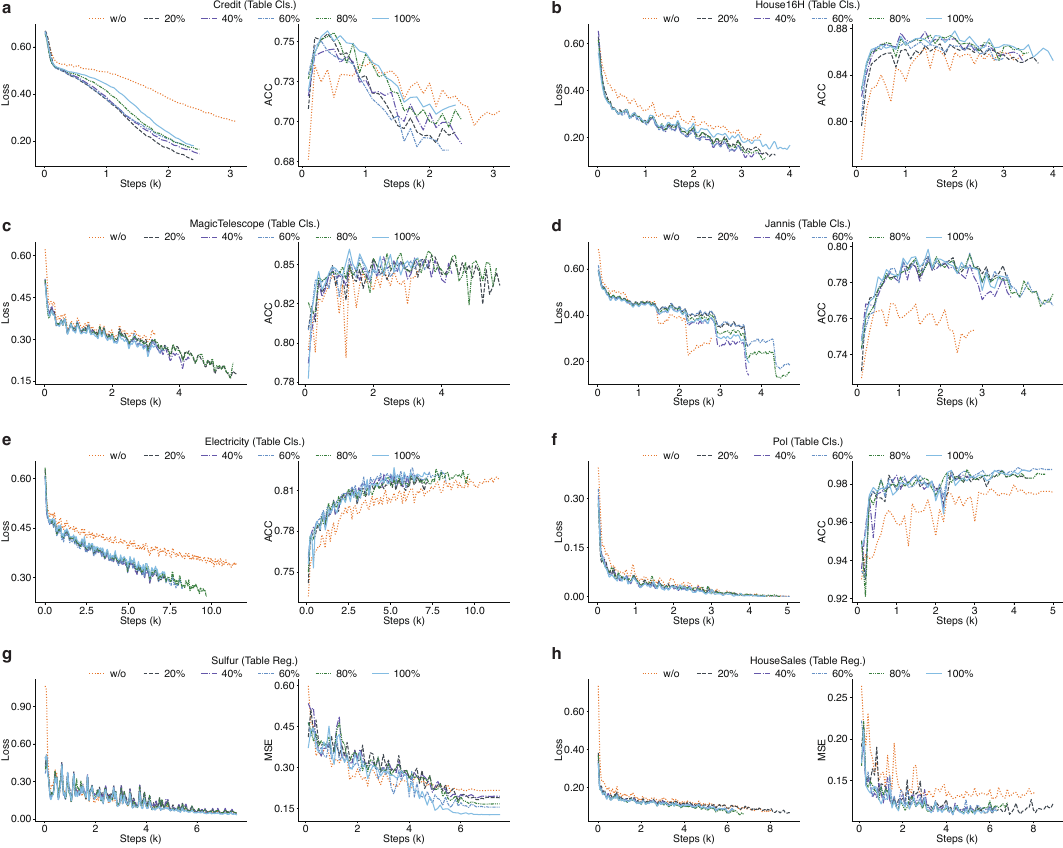}
    \caption{\textbf{Convergence curves of four-modal pre-trained \method{} across different percentages (0\% (w/o), 20\%, 40\%, 60\%, 80\%, 100\%) of pre-training data in downstream fine-tuning, with each pre-training modality sharing the same percentage.} More pre-training data lead to higher performance gains. \textbf{a}-\textbf{h}, Precision curves (ACC or MSE) and fine-tuning loss across various modality-specific downstream tasks, including table classification and regression. 
    }\label{app_fig:finetuning_precentage_data1}
\end{figure}

\clearpage

\begin{figure}[h]
    \centering
    \includegraphics[width=\textwidth,keepaspectratio=true]{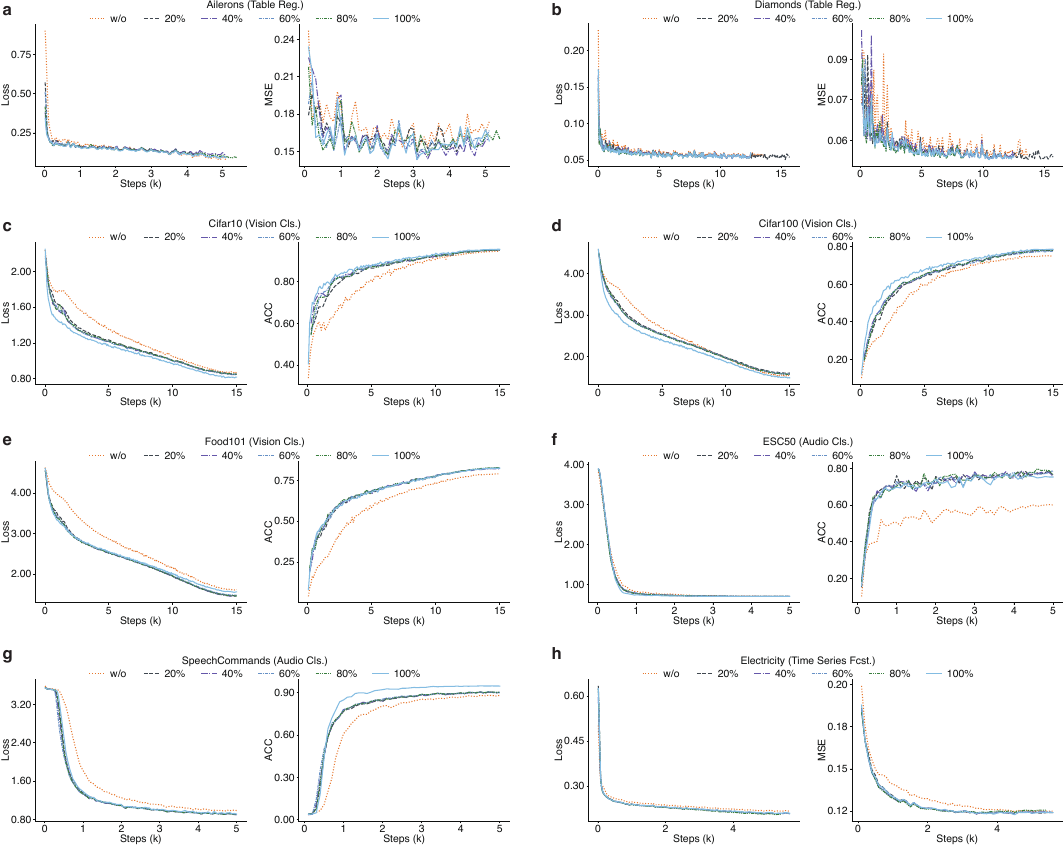} 
    \caption{\textbf{Convergence curves of four-modal pre-trained \method{} across different percentages (0\% (w/o), 20\%, 40\%, 60\%, 80\%, 100\%) of pre-training data in downstream fine-tuning, with each pre-training modality sharing the same percentage.} More pre-training data lead to higher performance gains. \textbf{a}-\textbf{h}, Precision curves (ACC or MSE) and fine-tuning loss across various modality-specific downstream tasks, including table regression, vision classification, audio classification and time series forecasting. 
    }\label{app_fig:finetuning_precentage_data2}
\end{figure}



\clearpage

\begin{table}[htbp]
\centering
\caption{Details of the pre-training datasets. For aesthetic consistency, all dataset names have been renamed using PascalCase. Datasets that share the same prefix are now renamed with the prefix-\{suffix1, suffix2, ...\} format.}
\label{tab:pretrain_data_size}
\newcolumntype{C}[1]{>{\centering\arraybackslash}p{#1}}
{\fontsize{6pt}{7pt}\selectfont
\setlength{\tabcolsep}{0.65mm}{
\begin{tabular}{@{}lcccp{10cm}}
\toprule
Modality & Domain    & Sample    & Number & \multicolumn{1}{c}{Name of datasets} \\ 
\midrule
\multirow{10}{*}{\raisebox{-40\height}{\textbf{Table}}} & Medicine   & 884,633  & 37 & \makecell[{{p{10cm}}}]{Crash, LungcancerShedden, OneHundredPlantsMargin, PrimaryTumor, Wisconsin, Pbc, Lowbwt, Pharynx, Pyrim, EchoMonths, BreastTumor, QsbrY2, Collins, Dionis, Helena, Diabetes, MedicalCharges, MedicalChargesNominal, Shapes, Lymph, ICU, EgyptianSkulls, BreastTissue, Cardiotocography, HeartLongBeach, HeartSwitzerland, VertebraColumn, HeartH, Hypothyroid, Sponge, Dermatology, Ecoli, Vowel, Pendigits, MfeatKarhunen, Soybean, AdultCensus} \\ \cmidrule{2-5}
& Economy    & 1,166,655    & 19 & \makecell[{{p{10cm}}}]{DeltaElevators, Triazines, AutoPrice, Elevators, Stock, PwLinear, Ailerons, HappinessRank2015, VancouverEmployee, Diamonds, Treasury, AirlinesDepDelay1M, House-\{Sales, Prices, SalesReduced, PricesNominal\}, Credit, AdaPrior, Confidence} \\ \cmidrule{2-5}
& Nature     & 6,964,534  & 82 & \makecell[{{p{10cm}}}]{QqdefectsNumeric, Brainsize, Acorns, BNG (Pharynx), Leaf, OneHundredPlants-\{Shape, Texture\}, Bolts, Cloud, CpuSmall, MachineCpu, Mv, MaunaLoaAtmosphericCo2, Andro, Jura, Rf1, Slump, Iris, RainfallBangladesh, CPMP2015-\{Regression, RuntimeRegression\}, CacaoFlavor, ForestFires, Laser, BalanceScale, MfeatFourier, MyIris, KrVsK, CovPokElec, Seeds, Volcanoes-\{a1, a2, a3, a4, b1, b2, b3, b4, b5, b6, c1, d1, d2, d3, d4, e1, e2, e3, e4, e5\}, AutoUnivAu6-\{1000, 750, 400\}, AutoUnivAu7-\{1100, 700, 500\}, PokerHand, Poker, MiniBooNE, Jm1, Mc2, Kc-\{1, 2, 3\}, BNG, CPUAct, Weather-\{Ankara, Izmir\}, Fishcatch, Airquality, Nursery, Abalone, Fruitfly, Covertype, SeismicBumps, WhiteClover, GrubDamage, CodrnaNorm, Pc-\{1, 2, 3, 4\} }\\ \cmidrule{2-5}
& Transport  & 1,143,622    & 10  & \makecell[{{p{10cm}}}]{Pubexpendat, AutoPrice, Titanic, Scm20d, DelaysZurichTransport, NycTaxiGreenDec2016, Datatrieve, Seattlecrime6, SubsampleDelaysZurichTransport, Bridges} \\ \cmidrule{2-5}
& Web        & 584,810  & 11  & \makecell[{{p{10cm}}}]{InternetUsage, ArticleInfluence, Youtube, NewFuelCar, TeachingAssistant, Desharnais, Badges2, KDDCup99, OnlineNewsPopularity, Kc1-\{Top5, Binary\}} \\\cmidrule{2-5}
& Industry   & 76,820  & 9  & Kin8nm, Puma-\{8NH, 32H\}, Sulfur, ESL, NasaNumeric, 2dplanes, SyntheticControl, Glass \\ \cmidrule{2-5}
& Society    & 1,907,006  & 25  & \makecell[{{p{10cm}}}]{FacultySalaries, Colleges, LEV, ERA, PopularKids, FirstOrderTheoremProving, BankMarketing, UserKnowledge, Ldpa, Mlr-\{Knn, Glmnet, Svm, Rpart\}-Rng, PageBlocks, USCrime, Baskball, Ozone, IpumsLa-\{98,99\}-Small,AartificialCharacters, Pokerhand, AnalcatdataReviewer, Flags, Cjs, AnalcatdataGermangss}  \\ \cmidrule{2-5}
& Retail & 1,687,392 & 4 & BlackFriday, ClickPredictionSmall, BrazilianHouses, House16H \\ \cmidrule{2-5}
& Energy     & 107,777  & 12  & \makecell[{{p{10cm}}}]{SMSA, EnergyEfficiency, AutoMpg, Gascons, Debutanizer, TamilnaduElectricity, Enb, DEE, YachtHydrodynamics, Electricity, BikeSharingDemand, ELE-2} \\ \cmidrule{2-5}
& Others     & 1,499,793  & 34 & \makecell[{{p{10cm}}}]{Kc1Numeric, MercuryinBass, WalkingActivity, AutoHorse, Cholesterol, House8L, Moneyball, ShortTrackSpeedSkating, Selwood, DatasetSales, SoilKsatDB, SWD, EyeMovements, Usp-\{05, 05-ft\}, WallRobotNavigation, Mozilla4, Mc1, Ar-\{1, 6\}, Mw1, Airlines, SpokenArabicDigit, Pol, MagicTelescope, FpsInVideoGames, RobotFailures-\{lp1, lp2, lp3, lp4, lp5\}, HumansNumeric, DatasetAutoHorseFixed, AnalcatdataDmft} \\ \cmidrule{2-5}
& Total     & \textbf{16,023,042}  & \textbf{243}  & - \\ \midrule
\multirow{8}{*}{\raisebox{-6\height}{\textbf{\shortstack{Time\\Series}}}} & Medicine   & 139,929  & 3 &  Hospital, USAirQuality1980present, AirQualityMadrid\\ \cmidrule{2-5}
&Economy    & 327,627    & 17 &  \makecell[{{p{10cm}}}]{Cif2016, ExchangeRate, FredMd, Nn5DailyWithMissing, Nn5Weekly, RideshareWithoutMissing, TourismMonthly, M4-\{Daily, Hourly, Monthly, Quarterly, Yearly, Weekly\}, Nasdaq100, QlibCNDaily-\{B, D\}, QlibCN1min} \\ \cmidrule{2-5}
& Transport  & 23,920   & 6  & Airpassengers, PedestrianCounts, Taxi30min, UberTlc-\{Daily, Hourly\}, VehicleTripsWithoutMissing \\ \cmidrule{2-5}
& Nature     & 972,422  & 6 & \makecell[{{p{10cm}}}]{CovidDeaths, Saugeenday, TemperatureRainWithoutMissing, Weather, GermanTemperatureData10minSelected, AkdotTdpTemperature} \\ \cmidrule{2-5}
& Web        & 28,605  & 2  & KaggleWebTrafficWeekly, WikiRollingNips\\ \cmidrule{2-5}
&Energy     & 1,976,988  & 12  & \makecell[{{p{10cm}}}]{AustralianElectricityDemand, Electricity, ElectricityWeekly, KddCup2018WithoutMissing, LondonSmartMetersWithoutMissing, SolarEnergy, Solar10Minutes, WindFarmsWithoutMissing, EuropeanSolarGeneration, HouseholdPowerConsumption, SimulatedCountryAggregatedHeatDemandAndCop} \\ \cmidrule{2-5}
&Others     & 2,400  & 3 &  Seasonal, TrendK-1-0, TrendK-0-1 \\ \cmidrule{2-5}
&Total     & \textbf{3,471,891}  & \textbf{48}  & - \\ \midrule
\textbf{\shortstack{Vision}} & Nature & 1,281,167 & 1 & ImageNet \\
\midrule
\textbf{\shortstack{Text}} & Other & 6,458,670 & 1 & English Wikipedia \\
\midrule
\textbf{\shortstack{Graph}} & Society & 917,127 & 3 & OgbnArxiv, OgbnMag, OgbnProducts \\
\midrule
\textbf{Total} & - & \textbf{28,151,897} & \textbf{296} & -\\
\bottomrule
\end{tabular}}}
\end{table}

\clearpage


\begin{table}[htbp]
\centering
\caption{Details of triplet construction for different modalities. For table and graph data, $d$ denotes the sample dimension. In vision data, 75\% of triplets are masked during pre-training, with the remaining 49 triplets (\(196 \times 25\% = 49\)) used to reconstruct the masked triplets. For point cloud data, the input has shape \(g \times k \times 3\), where \(g\) is the number of center points and each has \(k\) nearest neighbors represented by their 3D coordinates (x, y, z). Note that audio and point cloud data are not used during pre-training.}
\label{tab:triplet_info}
\begin{tabular}{@{}>{\raggedright\arraybackslash}p{3cm}p{3cm}p{3cm}p{3cm}@{}}
\toprule
\textbf{Modality} & \textbf{Input Dimension} & \textbf{Triplets in Pre-training} & \textbf{Triplets in Fine-tuning} \\
\midrule
Table       & \(d\) & \(d\)  & \(d\) \\
Time Series & 256   & 8     & 8 \\
Text        & 512   & 256   & 256\\
Vision      & \(224 \times 224 \times 3\) & 49    & 196 \\
Graph       & \(d\) & 32    & 32 \\
Audio       & \(512 \times 128 \times 3\) & --    & 256 \\
Point Cloud & \(g\times k\times 3\)     & -- & \(g/2\)\\
\bottomrule
\end{tabular}
\end{table}

\clearpage

\begin{table}[htbp]
\centering
\caption{Fine-tuning results for 45 general tasks, referenced from Fig.~\ref{fig:main}e. Higher values indicate better performance for ACC, AUC, and F1, while lower values are preferred for MSE.}
\label{tab:mian_res_info}
{\fontsize{8pt}{9pt}\selectfont
\begin{tabular}{@{}>{\raggedright\arraybackslash}p{2.9cm}p{2.8cm}p{0.6cm}p{1.3cm}p{1.2cm}p{1.4cm}@{}}
\toprule
Downstream Task & Dataset & Metric & \method{}$_\text{w/o}$ & \method{} & Competitors \\
\midrule
Table Classification&	Credit& ACC&	0.741&	0.755& 0.720\\
Table Classification&	Electricity& ACC&	0.821&	0.827& 0.765\\
Table Classification&	Pol& ACC&	0.979&	0.989& 0.968\\
Table Classification&	House16H& ACC&	0.866&	0.873& 0.866\\
Table Classification&	MagicTelescope& ACC&	0.847&	0.860& 0.834\\
Table Classification&	MiniBooNE& ACC&	0.918&	0.930& 0.924\\
Table Classification&	EyeMovements& ACC&	0.588&	0.600& 0.574\\
Table Classification&	DefaultOfCreditCardClients& ACC&	0.714&	0.718& 0.712\\
Table Classification&	Jannis& ACC&	0.770&	0.791& 0.780\\
Table Classification&	Diabetes130US& ACC&	0.603&	0.605& 0.604\\
Table Classification&	California& ACC&	0.883&	0.887& 0.818\\
Table Regression&	CpuAct& MSE&	0.017& 	0.015& 0.026\\
Table Regression&	Pol& MSE&	0.005&	0.003& 0.009\\
Table Regression&	Elevators& MSE&	0.084&	0.079& 0.109\\
Table Regression&	WineQuality& MSE&	0.572&	0.508& 0.510\\
Table Regression&	Ailerons& MSE&	0.151&	0.146& 0.156\\
Table Regression&	Houses& MSE&	0.184&	0.158& 0.159\\
Table Regression&	Diamonds& MSE&	0.054&	0.052& 0.052\\
Table Regression&	BrazilianHouses& MSE&	0.007&	0.006& 0.009\\
Table Regression&	BikeSharingDemand& MSE&	0.322&	0.297& 0.309\\
Table Regression&	HouseSales& MSE&	0.129&	0.108& 0.111\\
Table Regression&	Sulfur& MSE&	0.156&	0.120& 0.183\\
Table Regression&	MiamiHousing2016& MSE&	0.081&	0.072& 0.077\\
Table Regression&	YProb41& MSE&	0.812&	0.799& 0.805\\
Time Series Forecasting&	ETTh1& MSE&	0.388&	0.382& 0.386\\
Time Series Forecasting&	ETTh2& MSE&	0.300&	0.294& 0.297\\
Time Series Forecasting&	ETTm1& MSE&	0.325&	0.313& 0.334\\
Time Series Forecasting&	ETTm2& MSE&	0.179&	0.178& 0.180\\
Time Series Forecasting&	Electricity& MSE&	0.148&	0.147& 0.148\\
Time Series Forecasting&	Weather& MSE&	0.166&	0.166& 0.174\\
Vision Classification&	Cifar10& ACC&	0.947&	0.971& 0.963\\
Vision Classification&	Cifar100& ACC&	0.753&	0.783& 0.760\\
Vision Classification&	Food101& ACC&	0.790&	0.807& 0.798\\
Audio Classification&	ESC50& ACC&	0.638&	0.748& 0.664\\
Audio Classification&	SpeechCommands& ACC&	0.881&	0.970& 0.954\\
Graph Classification&	Roman-Empire& ACC&	0.855&	0.881& 0.652\\
Graph Classification&	Minesweeper& AUC&	0.885& 0.889& 0.882\\
Graph Classification&	Tolokers& AUC&	0.784&	0.787& 0.778\\
Point Cloud Classification&	OBJ-ONLY& ACC&	0.835&	0.857& 0.843\\
Point Cloud Classification&	OBJ-BG& ACC&	0.809&	0.855& 0.823\\
Point Cloud Classification&	PB-T50-RS& ACC&	0.752&	0.787& 0.779\\
Text Classification&	Emotion Balanced& F1&	0.853&	0.920& 0.916\\
Text Classification&	BoolQ& F1&	0.643&	0.660& 0.657\\
Text Classification&	Yelp Polarity& F1&	0.916&	0.960& 0.941\\
Text Classification&	MRPC& F1&	0.687&	0.723& 0.711\\
\bottomrule
\end{tabular}
}
\end{table}

\clearpage

\begin{table}[htbp]
\centering
\caption{Detailed results for 15 scientific tasks. Higher values indicate better performance for ACC, AUC, and F1, while lower values are preferred for MAE and RMSE.}
\label{tab:ai4s_res_info}
{\fontsize{8pt}{10pt}\selectfont
\begin{tabular}{@{}>{\raggedright\arraybackslash}p{5.5cm}p{4cm}p{0.6cm}p{1.2cm}p{1.4cm}@{}}
\toprule
Scientific Task & Scientific Subject & Metric & \method{} & Competitors \\
\midrule
Marine mammal vocalization classification&	Earth and environmental sciences
& ACC& 0.966& 0.962\\
Global temperature forecasting & Earth and environmental sciences& RMSE& 2.453& 3.204\\
Reservoir property estimation & Earth and environmental sciences & RMSE& 0.075& 0.106\\
Blood brain barrier permeability prediction&	Health sciences  & AUC& 0.941& 0.930\\
Prostate cancer grading&	Health sciences & ACC& 0.790& 0.701\\
Drug consumption prediction&	Humanities  & AUC& 0.756& 0.708\\
Stock movement prediction&	Business and commerce  & F1& 0.655& 0.605\\
Massive multitask language understanding&	Social science  & ACC& 0.278& 0.262\\
Cyclic peptide membrane permeability prediction&	Biological sciences  & MAE& 0.222& 0.355\\
Drug toxicity prediction&	Biological sciences  & AUC& 0.764& 0.730\\
Mathematical article subject classification&	Mathematics  & ACC& 0.674& 0.637\\
Active galactic nuclei classification&	Astronomy sciences  & ACC& 0.876& 0.869\\
High-energy particle identification&	Physical sciences  & AUC& 0.810& 0.798\\
Molecule electronic properties prediction&	Physical sciences  & MAE& 67.780& 81.900\\
Material band gap prediction&	Physical sciences  & MAE& 161.600& 253.030\\
\bottomrule
\end{tabular}
}
\end{table}

\end{document}